\documentclass[10pt]{article} 
\usepackage[accepted]{tmlr}

\usepackage{amsmath,amsfonts,bm}

\def\eqref#1{equation~\ref{#1}}

\def\1{\bm{1}}

\DeclareMathAlphabet{\mathsfit}{\encodingdefault}{\sfdefault}{m}{sl}
\SetMathAlphabet{\mathsfit}{bold}{\encodingdefault}{\sfdefault}{bx}{n}

\usepackage{url}

\usepackage[utf8]{inputenc} 
\usepackage[T1]{fontenc}    
\usepackage[colorlinks=true,allcolors=blue]{hyperref}
\usepackage{booktabs}       
\usepackage{amsfonts}       
\usepackage{nicefrac}       
\usepackage{microtype}      
\usepackage{xcolor}         
\usepackage{amsmath}
\usepackage{cleveref}
\usepackage{xspace}
\usepackage{enumitem}
\usepackage{graphicx}
\usepackage{caption}
\usepackage{booktabs}
\usepackage{makecell}
\usepackage{multirow}
\usepackage{adjustbox}
\usepackage{array}
\usepackage[caption=false]{subfig}

\title{Soft Merging of Experts with Adaptive Routing}

\author{\name Mohammed Muqeeth \email muqeeth101@gmail.com \\
      \addr University of North Carolina at Chapel Hill
      \AND
      \name Hoakun Liu \email haokunliu412@gmail.com \\
      \addr University of Toronto \\
      Vector Institute
      \AND
      \name Colin Raffel \email craffel@gmail.com\\
      \addr University of Toronto \\
      Vector Institute}

\def\month{05}  
\def\year{2024} 
\def\openreview{\url{https://openreview.net/forum?id=7I199lc54z}} 

\begin{document}

\maketitle

\begin{abstract}
Neural networks that learn to route their inputs through different ``expert'' subnetworks provide a form of modularity that standard dense models lack.
Despite their possible benefits, modular models with learned routing often underperform their parameter-matched dense counterparts as well as models that use non-learned heuristic routing strategies.
In this paper, we hypothesize that these shortcomings stem from the gradient estimation techniques used to train modular models that use non-differentiable discrete routing decisions.
To address this issue, we introduce \textbf{S}oft \textbf{M}erging of \textbf{E}xperts with \textbf{A}daptive \textbf{R}outing (SMEAR), which avoids discrete routing by using a single ``merged'' expert constructed via a weighted average of all of the experts' parameters.
By routing activations through a single merged expert, SMEAR does not incur a significant increase in computational costs and enables standard gradient-based training.
We empirically validate that models using SMEAR outperform models that route based on metadata or learn routing through gradient estimation.
Furthermore, we provide qualitative analysis demonstrating that the experts learned via SMEAR exhibit a significant amount of specialization.
All of the code used in our experiments is publicly available.\footnote{\url{https://github.com/r-three/smear}}
\end{abstract}

\section{Introduction}

Neural networks typically use all of their parameters to process a given input.
As such, the capabilities of a model are distributed across the parameters of a model in a self-organizing way \citep{zeiler2014visualizing,de2021editing,csordas2021neural,bau2020understanding,wang2022finding}.
Explicitly specializing different parts of a model to different capabilities can provide various benefits, including reduced interference across downstream tasks \citep{sanh2021multitask, wei2021finetuned, zamir2018taskonomy, bao2021beit} or languages \citep{pires2019multilingual, liu2020multilingual, xue2020mt5}.
Furthermore, dedicating specific parameters to specific capabilities enables a form of modularity where a capability can be added, removed, or modified by adding, removing, or modifying the corresponding parameters \citep{pfeiffer2023modular}.
Activating only a subset of the model's parameter for a given input also decouples the computational cost of a model from the number of parameters it has \citep{shazeer2017outrageously,fedus2021switch}, though we do not focus on this benefit in this paper.

\textit{Conditional computation} techniques provide a way to build models that adaptively choose a subset of their parameters to apply to a given input.
A common way to use conditional computation in this setting is to introduce specialized subnetworks called \textit{experts} that are controlled by \textit{routers} that decide which experts should be active.
When the model is trained on diverse data, this form of conditional computation can enable modular learning by allowing experts to specialize to different types of inputs and flexibly share knowledge \citep{ma2019snr}.
However, because routing involves making a discrete decision as to which expert to use, the loss on the model's prediction cannot back-propagate though the routing decision to update the router. 
Consequently, models with conditional computation often require gradient estimation techniques for training \citep{clark2022unified, fedus2021switch, bengio2013estimating}.
In practice, past work has shown that models with conditional computation do not always learn effective routing strategies.
For example, \citet{mittal2022modular} investigate models with a continuous router in a controlled setting and find the models do not route examples from the same group to the same experts and perform poorly compared to models with oracle routing.
However, models with task- or domain-specific subnetworks \citep{gururangan2021demix, kudugunta2021beyond} provide evidence that it is possible to train performant models with specialized experts.
As an extreme example, \citet{roller2021hash} achieves results comparable to learned routing with a fixed random routing.
Relatedly, \citet{fedus2021switch} find the gain from scaling up parameters by 30$\times$ with a sparsely activated model is smaller than scaling up both parameters and FLOPs by 3$\times$ in a dense model.
As a possible explanation, \citet{clark2022unified} study how models with conditional computation improve with scale and find a detrimental term that scales with the product of the log number of experts and active parameters.

\begin{figure*}
\centering
{\includegraphics[width=0.9\linewidth]{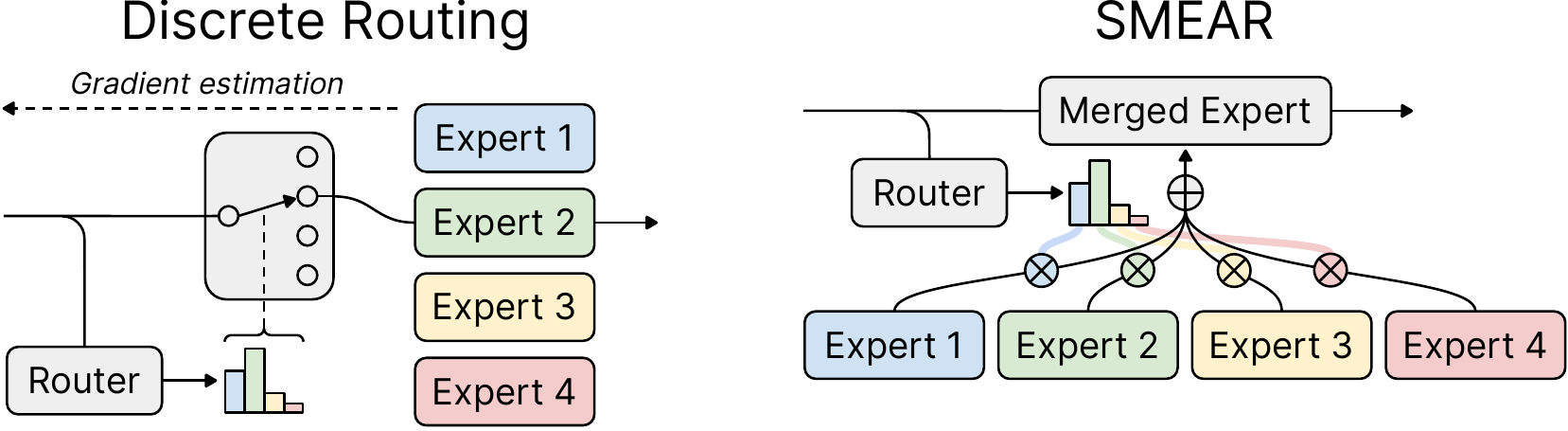}\label{fig:smear_vs_cc}}
\caption{The discrete routing decisions commonly used in models that route activations among experts require the use of gradient estimation (left). We propose SMEAR (right), which uses a given router's distribution to average the parameters of the corresponding experts and then routes the input through a single merged expert. SMEAR achieves better performance than models with discrete routing, can be trained with standard backpropagation, and does not incur significant additional computational costs.}
\end{figure*}

In this work, we hypothesize that issues with conditional computation stem from issues with gradient estimation.
Specifically, we focus on experimental settings where we can compare learned routing to a performant hand-designed heuristic routing scheme.
We find that the gradient estimation techniques we consider often produce models that underperform heuristic routing, despite the fact that they could in principle learn a better routing strategy.
To address this shortcoming, we introduce \textbf{S}oft \textbf{M}erging of \textbf{E}xperts with \textbf{A}daptive \textbf{R}outing (SMEAR), a method for training modular models with specialized experts and learned routing.
Given an input, which may be a sequence of tokens or an image, 
SMEAR works by using the router's distribution of the input over experts to compute a weighted average of the parameters of the individual experts.
The input activations are then sent through the \textit{merged} expert, which results in a similar computational cost to discrete routing with a single expert.
However, the fact that all components of SMEAR are fully differentiable enables standard gradient-based training.
Empirically, we show that SMEAR significantly attains a favorable performance/cost tradeoff to 1) discrete routing solutions found via gradient estimation, 2) heuristic routing schemes, and 3) state-of-the-art baselines for learning modular models.
We also qualitatively validate that the experts learned by SMEAR specialize to different types of inputs and share parameters across related tasks.
Put together, our results show that SMEAR provides an effective alternative for modular models that use adaptive routing among expert subnetworks.

After providing background on conditional computation models and gradient estimators in the following section, we define SMEAR in \cref{sec:smear}.
We then describe our experimental findings in \cref{sec:setup}, discuss related works in \cref{sec:related}, and conclude in \cref{sec:conclusion}.

\section{Background}
\label{sec:background}

To provide the necessary background for our work, we first explain how sparsely activated neural networks use conditional computation, then discuss gradient estimators that enable learning discrete routing strategies.
In addition, we discuss different ways to hand-design ``heuristic'' routing strategies as well as preexisting techniques for learning modular models that we use as baselines.

\subsection{Routing Among Experts}

In models that use discrete routing among experts (i.e.\ subnetworks), experts are organized into blocks that are incorporated as an intermediate layer in a neural network. 
An expert routing block $B$ comprises a set of $N$ experts $\{f_1, f_2, \dots, f_N\}$ and a router $R$. 
Experts in the same block accept inputs and produce outputs of the same dimensionality.
Given a hidden-state representation $u$, the output of the $i$-th expert with parameters $\theta_i$ is $f_i(u, \theta_i)$.  

\subsection{Gradient Estimators}
\label{sec:estimators}

In sparsely activated models that involve discrete adaptive routing, it is not possible to train the router's parameters with standard gradient-based learning.
Fortunately, gradient estimators can provide approximate gradients to the router parameters.
There are a few common designs shared by models that use gradient estimators to train routers. 
Their router $R$ often applies a lightweight network to some intermediate hidden states $v$ in the model.
The output of the lightweight routing network $R(v)$ parameterizes a discrete probability distribution over the $N$ experts.
Different gradient estimators vary in how they make the routing decision from $R(v)$ and how they construct the output from the chosen expert. 
Additionally, some estimators may introduce additional loss terms.

\paragraph{REINFORCE}
Gradients can be estimated through discrete operations using reinforcement learning techniques \citep{schulman2015gradient, bengio2013estimating}.
In reinforcement learning, a policy loss is used to train an agent to learn optimal actions in an environment. 
In this paper, we experiment with the REINFORCE algorithm which computes the policy loss as $\log(\pi) r$ where $r$ denotes the received reward for taking an action whose assigned probability is $\pi$.
When applied to models that use discrete routing among experts, the goal is to train the model to choose the optimal expert to process a given input. 
Here, the router $R$ acts as an agent that samples an expert to use according to the routing probabilities.
In order to train such a router, the router's assigned probability to the sampled expert is used as $\pi$ and the negative of the model's loss is used as the reward $r$.
The router is therefore trained to pick experts that maximize the reward which, in turn, minimizes the loss.
The REINFORCE estimator often suffers from high variance because of the sampling operation. 
This motivates the use of baselines, which reduce variance without changing the optimal solution.
In our work, we follow \citet{clark2022unified} and use a baseline $b$ that is generated by a small neural network with a single hidden layer that takes as input $v$ and is trained with the Huber loss.
The overall loss function is then
\begin{align*}
\label{eq:reinforce_estimator_eq}
    L = \;- \mathbb{E}_{i \sim R(v)} \alpha \log R(v)_i (r - b) - \beta R(v) \log R(v) + \gamma L_{\text{Huber}}(r,b)
\end{align*}
where $\alpha$, $\beta$, and $\gamma$ are hyperparameters that correspond to policy gradient weight, policy entropy weight, and value loss weight.
In practice, we approximate the expectation with a single sample.
During inference, the output of the block $B$ is just $f_i(u, \theta_i)$ where $i = \mathrm{arg}\max R(v)$.

\paragraph{Straight Through Gumbel-Softmax (ST-Gumbel)}  
The Gumbel-Softmax trick \citep{jang2016categorical,maddison2016concrete} provides a continuous differentiable approximation to sampling from a categorical distribution like the one parameterized by a router.
Specifically, Gumbel noise is added to the logits of the distribution and a temperature scale is applied in the softmax operation.
Denoting $g_i \sim \text{Gumbel(0, 1)}$ and $\tau$ as the temperature, the Gumbel-Softmax trick produces the following modified distribution:
\begin{equation*}
\label{eq:st_gumbel_estimator_eq}
    \hat{R}(v)_i = \frac{\exp((\log(R(v)_i) + g_i)/\tau)}{\sum_{j = 1}^N \exp((\log(R(v)_i) + g_i)/\tau)}
\end{equation*}
The expert $f_i$ with the highest assigned probability is chosen by applying an $\mathrm{arg}\max$ operation. 
In order to approximate gradients through the $\mathrm{arg}\max$ operation, we use the Straight-Through estimator which replaces $f_i(u, \theta_i)$ with $(1 - \text{sg}[\hat{R}(v)_i] + \hat{R}(v)_i) f_i(u, \theta_i)$ where $\text{sg}$ stands for the stop-gradient operator.
During forward pass, the multiplier for $f_i(u, \theta_i)$ becomes 1 and the multiplier receives gradients for the term $\hat{R}(v)_i$ in the backward pass.
In practice, the temperature $\tau$ is gradually annealed from a high to low value so that the approximated samples are more and more similar to discrete samples.
During inference, we choose an expert according to $\mathrm{arg}\max R(v)$.

\paragraph{Top-$k$}
\citet{shazeer2017outrageously} propose a gradient estimation scheme where the router sends the input through the $k$ experts that are assigned the highest probability.
\citet{fedus2021switch} later found that this router could be used effectively when $k = 1$. 
Specifically, the estimator selects the subnetwork with the highest probability and scales its output using its corresponding routing probability.
The output of the block is therefore $R(v)_i f_i(u, \theta_i)$, where $i= \mathrm{arg}\max R(v)$.

\paragraph{DSelect-$k$}
\citet{hazimeh2021dselect} proposed a differentiable approximation for the discrete Top-$k$ operation.
They parameterize each selection among $N$ experts using $m$ binary variables $z_1, z_2, \ldots z_m$, produced using a learnable weight $W$ as $z = W(v)$, where $m =\log_2(N)$. 
The selector function $r$ takes these variables and computes 
\begin{equation*}
r(z)_i = \prod_{j \in B(i-1)} (z_j) \prod_{j \in {1, \ldots, m} \setminus B(i-1)} (1-z_j)
\end{equation*}
where $i \in {1, \ldots, N}$ and $B(l)$ returns the non-zero indices in the binary representation of the integer $l$.
A differentiable step function $S$ based on a cubic polynomial is used for the $z$ variables. Finally, each selector operation happens $k$ times to get Top-$k$ selection and an additional learnable parameter $G$ provides a probability distribution over these $k$ selections using the softmax fuction.
In addition, entropy regularization is applied to the output of the selector function $r$, ensuring that it results in one-hot selection during inference. 
In our work, we consider $k=1$ to maintain a computational cost similar to other baselines.

\subsection{Heuristic Routing}
\label{sec:heuristic}

As a point of comparison for techniques that learn adaptive routing, we experiment with three baseline routing strategies that do not require a trained router.

\paragraph{Tag Routing}
If we have prior knowledge about the data that a model will be applied to, we can hand-design a heuristic routing strategy for choosing which expert to use for a given example based on data properties.
Tag routing takes advantage of ``tags'' associated with a given example (such as its domain or task) and associates each expert in an expert routing block with a particular tag.
In this work, we assume each example has a single tag and route each example to its tag's expert.

\paragraph{Hash Routing}
\citet{roller2021hash} propose hash routing, which uses a fixed hashing function to determine which expert to use for a given example.
Specifically, each example is assigned a random expert choice in each expert routing block which is used consistently over the course of training and inference.
This approach disregards any shared characteristics across examples.

\paragraph{Single-Expert}

As an additional baseline, we consider models where all inputs are routed to a single expert in each routing block. 
To provide a fair comparison to models with $N$ experts per block on the basis of both computational cost or parameter count, we consider models with a single expert with either the same number (compute-matched, referred to as ``$1\times$ compute'') or $N\times$ (parameter-matched, referred to as ``$1\times$ parameters'') as many parameters as a single expert.

\subsection{Methods for Learning Modular Models}
\label{sec:modular_baselines}

Beyond the simple baselines discussed above, we consider three recently proposed methods that aim to learn modular models.

\paragraph{Adamix}
Adamix \citep{wang2022adamix} uses random routing for each example during training and adds a consistency loss to encourage experts to share information and discourage divergence.
During inference, the parameters of all experts are averaged together to form a single expert and no adaptive routing is used.

\paragraph{Latent Skills}
Latent Skills \citep{ponti2022combining} assumes that the task for each example is known and trains a task-skill matrix that specifies which experts are active for a given task.
The binary task-skill matrix is fixed and learned via the Gumbel-Sigmoid trick \citep{maddison2016concrete}. 
During inference, a merged expert is formed for each task by averaging the parameters of the skill experts weighted according to the task-skill matrix. 

\paragraph{Soft MoE}
\citet{puigcerver2023sparse} recently proposed Soft MoE, which assigns ``slots'' to each expert and passes a weighted average of input tokens into each slot.
All operations in Soft MoE method are differentiable, avoiding the need for the gradient estimation. 
We consider Soft MoE with a single slot per expert to ensure fair comparison by having computational cost equivalent to other discrete routing baselines. 

\section{Soft Merging of Experts with Adaptive Routing}
\label{sec:smear}
As we will later show in \cref{sec:setup}, the gradient estimation techniques used to train models with discrete routing often fail to produce performant routing strategies.
Our goal in this work is therefore to explore whether it is possible to train models with adaptive routing among experts without resorting to gradient estimation.
Specifically, we aim to achieve better performance by designing an expert and router architecture that facilitates standard end-to-end gradient-based training but does not increase computational costs.

\paragraph{Ensemble Routing}

One simple idea would be to pass the input of a given expert routing block through \textit{every} expert, and then compute an average of the experts' outputs weighted according the router's distribution, i.e.\ exactly computing $\mathbb{E}_{i \sim R(v)} f_i(u, \theta_i)$.
We refer to this approach as an \textit{ensemble} routing strategy since it corresponds to using the ensemble prediction of the experts.
Since the operations involved in computing the average are all differentiable, using an ensemble routing strategy would allow for exact computation of gradients and end-to-end-learning.
Unfortunately, such an approach would incur a significant increase in computational costs because it requires computing the output of every expert rather than a single expert.

\paragraph{Merging Experts}

To explore an alternative fully-differentiable expert routing block, we take inspiration from recent work on \textit{merging} models \citep{matena2021merging,wortsman2022model,wortsman2022robust,choshen2022fusing,don2022cold,mcmahan2017communication}.
These works have shown that averaging the parameters of models that share a common architecture can often produce an aggregate model that shares the capabilities of the individual models.
Notably, \citet{wortsman2022model,matena2021merging} found that averaging the weights of multiple fine-tuned models produced a single model that performs comparably to an ensemble of the models.
In addition, both Adamix \citep{wang2022adamix} and Latent Skills \citep{ponti2022combining} include steps that involve averaging expert parameters, though neither of these methods learn an adaptive per-example routing strategy.
Motivated by these findings, we propose \textbf{S}oft \textbf{M}erging of \textbf{E}xperts with \textbf{A}daptive \textbf{R}outing (SMEAR), which constructs a single merged expert whose parameters are computed as the weighted average of the experts within a routing block.
Each expert's weight is set according to the corresponding routing probability generated by the router.
In SMEAR, the input to the routing block is fed into the merged expert and the merged expert's output is used as the output of the block.
By averaging parameters, SMEAR implicitly assumes that all experts in the routing block share an identical architecture (thereby inducing a natural one-to-one mapping between parameters in each expert).
To the best of our knowledge, all past works focused on routing among experts use experts with a common architecture, so we do not see this assumption as a major limitation.

More explicitly, we define SMEAR as computing the output of an expert routing block using a merged expert computed as
\begin{equation*}
\bar{f}(u, \sum_i R(v)_i \theta_i)
\end{equation*}
The merged expert shares the same architecture with the individual experts $f_i$.
Notably, the input of the routing block is only ever processed by $\bar{f}$; activations are never fed to any of the individual experts.
To break symmetry, all experts are randomly initialized with different parameter values.
Importantly, all operations in SMEAR are fully differentiable, enabling standard gradient-based end-to-end learning.
In addition, SMEAR retains the ability to learn an adaptive routing strategy that can route different examples to different experts without relying on hand-specified tags (as in Latent Skills and tag-based routing).
We will later show qualitatively that this leads to meaningful specialization of different experts in real-world experiments (\cref{sec:qualitative}).

\paragraph{Computational Costs}

Importantly, SMEAR only ever computes the output of a single expert, suggesting that SMEAR's computational cost could be comparable to single-expert discrete routing and significantly lower than ensemble routing.
However, the averaging operation in SMEAR incurs a nontrivial computational cost.
To quantify this cost, we focus on the common expert architecture comprising a dense layer that projects from $d$-dimensional activations to an $m$-dimensional vector followed by a nonlinearity and an additional dense layer projecting from $m$ dimensions back to $d$.
For simplicity, we ignore the (relatively minor) cost of the nonlinearity.
We assume the input is a length-$L$ sequence of activations with size $L \times d$.
In this case, computing the output of the merged experts incurs a computational cost of approximately $L \times 4 \times d \times m$ FLOPs and ensemble routing with $N$ experts would require $N \times L \times 4 \times d \times m$ FLOPs.
SMEAR additionally must average together the parameters of $N$ experts, which costs an additional $N \times 2 \times d \times m$ FLOPs.

Some past work on models with discrete routing has the router choose a different expert for each entry in the input sequence of activations (e.g. \citealp{fedus2021switch,lewis2021base,roller2021hash}).
This would require computing the expert average $L$ times, which would make the cost of SMEAR similar to that of ensemble routing.
We therefore focus on settings where models make a \textit{single} routing choice for an entire input example (e.g.\ \citealp{gururangan2021demix,kudugunta2021beyond,ye2022eliciting}).
This results in a total cost of approximately $(L \times 4 + N \times 2) \times d \times m$ for SMEAR.
Consequently, as long as $L \times 4 \gg N \times 2$, SMEAR and discrete routing have roughly the same computational costs.
Given that $L$ is on the order of hundreds or thousands of tokens for text-based tasks and on the order of thousands for vision tasks, $L \times 4$ will be much larger than $N \times 2$ as long as there is a modest number of experts.
In our experiments within the T5-GLUE setting, where $L = 128$ and $N = 8$, this results in a minimal runtime difference.
Furthermore, we would expect SMEAR to be approximately $\frac{N \times L}{N + L}$ times cheaper than ensemble routing.
More concretely, we will later experimentally validate 
that the wall-clock time required to process an example with SMEAR in real-world experiments is roughly the same as using discrete routing and significant faster than ensemble routing. 

\section{Experiments}
\label{sec:setup}
In order to thoroughly evaluate the effectiveness of SMEAR, we perform experiments in two real-world settings that differ in model architecture and modality.
We are particularly interested in whether a given approach for learning routing outperforms the heuristic routing strategies described in \cref{sec:background}.
As such, we focus on experimental settings where a performant ``tag routing'' baseline can be designed, i.e.\ where we have oracle access to metadata that can be used to appropriately route examples.
Specifically, we experiment with fine-tuning T5.1.1 Base \citep{raffel2020exploring} on datasets from GLUE \citep{wang2018glue} (referred to as T5-GLUE) and fine-tuning a ResNet18 \citep{he2016deep} on DomainNet \citep{peng2019moment} (ResNet-DomainNet).
In these settings, we add experts to an existing pre-trained backbone in the same way that Adapters are used for parameter-efficient fine-tuning \citep{houlsby2019parameter}.
While past work has also considered using discrete routing among experts to train large-scale models from scratch \citep{fedus2021switch,shazeer2017outrageously}, we focus on modular fine-tuned models in this work and leave large-scale experiments for future work.

\subsection{Settings}

\paragraph{T5-GLUE}

In this setting, we focus on training a T5 model \citep{raffel2020exploring} on the GLUE meta-benchmark \citep{wang2018glue} for natural language understanding. 
We provide background on the GLUE dataset and the example format we use in \cref{sec:t5_glue_hyps}.
We follow the approach of \citet{mahabadi2021parameter} for splitting each GLUE dataset into train, eval, and test splits.
Past work has demonstrated improved performance on RTE by co-training with MNLI \citep{phang2018sentence, devlin2018bert, pruksachatkun2020intermediate, vu2020exploring, choshen2022start}, and we congruously found that sharing an expert between RTE and MNLI produced a stronger tag routing strategy.
In the interest of making our baselines as strong as possible, we use this improved tag routing scheme in all experiments.
We use the pretrained T5.1.1 Base model as the backbone and adapt the model in a way similar to adding adapters \citep{houlsby2019parameter} for a single task, i.e.\ we keep all pretrained parameters frozen except for layer normalization parameters and insert expert routing blocks after self-attention, feed-forward and cross-attention modules.
The T5 1.1 Base model has 12 Transformer layers in both the encoder and decoder, resulting in a total of $12\times2 = 24$ blocks in the encoder and $12 \times 3 = 36$ blocks in the decoder, or $60$ expert routing blocks in total.
In each block, we introduce eight experts (one for each dataset in GLUE).
The router architecture is simply a linear classifier, i.e.\ a linear projection layer consisting of a weight matrix of shape $d$ x $N$, where $d$ is the model dimension and $N$ is the number of experts in the MoE layer, followed by a softmax nonlinearity.
To help avoid saturating the softmax nonlinearity, we apply layer normalization both to the input of the router as well as the rows of the linear layer.
In the encoder, the router takes as input the preceding hidden states, which are averaged across the sequence and fed into the router. 
In the decoder, the routers receive the average of the encoder's final hidden states instead of the decoder hidden states to prevent information leakage from later target tokens. 
We also include expert dropout \citep{liu2022gating} where each expert is dropped with a probability of 0.1 wherever it was found to be beneficial (a detailed ablation can be found in \cref{tab:dropout_ablation}).
In GLUE, dataset sizes vary by three orders of magnitude, and we therefore found that load-balancing losses (as used e.g.\ in \citep{shazeer2017outrageously,fedus2021switch,lepikhin2020gshard} to encourage uniform usage across experts) tended to hurt performance, so we did not include them.

\paragraph{ResNet-DomainNet}

In this setting, we focus on adapting an ImageNet pre-trained ResNet18 model \citep{he2016deep} to datasets within DomainNet \citep{peng2019moment}.
DomainNet is a collection of object recognition datasets that cover six distinct domains and all share the same label space corresponding to 345 object categories.
We treat the domain of each example as its tag.
As in the T5-GLUE setting, we freeze the pretrained model and insert eight expert routing blocks after each of the eight residual layer groups in the model.
Each block includes six experts corresponding to the number of domains.
We use the same architecture for routers as in T5-GLUE and feed average-pooled hidden states into the router to compute the routing probability.
Experts in this setting use batch normalization on their input instead of layer normalization in the output, following \cite{rebuffi2017learning}. 
As in T5-GLUE, we omit load-balancing losses due to dramatically different sizes across domains in DomainNet.

We only provide the average performance across datasets/domains in the main paper due to space limitations.
Full results are provided in \cref{sec:full_results}.

\begin{figure*}[t]
\centering
\includegraphics[width=0.9\textwidth]{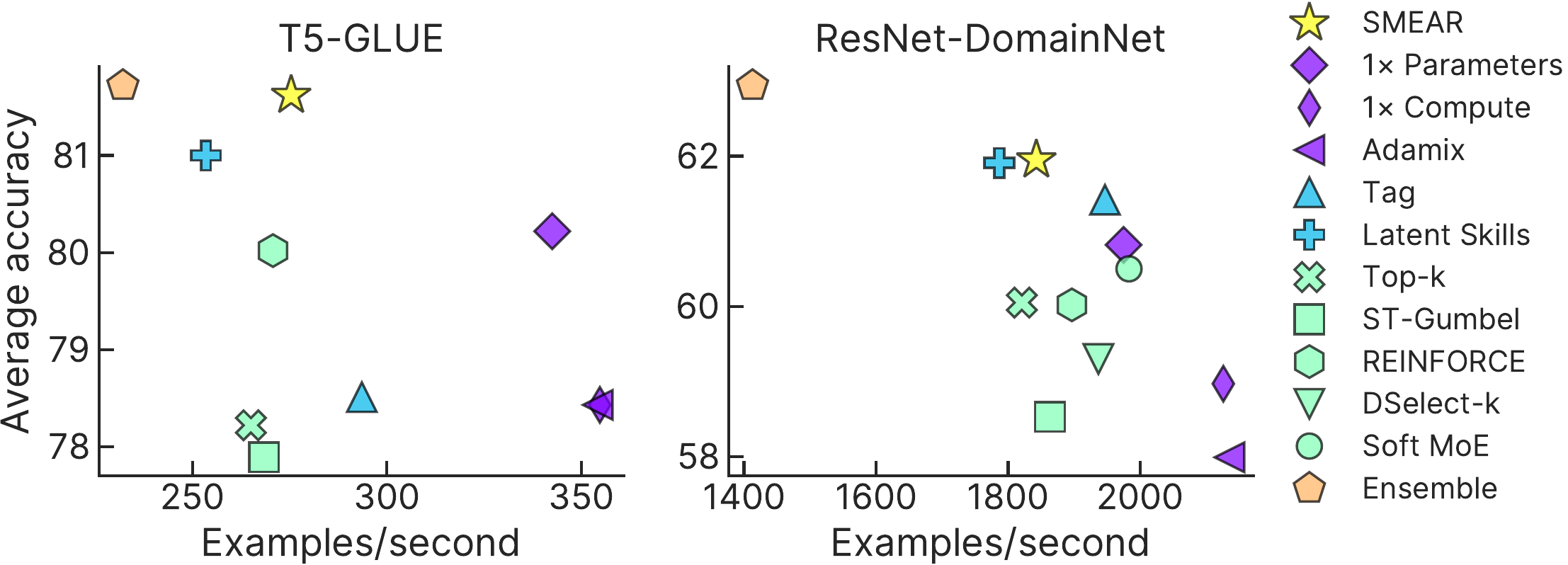}
\caption{Average accuracy and inference speed (in examples processed per second) for models using different routing approaches on our T5-GLUE and ResNet-DomainNet settings. Routing approaches are grouped by color; groups are (in order of the legend) our method (SMEAR), methods that do not use adaptive routing ($1\times$ compute, $1\times$ parameters, Adamix, and Hash), methods that make use of metadata (Tag and Latent Skills), methods that learn adaptive routing (Top-$k$, ST-Gumbel, REINFORCE, DSelect-$k$, and Soft MoE), and methods that ensemble expert outputs (Ensemble). We omit Hash routing from the plots because its poor performance (66.9\% on T5-GLUE and 52.4\% on ResNet-DomainNet) hampers readability. Exact numerical results for all methods and standard deviation across five runs are provided in \cref{sec:full_results}.}
\label{fig:results}
\end{figure*}

\subsection{Results}
\label{sec:results}

To assess the overall effectiveness of routing strategies learned with SMEAR, we compare to learned routing using the gradient estimators, heuristic routing strategies, and modular baselines from \cref{sec:background}.
A summary of our results is shown in \cref{fig:results}.
First, we find that models using routing strategies learned through gradient estimation often underperform heuristic routing strategies -- while the best-performing estimator (REINFORCE) in T5-GLUE outperforms tag routing, all estimators perform worse than tag routing in ResNet-DomainNet.
On the other hand, we observed some cases where gradient estimation-based routing outperforms hash or single-expert routing, which suggests that the learned routing strategies were nontrivial.
Pertinently, in all experimental settings, SMEAR matches or outperforms every other routing strategy, including both routing learned by gradient estimators and all heuristic routing strategies.
In particular, SMEAR achieves 2.7\% improvement over tag routing in T5-GLUE and 0.6\% improvement over tag routing in ResNet-DomainNet, suggesting effective specialization and sharing of experts.
SMEAR additionally outperforms the single-expert parameter-matched baseline ($1\times$ parameters) by 1.4\% in T5-GLUE and 1.2\% in ResNet-DomainNet, further highlighting the importance of modularity. 

\begin{figure*}[t]
\centering
\subfloat[T5-GLUE model]{
    \includegraphics[width=0.7\textwidth]{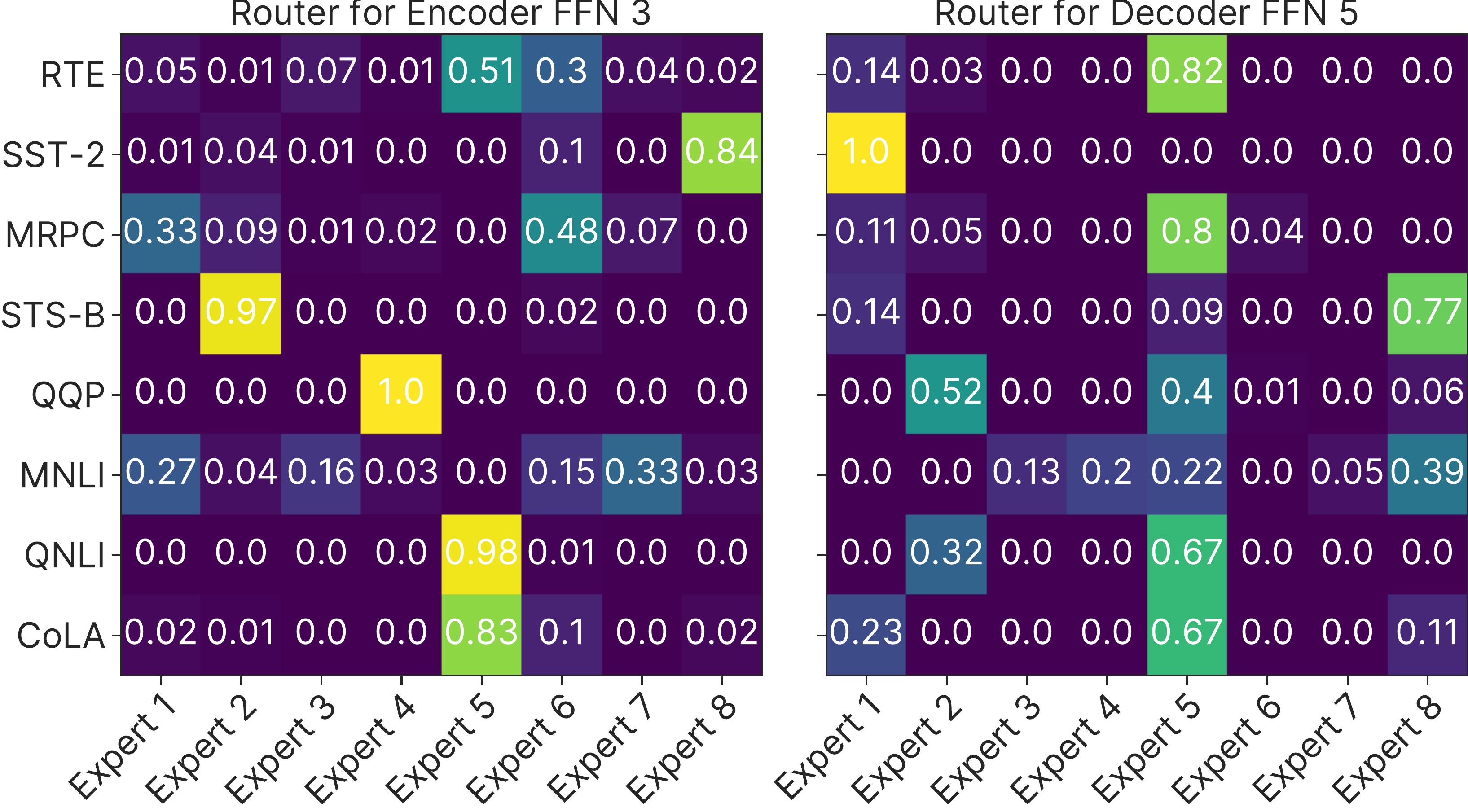}
    \label{fig:glue_sample_routing}
}

\subfloat[ResNet-DomainNet model]{
    \includegraphics[width=0.7\textwidth]{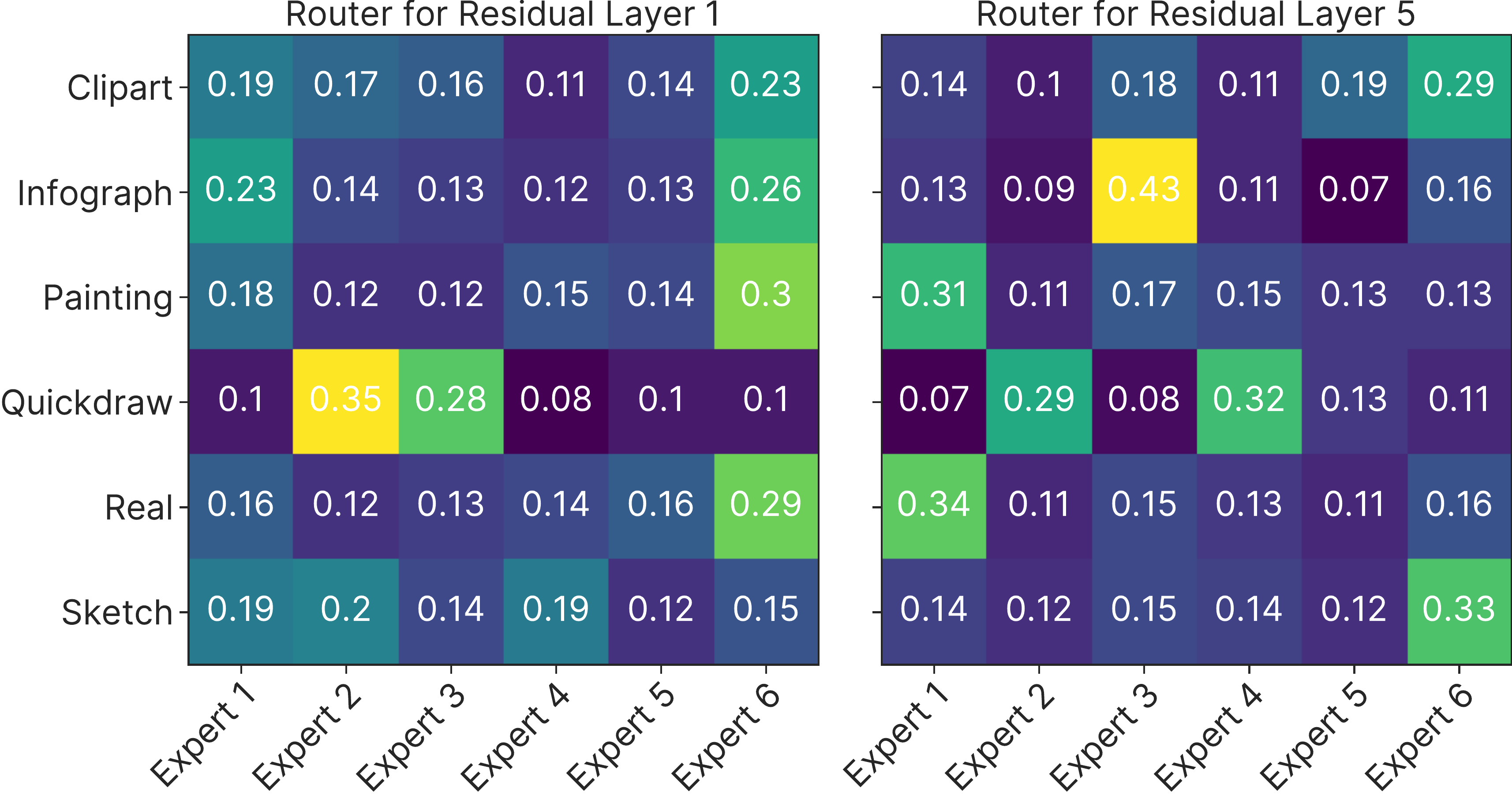}
    \label{fig:domainnet_sample_routing}
}
\caption{Average routing distributions produced by SMEAR for two routers from the T5-GLUE model and two from the ResNet-DomainNet model. For a given router, we average all routing distributions across all examples from a given dataset.}
\label{fig:sample_routing_distributions}
\end{figure*}

As an upper bound on performance, we also compare SMEAR to expert ensembling (``Ensemble'') which averages the outputs of all experts and incurs significantly higher computational cost. 
SMEAR matches the performance of ensemble routing in T5-GLUE and modestly underperforms it in ResNet-DomainNet, despite being significantly computationally cheaper.
Compared to Adamix, which similarly averages experts but does not learn a routing strategy, SMEAR achieves 3.2\% higher performance in T5-GLUE and 4\% higher in ResNet-DomainNet.
Since the Soft MoE method averages input tokens, it's inapplicable for the encoder-decoder model in T5-GLUE, where future tokens are not available for averaging in the decoder during inference. 
Hence, we include Soft MoE for ResNet-DomainNet where SMEAR outperforms it by 1.5\%.
Additionally, SMEAR exceeds the DSelect-$k$ method by 2.7\% in ResNet-DomainNet. 
However, despite extensive hyperparameter tuning we encountered training instabilites with the DSelect-$k$ method in T5-GLUE and therefore omit those results.
Moreover, while the performance improvement of SMEAR over Latent Skills is relatively small (0.6\% in T5-GLUE and 0.1\% in ResNet-DomainNet), a major advantage of SMEAR over Latent Skills is that it does not assume access to oracle tags (which are not always available in real-world settings) and instead learns an adaptive routing strategy. 
Finally, we highlight the consistency of improvements achieved by SMEAR across a diverse range of datasets and architectures, confirming its generality and robustness.

We additionally plot the inference speed (in terms of number of examples processed per second) of each method in \cref{fig:results}.
The single-expert, Adamix, Hash, and Tag routing methods are the fastest since they do not use any routing networks.
Despite the slight overhead of averaging the weights in SMEAR, we observe that its inference speed is almost identical to that of discrete adaptive routing (as learned via gradient estimation techniques).
This confirms that the performance gains attained by SMEAR do not incur significant additional costs.
Ensembling expert outputs is the slowest, with a $1.2 \times$ slowdown in T5-GLUE and $1.3 \times$ slowdown in ResNet-DomainNet compared to SMEAR.

\paragraph{Scaling}

Thus far, we have always set the number of experts equal to the number of tasks (in T5-GLUE) or domains (in DomainNet).
However, with learned routing there is no reason to force this constraint, so we therefore tested the scalability of SMEAR by evaluating its performance with twice as many experts (16 for T5-GLUE and 12 for ResNet-DomainNet).
We found a significant improvement (0.8\%) when doubling the number of experts on ResNet-DomainNet, but no significant change on T5-GLUE ($81.3 \pm 1.1$ vs. $81.6 \pm 1.1$).
This suggests there is no benefit to increasing capacity in the T5-GLUE setting.
The complete results for doubling the number of experts are presented in \cref{sec:full_results} (labeled ``SMEAR $ 2 \times$''). 

\subsection{Qualitative Analysis}
\label{sec:qualitative}

In this section, we provide qualitative analysis of the routing learned by SMEAR by visualizing the average router distribution across all examples in a given dataset for every router in each model.
\Cref{fig:sample_routing_distributions} shows four select visualizations (two from a SMEAR-based model trained in T5-GLUE and two from ResNet-DomainNet).
Across the two T5-GLUE router distributions shown in \cref{fig:sample_routing_distributions}, we observe significantly different behavior -- one mainly follows a tag routing-style strategy whereas the other routes most datasets to the same expert.
However, we note that the tag-style router utilizes shared experts for RTE, MRPC, and MNLI; notably, these tasks are somewhat similar in that they all involve determining similarity among pairs of sentences.
In the single-expert-style router, STS-B (the only regression task) and SST-2 (which has a distinct output vocabulary) are given dedicated experts, and MNLI (a large and relatively challenging dataset) is routed through many different experts.
More broadly, we highlight that there is generally a great deal of sparsity in the learned routing distributions, suggesting a significant amount of expert specialization.
In ResNet-DomainNet, we can see that examples from the Quickdraw domain are routed to two specific experts in both cases.
Additionally, we observe that the router distribution of the Painting and Real domains are highly correlated. 
Other domains such as Clipart and Sketch seem to evenly use experts.
Interestingly, there is less expert specialization in the ResNet-DomainNet model, suggesting that there may be more similarities between the individual domains in DomainNet compared to the tasks in GLUE.

In general, other approaches for learning routing did not exhibit as intuitive or meaningful specialization and sharing.
In ResNet-DomainNet, Top-$k$ demonstrates uniform routing in the initial layers but chooses a single expert in the last layer.  REINFORCE, ST-Gumbel, and DSelect-$k$ tend to exhibit mostly degenerate single-expert routing. 
Interestingly, all these gradient estimators learn to assign a distinct expert for the Quickdraw dataset. 
However, this degree of specialization is insufficient for achieving superior performance scores. 
In T5-GLUE, these estimators display degenerate routing in some layers, while showing a tendency to share a few experts (approximately 3 out of 8) in other layers across tasks. 
Methods such as Latent Skills and Ensemble utilize most experts in the MoE layer (similar to SMEAR).
Routing distribution visualizations for all methods and all layers can be found in \cref{sec:router_dist_all_layers}.

\section{Related Work}
\label{sec:related}

\paragraph{Models with Conditional Computation}

Given the difficulties of training models with discrete routing strategies, there is a large body of literature investigating effective methods for training such models.
\citet{deecke2020latent,hazimeh2021dselect,dua2021tricks} start training with most of the experts activated and gradually introduce sparsity.
\citet{kudugunta2021beyond, ponti2022combining, ma2019snr, Gupta2022SparselyAM} group examples from the same task together and introduce task-specific parameters in the router.
\citet{zuo2021taming, wang2022adamix} explore random routing with consistency regularization as a form of model ensembling. 
Other works avoid learned routing by hand-crafting heuristic routing strategies.
\citet{gururangan2021demix} built sparsely activated language models where different domains use separate experts and then weights the experts for new domains.
\citet{Tang2022OneMM, Pfeiffer2022LiftingTC, Pfeiffer2020MADXAA} assign experts based on task-related human knowledge.
Our focus on settings where performant routing schemes can be hand-designed takes inspiration from this line of work.

Because sparsely activated models disentangle computation and parameter count, significant effort has gone into leveraging conditional computation to create massive pre-trained models with a feasible computation cost \citep{fedus2022review, shazeer2017outrageously, fedus2021switch,du2022glam,zoph2022designing,yu2022efficient}.
Many works explore different routing methods in this setting, with a major focus on balancing the load across experts \citep{lewis2021base, zhou2022mixture, kool2021unbiased, roller2021hash}.
\citet{li2022sparse} demonstrate that models structured as sparse mixture-of-experts generalize effectively to novel domains, as compared to other domain generalization algorithms in vision transformers. 
Another line of work aims to introduce ways to convert trained dense models into similar-sized sparse models with a lower computational footprint \citep{lee2022sparse, zhang2022moefication, komatsuzaki2022sparse}.
More generally, there are other forms of conditional computation beyond the formulation we describe in this work \citep{han2021dynamic}.
Modular networks \citep{kirsch2018modular, jiang2019self, hu2017learning} use a router to assemble heterogeneous subnetworks into a network layout specific to a given input example.
Early exiting \citep{graves2016adaptive, xin2020deebert, liu2020fastbert, bengio2015conditional} saves computation by terminating inference in earlier layers or recurrent steps.

\paragraph{Gradient Estimation Techniques}

Many gradient estimators have been proposed to produce approximate gradients for backpropagating through discrete steps. 
\citet{clark2022unified} uses a learned baseline to reduce the variance of the REINFORCE estimator. 
The REBAR estimator \citep{tucker2017rebar} adds a reparameterizable term to REINFORCE as a baseline that results in a more effective unbiased estimator. 
This additional term incorporates a relaxed sample similar to Gumbel-Softmax \citep{jang2016categorical}. 
RELAX \citep{grathwohl2017backpropagation} is similar to REBAR but uses a learnable neural network for the reparameterizable term. 
\citet{kool2019buy} uses additional samples as a built-in baseline for REINFORCE. 
\citet{yin2018arm} and \citet{dong2020disarm} use the idea of coupling between multiple samples to reduce the variance of the gradient estimator for improved training of models with binary latent variables.
\citet{dong2021coupled} improve upon \citet{yin2018arm} and \citet{dong2020disarm} by extending the estimator to categorical variables.
These studies have theoretically analyzed gradient estimators, focusing on the bias and variance of these gradients and suggesting enhancements through improved relaxation techniques and variance reduction techniques. 
The fundamental theoretical advantage of our method lies in its ability to enable exact gradient computation through standard backpropagation.
In preliminary experiments, we did not find any major gains from using more sophisticated gradient estimation techniques, but designing gradient estimators with discrete routing in mind could yield more performant routing strategies.

\paragraph{Issues with Conditional Computation}

A great deal of past work has highlighted issues with models that use conditional computation.
\citet{clark2022unified} study the scaling laws of sparse language models and discovered a computational cutoff above which no additional benefits are observed.
Relatedly, \citet{du2022glam} observe worse results when further scaling up the number of experts.
\citet{chi2022representation} highlight that using the model's activations as input to the router can cause the representations to ``collapse''.
\citet{dai2022stablemoe} demonstrate that learned routing decisions can fluctuate significantly over training.
\citet{mittal2022modular} create a set of simple and compositional data distributions and show that systems with modular architecture can not find the most performant solution when trained end-to-end.
\citet{ye2022eliciting} experiment with various designs for multi-task learning with task-level routing and find that the performance never surpasses simple multi-task baselines.
We show a possibility to avoid these issues with a fully differentiable routing strategy that does not increase computational costs.

\paragraph{Weight Averaging Methods}
Many prior works utilize parameter averaging for ensembling.
\citet{wortsman2022robust,ilharco2022patching} average the weights of a pre-trained and a fine-tuned model to improve performance on target tasks as well as robustness to distribution shift.
\citet{choshen2022fusing} similarly show that merging multiple models fine-tuned on different datasets can provide a better initialization than using the original pre-trained model for further fine-tuning on new unseen datasets.
The phenomenon of linear mode connectivity, as observed by \citet{frankle2020linear}, \citet{qin2022exploring} and \citet{entezari2021role}, refers to the tendency of models to converge to a single low-loss basin, which makes it possible to average the weights of models without increasing loss.
\citet{juneja2022linear} found that tasks with different generalization strategies have high barriers when linearly interpolated, whereas tasks with similar generalization are linearly mode connected and can be averaged.
They defined several generalization strategies and found that sets of tasks form clusters with those strategies.
In our case, the router plays the role of identifying experts with similar generalization properties when conditioned on the input and performs the averaging.

\citet{yang2019condconv, zhang2021basisnet} compute convolution kernels by averaging weights of individual kernels. 
\citet{maziarz2019flexible} learn a task-specific binary allocation matrix, where each allocation corresponds to a convolution kernel. 
Since the convolution operation is linear, weight averaging and ensembling are mathematically equivalent in the aforementioned works. 
However, SMEAR performs averaging on non-linear and parameter efficient experts that, when trained alone, can match the performance of the fully fine-tuned model \citep{houlsby2019parameter}.  
$\pi$-Tuning \citep{wu2023pi} employs a set of existing task specific experts, retrieving the top $k$ experts for a downstream task and learns to interpolate among these experts for the downstream task. 
While $\pi$-Tuning enables transfer learning to a new downstream task by learning to interpolate, our focus is on developing a routing algorithm that learns how to share or specialize experts without using any metadata.
\citet{wortsman2022model} average multiple models fine-tuned from large pre-trained models with different hyperparameters, using techniques such as uniform averaging and greedy averaging by picking those models that increase the average performance using a validation set.
\citet{gururangan2021demix, Li2022BranchTrainMergeEP} use data in novel domains to estimate weights to average each of their domain-specific subnetworks.
\citet{izmailov2018averaging} average model parameters along a training trajectory and show that the resultant model generalizes better than the solution found by SGD at the end of training.
Our work instead learns a router that implements an adaptive routing strategy over the course of the training process.

Concurrent work by \citet{zadouri2023pushing} used SMEAR for training a mixture-of-experts-style model with LoRA \citep{hu2021lora}- and ${IA}^3$ \citep{liu2022few}-based experts.
They show that the resultant model outperforms full model finetuning in generalizing to unseen tasks. 
Model averaging is also a common step in distributed optimization, where it is widely used in federated learning \citep{mcmahan2017communication} and has recently been used for distributed fine-tuning \citep{wortsman2022fi}, multi-domain training \citep{Li2022BranchTrainMergeEP}, and multitask training \citep{don2022cold}.
There are also works that utilize different styles of merging instead of weight averaging of parameters, such as reweighting parameters in accordance with their approximate Fisher information \citep{matena2021merging}, aligning features by fitting a linear projection \citep{jin2022dataless}, and permuting columns to account for permutation symmetries \citep{ainsworth2022git}.
We are interested in applying these more sophisticated merging methods to SMEAR in future work.

\section{Conclusion}
\label{sec:conclusion}

In this work, we sought to address shortcomings of models with discrete routing among experts that can lead them to underperform heuristic non-learned routing.
We hypothesized that these issues stem from the gradient estimation techniques required to propagate gradients through discrete routing decisions and therefore focused on designing an expert routing architecture that allows exact calculation of gradients.
Our approach, called SMEAR, works by computing a weighted average of expert parameters where the weighting is set according to the output of a learned router.
We compared the performance of models using SMEAR to discrete routing models that were trained via various gradient estimation techniques.
In experimental settings covering different modalities and model architectures, we found that SMEAR outperformed all models with discrete routing as well as performant heursitic routing strategies.
Notably, this performance boost comes with no increase in computational costs.
SMEAR also matched or outperformed existing state-of-the-art methods for learning modular models through expert averaging while removing the requirement for oracle task labels.
Through qualitative analysis, we further confirmed that the experts learned in a model using SMEAR specialize to different types of inputs and that the router learns a nontrivial strategy that exploits commonalities across different examples.
In future work, we are interested in exploring different expert architectures \citep{liu2022few,hu2021lora} and improved merging methods \citep{matena2021merging,ainsworth2022git,jin2022dataless}.
Given access to a larger amount of compute, we would also be excited to try out SMEAR in the large-scale settings where discrete routing has been used \citep{fedus2021switch,zoph2022designing,du2022glam} to see whether it helps fix the poor scaling properties of models with discrete routing \citep{clark2022unified}.

\bibliographystyle{tmlr}
\bibliography{general.bib}

\begin{thebibliography}{111}
\providecommand{\natexlab}[1]{#1}
\providecommand{\url}[1]{\texttt{#1}}
\expandafter\ifx\csname urlstyle\endcsname\relax
  \providecommand{\doi}[1]{doi: #1}\else
  \providecommand{\doi}{doi: \begingroup \urlstyle{rm}\Url}\fi

\bibitem[Ainsworth et~al.(2022)Ainsworth, Hayase, and
  Srinivasa]{ainsworth2022git}
Samuel~K Ainsworth, Jonathan Hayase, and Siddhartha Srinivasa.
\newblock Git re-basin: Merging models modulo permutation symmetries.
\newblock \emph{arXiv preprint arXiv:2209.04836}, 2022.

\bibitem[Bach et~al.(2022)Bach, Sanh, Yong, Webson, Raffel, Nayak, Sharma, Kim,
  Bari, Fevry, et~al.]{bach2022promptsource}
Stephen~H Bach, Victor Sanh, Zheng-Xin Yong, Albert Webson, Colin Raffel,
  Nihal~V Nayak, Abheesht Sharma, Taewoon Kim, M~Saiful Bari, Thibault Fevry,
  et~al.
\newblock Promptsource: An integrated development environment and repository
  for natural language prompts.
\newblock \emph{arXiv preprint arXiv:2202.01279}, 2022.

\bibitem[Bao et~al.(2021)Bao, Dong, and Wei]{bao2021beit}
Hangbo Bao, Li~Dong, and Furu Wei.
\newblock Beit: Bert pre-training of image transformers.
\newblock \emph{arXiv preprint arXiv:2106.08254}, 2021.

\bibitem[Bau et~al.(2020)Bau, Zhu, Strobelt, Lapedriza, Zhou, and
  Torralba]{bau2020understanding}
David Bau, Jun-Yan Zhu, Hendrik Strobelt, Agata Lapedriza, Bolei Zhou, and
  Antonio Torralba.
\newblock Understanding the role of individual units in a deep neural network.
\newblock \emph{Proceedings of the National Academy of Sciences}, 117\penalty0
  (48):\penalty0 30071--30078, 2020.

\bibitem[Bengio et~al.(2015)Bengio, Bacon, Pineau, and
  Precup]{bengio2015conditional}
Emmanuel Bengio, Pierre-Luc Bacon, Joelle Pineau, and Doina Precup.
\newblock Conditional computation in neural networks for faster models.
\newblock \emph{arXiv preprint arXiv:1511.06297}, 2015.

\bibitem[Bengio et~al.(2013)Bengio, L{\'e}onard, and
  Courville]{bengio2013estimating}
Yoshua Bengio, Nicholas L{\'e}onard, and Aaron Courville.
\newblock Estimating or propagating gradients through stochastic neurons for
  conditional computation.
\newblock \emph{arXiv preprint arXiv:1308.3432}, 2013.

\bibitem[Bentivogli et~al.(2009)Bentivogli, Clark, Dagan, and
  Giampiccolo]{bentivogli2009fifth}
Luisa Bentivogli, Peter Clark, Ido Dagan, and Danilo Giampiccolo.
\newblock The fifth pascal recognizing textual entailment challenge.
\newblock In \emph{TAC}, 2009.

\bibitem[Cer et~al.(2017)Cer, Diab, Agirre, Lopez-Gazpio, and
  Specia]{cer2017semeval}
Daniel Cer, Mona Diab, Eneko Agirre, Inigo Lopez-Gazpio, and Lucia Specia.
\newblock Semeval-2017 task 1: Semantic textual similarity-multilingual and
  cross-lingual focused evaluation.
\newblock \emph{arXiv preprint arXiv:1708.00055}, 2017.

\bibitem[Chi et~al.(2022)Chi, Dong, Huang, Dai, Ma, Patra, Singhal, Bajaj,
  Song, and Wei]{chi2022representation}
Zewen Chi, Li~Dong, Shaohan Huang, Damai Dai, Shuming Ma, Barun Patra, Saksham
  Singhal, Payal Bajaj, Xia Song, and Furu Wei.
\newblock On the representation collapse of sparse mixture of experts.
\newblock \emph{arXiv preprint arXiv:2204.09179}, 2022.

\bibitem[Choshen et~al.(2022{\natexlab{a}})Choshen, Venezian, Don-Yehia,
  Slonim, and Katz]{choshen2022start}
Leshem Choshen, Elad Venezian, Shachar Don-Yehia, Noam Slonim, and Yoav Katz.
\newblock Where to start? analyzing the potential value of intermediate models.
\newblock \emph{arXiv preprint arXiv:2211.00107}, 2022{\natexlab{a}}.

\bibitem[Choshen et~al.(2022{\natexlab{b}})Choshen, Venezian, Slonim, and
  Katz]{choshen2022fusing}
Leshem Choshen, Elad Venezian, Noam Slonim, and Yoav Katz.
\newblock Fusing finetuned models for better pretraining.
\newblock \emph{arXiv preprint arXiv:2204.03044}, 2022{\natexlab{b}}.

\bibitem[Clark et~al.(2022)Clark, de~Las~Casas, Guy, Mensch, Paganini,
  Hoffmann, Damoc, Hechtman, Cai, Borgeaud, et~al.]{clark2022unified}
Aidan Clark, Diego de~Las~Casas, Aurelia Guy, Arthur Mensch, Michela Paganini,
  Jordan Hoffmann, Bogdan Damoc, Blake Hechtman, Trevor Cai, Sebastian
  Borgeaud, et~al.
\newblock Unified scaling laws for routed language models.
\newblock In \emph{International Conference on Machine Learning}, pp.\
  4057--4086. PMLR, 2022.

\bibitem[Csord{\'a}s et~al.(2021)Csord{\'a}s, van Steenkiste, and
  Schmidhuber]{csordas2021neural}
R{\'o}bert Csord{\'a}s, Sjoerd van Steenkiste, and J{\"u}rgen Schmidhuber.
\newblock Are neural nets modular? inspecting functional modularity through
  differentiable weight masks.
\newblock In \emph{International Conference on Learning Representations}, 2021.

\bibitem[Dai et~al.(2022)Dai, Dong, Ma, Zheng, Sui, Chang, and
  Wei]{dai2022stablemoe}
Damai Dai, Li~Dong, Shuming Ma, Bo~Zheng, Zhifang Sui, Baobao Chang, and Furu
  Wei.
\newblock Stablemoe: Stable routing strategy for mixture of experts.
\newblock \emph{arXiv preprint arXiv:2204.08396}, 2022.

\bibitem[De~Cao et~al.(2021)De~Cao, Aziz, and Titov]{de2021editing}
Nicola De~Cao, Wilker Aziz, and Ivan Titov.
\newblock Editing factual knowledge in language models.
\newblock In \emph{Proceedings of the 2021 Conference on Empirical Methods in
  Natural Language Processing}, 2021.

\bibitem[Deecke et~al.(2020)Deecke, Hospedales, and Bilen]{deecke2020latent}
Lucas Deecke, Timothy Hospedales, and Hakan Bilen.
\newblock Latent domain learning with dynamic residual adapters.
\newblock \emph{arXiv preprint arXiv:2006.00996}, 2020.

\bibitem[Devlin et~al.(2018)Devlin, Chang, Lee, and Toutanova]{devlin2018bert}
Jacob Devlin, Ming-Wei Chang, Kenton Lee, and Kristina Toutanova.
\newblock Bert: Pre-training of deep bidirectional transformers for language
  understanding.
\newblock \emph{arXiv preprint arXiv:1810.04805}, 2018.

\bibitem[Dolan \& Brockett(2005)Dolan and Brockett]{dolan2005automatically}
Bill Dolan and Chris Brockett.
\newblock Automatically constructing a corpus of sentential paraphrases.
\newblock In \emph{Third International Workshop on Paraphrasing (IWP2005)},
  2005.

\bibitem[Don-Yehiya et~al.(2022)Don-Yehiya, Venezian, Raffel, Slonim, Katz, and
  Choshen]{don2022cold}
Shachar Don-Yehiya, Elad Venezian, Colin Raffel, Noam Slonim, Yoav Katz, and
  Leshem Choshen.
\newblock Cold fusion: Collaborative descent for distributed multitask
  finetuning.
\newblock \emph{arXiv preprint arXiv:2212.01378}, 2022.

\bibitem[Dong et~al.(2020)Dong, Mnih, and Tucker]{dong2020disarm}
Zhe Dong, Andriy Mnih, and George Tucker.
\newblock Disarm: An antithetic gradient estimator for binary latent variables.
\newblock \emph{Advances in neural information processing systems},
  33:\penalty0 18637--18647, 2020.

\bibitem[Dong et~al.(2021)Dong, Mnih, and Tucker]{dong2021coupled}
Zhe Dong, Andriy Mnih, and George Tucker.
\newblock Coupled gradient estimators for discrete latent variables.
\newblock \emph{Advances in Neural Information Processing Systems},
  34:\penalty0 24498--24508, 2021.

\bibitem[Du et~al.(2022)Du, Huang, Dai, Tong, Lepikhin, Xu, Krikun, Zhou, Yu,
  Firat, et~al.]{du2022glam}
Nan Du, Yanping Huang, Andrew~M Dai, Simon Tong, Dmitry Lepikhin, Yuanzhong Xu,
  Maxim Krikun, Yanqi Zhou, Adams~Wei Yu, Orhan Firat, et~al.
\newblock Glam: Efficient scaling of language models with mixture-of-experts.
\newblock In \emph{International Conference on Machine Learning}, pp.\
  5547--5569. PMLR, 2022.

\bibitem[Dua et~al.(2021)Dua, Bhosale, Goswami, Cross, Lewis, and
  Fan]{dua2021tricks}
Dheeru Dua, Shruti Bhosale, Vedanuj Goswami, James Cross, Mike Lewis, and
  Angela Fan.
\newblock Tricks for training sparse translation models.
\newblock \emph{arXiv preprint arXiv:2110.08246}, 2021.

\bibitem[Entezari et~al.(2021)Entezari, Sedghi, Saukh, and
  Neyshabur]{entezari2021role}
Rahim Entezari, Hanie Sedghi, Olga Saukh, and Behnam Neyshabur.
\newblock The role of permutation invariance in linear mode connectivity of
  neural networks.
\newblock \emph{arXiv preprint arXiv:2110.06296}, 2021.

\bibitem[Fedus et~al.(2021)Fedus, Zoph, and Shazeer]{fedus2021switch}
William Fedus, Barret Zoph, and Noam Shazeer.
\newblock Switch transformers: Scaling to trillion parameter models with simple
  and efficient sparsity, 2021.

\bibitem[Fedus et~al.(2022)Fedus, Dean, and Zoph]{fedus2022review}
William Fedus, Jeff Dean, and Barret Zoph.
\newblock A review of sparse expert models in deep learning.
\newblock \emph{arXiv preprint arXiv:2209.01667}, 2022.

\bibitem[Frankle et~al.(2020)Frankle, Dziugaite, Roy, and
  Carbin]{frankle2020linear}
Jonathan Frankle, Gintare~Karolina Dziugaite, Daniel Roy, and Michael Carbin.
\newblock Linear mode connectivity and the lottery ticket hypothesis.
\newblock In \emph{International Conference on Machine Learning}, pp.\
  3259--3269. PMLR, 2020.

\bibitem[Grathwohl et~al.(2017)Grathwohl, Choi, Wu, Roeder, and
  Duvenaud]{grathwohl2017backpropagation}
Will Grathwohl, Dami Choi, Yuhuai Wu, Geoffrey Roeder, and David Duvenaud.
\newblock Backpropagation through the void: Optimizing control variates for
  black-box gradient estimation.
\newblock \emph{arXiv preprint arXiv:1711.00123}, 2017.

\bibitem[Graves(2016)]{graves2016adaptive}
Alex Graves.
\newblock Adaptive computation time for recurrent neural networks.
\newblock \emph{arXiv preprint arXiv:1603.08983}, 2016.

\bibitem[Gupta et~al.(2022)Gupta, Mukherjee, Subudhi, Gonzalez, Jose,
  Awadallah, and Gao]{Gupta2022SparselyAM}
Shashank Gupta, Subhabrata Mukherjee, Krishan Subudhi, Eduardo Gonzalez, Damien
  Jose, Ahmed~Hassan Awadallah, and Jianfeng Gao.
\newblock Sparsely activated mixture-of-experts are robust multi-task learners.
\newblock \emph{ArXiv}, abs/2204.07689, 2022.

\bibitem[Gururangan et~al.(2021)Gururangan, Lewis, Holtzman, Smith, and
  Zettlemoyer]{gururangan2021demix}
Suchin Gururangan, Mike Lewis, Ari Holtzman, Noah~A Smith, and Luke
  Zettlemoyer.
\newblock Demix layers: Disentangling domains for modular language modeling.
\newblock \emph{arXiv preprint arXiv:2108.05036}, 2021.

\bibitem[Han et~al.(2021)Han, Huang, Song, Yang, Wang, and
  Wang]{han2021dynamic}
Yizeng Han, Gao Huang, Shiji Song, Le~Yang, Honghui Wang, and Yulin Wang.
\newblock Dynamic neural networks: A survey.
\newblock \emph{IEEE Transactions on Pattern Analysis and Machine
  Intelligence}, 2021.

\bibitem[Hazimeh et~al.(2021)Hazimeh, Zhao, Chowdhery, Sathiamoorthy, Chen,
  Mazumder, Hong, and Chi]{hazimeh2021dselect}
Hussein Hazimeh, Zhe Zhao, Aakanksha Chowdhery, Maheswaran Sathiamoorthy, Yihua
  Chen, Rahul Mazumder, Lichan Hong, and Ed~Chi.
\newblock Dselect-k: Differentiable selection in the mixture of experts with
  applications to multi-task learning.
\newblock \emph{Advances in Neural Information Processing Systems},
  34:\penalty0 29335--29347, 2021.

\bibitem[He et~al.(2016)He, Zhang, Ren, and Sun]{he2016deep}
Kaiming He, Xiangyu Zhang, Shaoqing Ren, and Jian Sun.
\newblock Deep residual learning for image recognition.
\newblock In \emph{Proceedings of the IEEE conference on computer vision and
  pattern recognition}, pp.\  770--778, 2016.

\bibitem[Houlsby et~al.(2019)Houlsby, Giurgiu, Jastrzebski, Morrone,
  De~Laroussilhe, Gesmundo, Attariyan, and Gelly]{houlsby2019parameter}
Neil Houlsby, Andrei Giurgiu, Stanislaw Jastrzebski, Bruna Morrone, Quentin
  De~Laroussilhe, Andrea Gesmundo, Mona Attariyan, and Sylvain Gelly.
\newblock Parameter-efficient transfer learning for nlp.
\newblock In \emph{International Conference on Machine Learning}, pp.\
  2790--2799. PMLR, 2019.

\bibitem[Hu et~al.(2021)Hu, Shen, Wallis, Allen-Zhu, Li, Wang, Wang, and
  Chen]{hu2021lora}
Edward~J Hu, Yelong Shen, Phillip Wallis, Zeyuan Allen-Zhu, Yuanzhi Li, Shean
  Wang, Lu~Wang, and Weizhu Chen.
\newblock Lora: Low-rank adaptation of large language models.
\newblock \emph{arXiv preprint arXiv:2106.09685}, 2021.

\bibitem[Hu et~al.(2017)Hu, Andreas, Rohrbach, Darrell, and
  Saenko]{hu2017learning}
Ronghang Hu, Jacob Andreas, Marcus Rohrbach, Trevor Darrell, and Kate Saenko.
\newblock Learning to reason: End-to-end module networks for visual question
  answering.
\newblock In \emph{Proceedings of the IEEE international conference on computer
  vision}, pp.\  804--813, 2017.

\bibitem[Ilharco et~al.(2022)Ilharco, Wortsman, Gadre, Song, Hajishirzi,
  Kornblith, Farhadi, and Schmidt]{ilharco2022patching}
Gabriel Ilharco, Mitchell Wortsman, Samir~Yitzhak Gadre, Shuran Song, Hannaneh
  Hajishirzi, Simon Kornblith, Ali Farhadi, and Ludwig Schmidt.
\newblock Patching open-vocabulary models by interpolating weights.
\newblock \emph{arXiv preprint arXiv:2208.05592}, 2022.

\bibitem[Izmailov et~al.(2018)Izmailov, Podoprikhin, Garipov, Vetrov, and
  Wilson]{izmailov2018averaging}
Pavel Izmailov, Dmitrii Podoprikhin, Timur Garipov, Dmitry Vetrov, and
  Andrew~Gordon Wilson.
\newblock Averaging weights leads to wider optima and better generalization.
\newblock \emph{arXiv preprint arXiv:1803.05407}, 2018.

\bibitem[Jang et~al.(2016)Jang, Gu, and Poole]{jang2016categorical}
Eric Jang, Shixiang Gu, and Ben Poole.
\newblock Categorical reparameterization with gumbel-softmax.
\newblock \emph{arXiv preprint arXiv:1611.01144}, 2016.

\bibitem[Jiang \& Bansal(2019)Jiang and Bansal]{jiang2019self}
Yichen Jiang and Mohit Bansal.
\newblock Self-assembling modular networks for interpretable multi-hop
  reasoning.
\newblock \emph{arXiv preprint arXiv:1909.05803}, 2019.

\bibitem[Jin et~al.(2022)Jin, Ren, Preotiuc-Pietro, and Cheng]{jin2022dataless}
Xisen Jin, Xiang Ren, Daniel Preotiuc-Pietro, and Pengxiang Cheng.
\newblock Dataless knowledge fusion by merging weights of language models.
\newblock \emph{arXiv preprint arXiv:2212.09849}, 2022.

\bibitem[Juneja et~al.(2022)Juneja, Bansal, Cho, Sedoc, and
  Saphra]{juneja2022linear}
Jeevesh Juneja, Rachit Bansal, Kyunghyun Cho, Jo{\~a}o Sedoc, and Naomi Saphra.
\newblock Linear connectivity reveals generalization strategies.
\newblock \emph{arXiv preprint arXiv:2205.12411}, 2022.

\bibitem[Kirsch et~al.(2018)Kirsch, Kunze, and Barber]{kirsch2018modular}
Louis Kirsch, Julius Kunze, and David Barber.
\newblock Modular networks: Learning to decompose neural computation.
\newblock \emph{Advances in neural information processing systems}, 31, 2018.

\bibitem[Komatsuzaki et~al.(2022)Komatsuzaki, Puigcerver, Lee-Thorp, Ruiz,
  Mustafa, Ainslie, Tay, Dehghani, and Houlsby]{komatsuzaki2022sparse}
Aran Komatsuzaki, Joan Puigcerver, James Lee-Thorp, Carlos~Riquelme Ruiz, Basil
  Mustafa, Joshua Ainslie, Yi~Tay, Mostafa Dehghani, and Neil Houlsby.
\newblock Sparse upcycling: Training mixture-of-experts from dense checkpoints.
\newblock \emph{arXiv preprint arXiv:2212.05055}, 2022.

\bibitem[Kool et~al.(2019)Kool, van Hoof, and Welling]{kool2019buy}
Wouter Kool, Herke van Hoof, and Max Welling.
\newblock Buy 4 reinforce samples, get a baseline for free!
\newblock 2019.

\bibitem[Kool et~al.(2021)Kool, Maddison, and Mnih]{kool2021unbiased}
Wouter Kool, Chris~J Maddison, and Andriy Mnih.
\newblock Unbiased gradient estimation with balanced assignments for mixtures
  of experts.
\newblock \emph{arXiv preprint arXiv:2109.11817}, 2021.

\bibitem[Kudugunta et~al.(2021)Kudugunta, Huang, Bapna, Krikun, Lepikhin,
  Luong, and Firat]{kudugunta2021beyond}
Sneha Kudugunta, Yanping Huang, Ankur Bapna, Maxim Krikun, Dmitry Lepikhin,
  Minh-Thang Luong, and Orhan Firat.
\newblock Beyond distillation: Task-level mixture-of-experts for efficient
  inference.
\newblock \emph{arXiv preprint arXiv:2110.03742}, 2021.

\bibitem[Lee-Thorp \& Ainslie(2022)Lee-Thorp and Ainslie]{lee2022sparse}
James Lee-Thorp and Joshua Ainslie.
\newblock Sparse mixers: Combining moe and mixing to build a more efficient
  bert.
\newblock \emph{arXiv preprint arXiv:2205.12399}, 2022.

\bibitem[Lepikhin et~al.(2020)Lepikhin, Lee, Xu, Chen, Firat, Huang, Krikun,
  Shazeer, and Chen]{lepikhin2020gshard}
Dmitry Lepikhin, HyoukJoong Lee, Yuanzhong Xu, Dehao Chen, Orhan Firat, Yanping
  Huang, Maxim Krikun, Noam Shazeer, and Zhifeng Chen.
\newblock Gshard: Scaling giant models with conditional computation and
  automatic sharding.
\newblock \emph{arXiv preprint arXiv:2006.16668}, 2020.

\bibitem[Levesque et~al.(2012)Levesque, Davis, and
  Morgenstern]{levesque2012winograd}
Hector Levesque, Ernest Davis, and Leora Morgenstern.
\newblock The winograd schema challenge.
\newblock In \emph{Thirteenth international conference on the principles of
  knowledge representation and reasoning}, 2012.

\bibitem[Lewis et~al.(2021)Lewis, Bhosale, Dettmers, Goyal, and
  Zettlemoyer]{lewis2021base}
Mike Lewis, Shruti Bhosale, Tim Dettmers, Naman Goyal, and Luke Zettlemoyer.
\newblock Base layers: Simplifying training of large, sparse models.
\newblock In \emph{International Conference on Machine Learning}, pp.\
  6265--6274. PMLR, 2021.

\bibitem[Li et~al.(2022{\natexlab{a}})Li, Yang, Ren, Wang, and
  Liu]{li2022sparse}
Bo~Li, Jingkang Yang, Jiawei Ren, Yezhen Wang, and Ziwei Liu.
\newblock Sparse fusion mixture-of-experts are domain generalizable learners.
\newblock \emph{arXiv preprint arXiv:2206.04046}, 2022{\natexlab{a}}.

\bibitem[Li et~al.(2022{\natexlab{b}})Li, Gururangan, Dettmers, Lewis, Althoff,
  Smith, and Zettlemoyer]{Li2022BranchTrainMergeEP}
Margaret Li, Suchin Gururangan, Tim Dettmers, Mike Lewis, Tim Althoff, Noah~A.
  Smith, and Luke Zettlemoyer.
\newblock Branch-train-merge: Embarrassingly parallel training of expert
  language models.
\newblock \emph{ArXiv}, abs/2208.03306, 2022{\natexlab{b}}.

\bibitem[Liu et~al.(2022{\natexlab{a}})Liu, Tam, Muqeeth, Mohta, Huang, Bansal,
  and Raffel]{liu2022few}
Haokun Liu, Derek Tam, Mohammed Muqeeth, Jay Mohta, Tenghao Huang, Mohit
  Bansal, and Colin Raffel.
\newblock Few-shot parameter-efficient fine-tuning is better and cheaper than
  in-context learning.
\newblock \emph{arXiv preprint arXiv:2205.05638}, 2022{\natexlab{a}}.

\bibitem[Liu et~al.(2022{\natexlab{b}})Liu, Kim, Muzio, and
  Hassan]{liu2022gating}
Rui Liu, Young~Jin Kim, Alexandre Muzio, and Hany Hassan.
\newblock Gating dropout: Communication-efficient regularization for sparsely
  activated transformers.
\newblock In \emph{International Conference on Machine Learning}, pp.\
  13782--13792. PMLR, 2022{\natexlab{b}}.

\bibitem[Liu et~al.(2020{\natexlab{a}})Liu, Zhou, Zhao, Wang, Deng, and
  Ju]{liu2020fastbert}
Weijie Liu, Peng Zhou, Zhe Zhao, Zhiruo Wang, Haotang Deng, and Qi~Ju.
\newblock Fastbert: a self-distilling bert with adaptive inference time.
\newblock \emph{arXiv preprint arXiv:2004.02178}, 2020{\natexlab{a}}.

\bibitem[Liu et~al.(2020{\natexlab{b}})Liu, Gu, Goyal, Li, Edunov,
  Ghazvininejad, Lewis, and Zettlemoyer]{liu2020multilingual}
Yinhan Liu, Jiatao Gu, Naman Goyal, Xian Li, Sergey Edunov, Marjan
  Ghazvininejad, Mike Lewis, and Luke Zettlemoyer.
\newblock Multilingual denoising pre-training for neural machine translation.
\newblock \emph{Transactions of the Association for Computational Linguistics},
  8:\penalty0 726--742, 2020{\natexlab{b}}.

\bibitem[Ma et~al.(2019)Ma, Zhao, Chen, Li, Hong, and Chi]{ma2019snr}
Jiaqi Ma, Zhe Zhao, Jilin Chen, Ang Li, Lichan Hong, and Ed~H Chi.
\newblock Snr: Sub-network routing for flexible parameter sharing in multi-task
  learning.
\newblock In \emph{Proceedings of the AAAI Conference on Artificial
  Intelligence}, volume~33, pp.\  216--223, 2019.

\bibitem[Maddison et~al.(2016)Maddison, Mnih, and Teh]{maddison2016concrete}
Chris~J Maddison, Andriy Mnih, and Yee~Whye Teh.
\newblock The concrete distribution: A continuous relaxation of discrete random
  variables.
\newblock \emph{arXiv preprint arXiv:1611.00712}, 2016.

\bibitem[Mahabadi et~al.(2021)Mahabadi, Ruder, Dehghani, and
  Henderson]{mahabadi2021parameter}
Rabeeh~Karimi Mahabadi, Sebastian Ruder, Mostafa Dehghani, and James Henderson.
\newblock Parameter-efficient multi-task fine-tuning for transformers via
  shared hypernetworks.
\newblock \emph{arXiv preprint arXiv:2106.04489}, 2021.

\bibitem[Matena \& Raffel(2021)Matena and Raffel]{matena2021merging}
Michael Matena and Colin Raffel.
\newblock Merging models with fisher-weighted averaging.
\newblock \emph{arXiv preprint arXiv:2111.09832}, 2021.

\bibitem[Maziarz et~al.(2019)Maziarz, Kokiopoulou, Gesmundo, Sbaiz, Bartok, and
  Berent]{maziarz2019flexible}
Krzysztof Maziarz, Efi Kokiopoulou, Andrea Gesmundo, Luciano Sbaiz, Gabor
  Bartok, and Jesse Berent.
\newblock Flexible multi-task networks by learning parameter allocation.
\newblock \emph{arXiv preprint arXiv:1910.04915}, 2019.

\bibitem[McMahan et~al.(2017)McMahan, Moore, Ramage, Hampson, and
  Arcas]{mcmahan2017communication}
Brendan McMahan, Eider Moore, Daniel Ramage, Seth Hampson, and Blaise Aguera~y
  Arcas.
\newblock Communication-efficient learning of deep networks from decentralized
  data.
\newblock In \emph{Artificial Intelligence and Sstatistics}, 2017.

\bibitem[Mittal et~al.(2022)Mittal, Bengio, and Lajoie]{mittal2022modular}
Sarthak Mittal, Yoshua Bengio, and Guillaume Lajoie.
\newblock Is a modular architecture enough?
\newblock \emph{arXiv preprint arXiv:2206.02713}, 2022.

\bibitem[Peng et~al.(2019)Peng, Bai, Xia, Huang, Saenko, and
  Wang]{peng2019moment}
Xingchao Peng, Qinxun Bai, Xide Xia, Zijun Huang, Kate Saenko, and Bo~Wang.
\newblock Moment matching for multi-source domain adaptation.
\newblock In \emph{Proceedings of the IEEE/CVF international conference on
  computer vision}, pp.\  1406--1415, 2019.

\bibitem[Pfeiffer et~al.(2020)Pfeiffer, Vulic, Gurevych, and
  Ruder]{Pfeiffer2020MADXAA}
Jonas Pfeiffer, Ivan Vulic, Iryna Gurevych, and Sebastian Ruder.
\newblock Mad-x: An adapter-based framework for multi-task cross-lingual
  transfer.
\newblock In \emph{EMNLP}, 2020.

\bibitem[Pfeiffer et~al.(2022)Pfeiffer, Goyal, Lin, Li, Cross, Riedel, and
  Artetxe]{Pfeiffer2022LiftingTC}
Jonas Pfeiffer, Naman Goyal, Xi~Victoria Lin, Xian Li, James Cross, Sebastian
  Riedel, and Mikel Artetxe.
\newblock Lifting the curse of multilinguality by pre-training modular
  transformers.
\newblock In \emph{NAACL}, 2022.

\bibitem[Pfeiffer et~al.(2023)Pfeiffer, Ruder, Vuli{\'c}, and
  Ponti]{pfeiffer2023modular}
Jonas Pfeiffer, Sebastian Ruder, Ivan Vuli{\'c}, and Edoardo~Maria Ponti.
\newblock Modular deep learning.
\newblock \emph{arXiv preprint arXiv:2302.11529}, 2023.

\bibitem[Phang et~al.(2018)Phang, F{\'e}vry, and Bowman]{phang2018sentence}
Jason Phang, Thibault F{\'e}vry, and Samuel~R Bowman.
\newblock Sentence encoders on stilts: Supplementary training on intermediate
  labeled-data tasks.
\newblock \emph{arXiv preprint arXiv:1811.01088}, 2018.

\bibitem[Pires et~al.(2019)Pires, Schlinger, and
  Garrette]{pires2019multilingual}
Telmo Pires, Eva Schlinger, and Dan Garrette.
\newblock How multilingual is multilingual bert?
\newblock \emph{arXiv preprint arXiv:1906.01502}, 2019.

\bibitem[Ponti et~al.(2022)Ponti, Sordoni, and Reddy]{ponti2022combining}
Edoardo~M Ponti, Alessandro Sordoni, and Siva Reddy.
\newblock Combining modular skills in multitask learning.
\newblock \emph{arXiv preprint arXiv:2202.13914}, 2022.

\bibitem[Pruksachatkun et~al.(2020)Pruksachatkun, Phang, Liu, Htut, Zhang,
  Pang, Vania, Kann, and Bowman]{pruksachatkun2020intermediate}
Yada Pruksachatkun, Jason Phang, Haokun Liu, Phu~Mon Htut, Xiaoyi Zhang,
  Richard~Yuanzhe Pang, Clara Vania, Katharina Kann, and Samuel~R Bowman.
\newblock Intermediate-task transfer learning with pretrained models for
  natural language understanding: When and why does it work?
\newblock \emph{arXiv preprint arXiv:2005.00628}, 2020.

\bibitem[Puigcerver et~al.(2023)Puigcerver, Riquelme, Mustafa, and
  Houlsby]{puigcerver2023sparse}
Joan Puigcerver, Carlos Riquelme, Basil Mustafa, and Neil Houlsby.
\newblock From sparse to soft mixtures of experts.
\newblock \emph{arXiv preprint arXiv:2308.00951}, 2023.

\bibitem[Qin et~al.(2022)Qin, Qian, Yi, Chen, Lin, Han, Liu, Sun, and
  Zhou]{qin2022exploring}
Yujia Qin, Cheng Qian, Jing Yi, Weize Chen, Yankai Lin, Xu~Han, Zhiyuan Liu,
  Maosong Sun, and Jie Zhou.
\newblock Exploring mode connectivity for pre-trained language models.
\newblock \emph{arXiv preprint arXiv:2210.14102}, 2022.

\bibitem[Raffel et~al.(2020)Raffel, Shazeer, Roberts, Lee, Narang, Matena,
  Zhou, Li, Liu, et~al.]{raffel2020exploring}
Colin Raffel, Noam Shazeer, Adam Roberts, Katherine Lee, Sharan Narang, Michael
  Matena, Yanqi Zhou, Wei Li, Peter~J Liu, et~al.
\newblock Exploring the limits of transfer learning with a unified text-to-text
  transformer.
\newblock \emph{J. Mach. Learn. Res.}, 21\penalty0 (140):\penalty0 1--67, 2020.

\bibitem[Rajpurkar et~al.(2016)Rajpurkar, Zhang, Lopyrev, and
  Liang]{rajpurkar2016squad}
Pranav Rajpurkar, Jian Zhang, Konstantin Lopyrev, and Percy Liang.
\newblock Squad: 100,000+ questions for machine comprehension of text.
\newblock \emph{arXiv preprint arXiv:1606.05250}, 2016.

\bibitem[Rebuffi et~al.(2017)Rebuffi, Bilen, and Vedaldi]{rebuffi2017learning}
Sylvestre-Alvise Rebuffi, Hakan Bilen, and Andrea Vedaldi.
\newblock Learning multiple visual domains with residual adapters.
\newblock \emph{Advances in neural information processing systems}, 30, 2017.

\bibitem[Roller et~al.(2021)Roller, Sukhbaatar, Weston, et~al.]{roller2021hash}
Stephen Roller, Sainbayar Sukhbaatar, Jason Weston, et~al.
\newblock Hash layers for large sparse models.
\newblock \emph{Advances in Neural Information Processing Systems},
  34:\penalty0 17555--17566, 2021.

\bibitem[Sanh et~al.(2021)Sanh, Webson, Raffel, Bach, Sutawika, Alyafeai,
  Chaffin, Stiegler, Scao, Raja, et~al.]{sanh2021multitask}
Victor Sanh, Albert Webson, Colin Raffel, Stephen~H Bach, Lintang Sutawika,
  Zaid Alyafeai, Antoine Chaffin, Arnaud Stiegler, Teven~Le Scao, Arun Raja,
  et~al.
\newblock Multitask prompted training enables zero-shot task generalization.
\newblock \emph{arXiv preprint arXiv:2110.08207}, 2021.

\bibitem[Schulman et~al.(2015)Schulman, Heess, Weber, and
  Abbeel]{schulman2015gradient}
John Schulman, Nicolas Heess, Theophane Weber, and Pieter Abbeel.
\newblock Gradient estimation using stochastic computation graphs.
\newblock \emph{Advances in Neural Information Processing Systems}, 28, 2015.

\bibitem[Shankar et~al.(2017)Shankar, Nikhil, and Kornel]{shankar2017first}
Iyer Shankar, Dandekar Nikhil, and Csernai Kornel.
\newblock First quora dataset release: question pairs (2017).
\newblock
  \url{https://www.quora.com/q/quoradata/First-Quora-Dataset-Release-Question-Pairs},
  2017.

\bibitem[Shazeer et~al.(2017)Shazeer, Mirhoseini, Maziarz, Davis, Le, Hinton,
  and Dean]{shazeer2017outrageously}
Noam Shazeer, Azalia Mirhoseini, Krzysztof Maziarz, Andy Davis, Quoc Le,
  Geoffrey Hinton, and Jeff Dean.
\newblock Outrageously large neural networks: The sparsely-gated
  mixture-of-experts layer.
\newblock \emph{arXiv preprint arXiv:1701.06538}, 2017.

\bibitem[Socher et~al.(2013)Socher, Perelygin, Wu, Chuang, Manning, Ng, and
  Potts]{socher2013recursive}
Richard Socher, Alex Perelygin, Jean Wu, Jason Chuang, Christopher~D Manning,
  Andrew~Y Ng, and Christopher Potts.
\newblock Recursive deep models for semantic compositionality over a sentiment
  treebank.
\newblock In \emph{Proceedings of the 2013 conference on empirical methods in
  natural language processing}, pp.\  1631--1642, 2013.

\bibitem[Tang et~al.(2022)Tang, Zhang, Dai, Zhou, Wu, and Shi]{Tang2022OneMM}
Duyu Tang, Fan Zhang, Yong Dai, Cong Zhou, Shuangzhi Wu, and Shuming Shi.
\newblock One model, multiple tasks: Pathways for natural language
  understanding.
\newblock \emph{ArXiv}, abs/2203.03312, 2022.

\bibitem[Tucker et~al.(2017)Tucker, Mnih, Maddison, Lawson, and
  Sohl-Dickstein]{tucker2017rebar}
George Tucker, Andriy Mnih, Chris~J Maddison, John Lawson, and Jascha
  Sohl-Dickstein.
\newblock Rebar: Low-variance, unbiased gradient estimates for discrete latent
  variable models.
\newblock \emph{Advances in Neural Information Processing Systems}, 30, 2017.

\bibitem[Vu et~al.(2020)Vu, Wang, Munkhdalai, Sordoni, Trischler,
  Mattarella-Micke, Maji, and Iyyer]{vu2020exploring}
Tu~Vu, Tong Wang, Tsendsuren Munkhdalai, Alessandro Sordoni, Adam Trischler,
  Andrew Mattarella-Micke, Subhransu Maji, and Mohit Iyyer.
\newblock Exploring and predicting transferability across nlp tasks.
\newblock \emph{arXiv preprint arXiv:2005.00770}, 2020.

\bibitem[Wang et~al.(2018)Wang, Singh, Michael, Hill, Levy, and
  Bowman]{wang2018glue}
Alex Wang, Amanpreet Singh, Julian Michael, Felix Hill, Omer Levy, and Samuel~R
  Bowman.
\newblock Glue: A multi-task benchmark and analysis platform for natural
  language understanding.
\newblock \emph{arXiv preprint arXiv:1804.07461}, 2018.

\bibitem[Wang et~al.(2022{\natexlab{a}})Wang, Wen, Zhang, Hou, Liu, and
  Li]{wang2022finding}
Xiaozhi Wang, Kaiyue Wen, Zhengyan Zhang, Lei Hou, Zhiyuan Liu, and Juanzi Li.
\newblock Finding skill neurons in pre-trained transformer-based language
  models.
\newblock \emph{arXiv preprint arXiv:2211.07349}, 2022{\natexlab{a}}.

\bibitem[Wang et~al.(2022{\natexlab{b}})Wang, Mukherjee, Liu, Gao, Awadallah,
  and Gao]{wang2022adamix}
Yaqing Wang, Subhabrata Mukherjee, Xiaodong Liu, Jing Gao, Ahmed~Hassan
  Awadallah, and Jianfeng Gao.
\newblock Adamix: Mixture-of-adapter for parameter-efficient tuning of large
  language models.
\newblock \emph{arXiv preprint arXiv:2205.12410}, 2022{\natexlab{b}}.

\bibitem[Warstadt et~al.(2019)Warstadt, Singh, and Bowman]{warstadt2019neural}
Alex Warstadt, Amanpreet Singh, and Samuel~R Bowman.
\newblock Neural network acceptability judgments.
\newblock \emph{Transactions of the Association for Computational Linguistics},
  7:\penalty0 625--641, 2019.

\bibitem[Wei et~al.(2021)Wei, Bosma, Zhao, Guu, Yu, Lester, Du, Dai, and
  Le]{wei2021finetuned}
Jason Wei, Maarten Bosma, Vincent~Y Zhao, Kelvin Guu, Adams~Wei Yu, Brian
  Lester, Nan Du, Andrew~M Dai, and Quoc~V Le.
\newblock Finetuned language models are zero-shot learners.
\newblock \emph{arXiv preprint arXiv:2109.01652}, 2021.

\bibitem[Williams et~al.(2017)Williams, Nangia, and Bowman]{williams2017broad}
Adina Williams, Nikita Nangia, and Samuel~R Bowman.
\newblock A broad-coverage challenge corpus for sentence understanding through
  inference.
\newblock \emph{arXiv preprint arXiv:1704.05426}, 2017.

\bibitem[Wortsman et~al.(2022{\natexlab{a}})Wortsman, Gururangan, Li, Farhadi,
  Schmidt, Rabbat, and Morcos]{wortsman2022fi}
Mitchell Wortsman, Suchin Gururangan, Shen Li, Ali Farhadi, Ludwig Schmidt,
  Michael Rabbat, and Ari~S Morcos.
\newblock lo-fi: distributed fine-tuning without communication.
\newblock \emph{arXiv preprint arXiv:2210.11948}, 2022{\natexlab{a}}.

\bibitem[Wortsman et~al.(2022{\natexlab{b}})Wortsman, Ilharco, Gadre, Roelofs,
  Gontijo-Lopes, Morcos, Namkoong, Farhadi, Carmon, Kornblith,
  et~al.]{wortsman2022model}
Mitchell Wortsman, Gabriel Ilharco, Samir~Ya Gadre, Rebecca Roelofs, Raphael
  Gontijo-Lopes, Ari~S Morcos, Hongseok Namkoong, Ali Farhadi, Yair Carmon,
  Simon Kornblith, et~al.
\newblock Model soups: averaging weights of multiple fine-tuned models improves
  accuracy without increasing inference time.
\newblock In \emph{International Conference on Machine Learning}, pp.\
  23965--23998. PMLR, 2022{\natexlab{b}}.

\bibitem[Wortsman et~al.(2022{\natexlab{c}})Wortsman, Ilharco, Kim, Li,
  Kornblith, Roelofs, Lopes, Hajishirzi, Farhadi, Namkoong,
  et~al.]{wortsman2022robust}
Mitchell Wortsman, Gabriel Ilharco, Jong~Wook Kim, Mike Li, Simon Kornblith,
  Rebecca Roelofs, Raphael~Gontijo Lopes, Hannaneh Hajishirzi, Ali Farhadi,
  Hongseok Namkoong, et~al.
\newblock Robust fine-tuning of zero-shot models.
\newblock In \emph{Proceedings of the IEEE/CVF Conference on Computer Vision
  and Pattern Recognition}, pp.\  7959--7971, 2022{\natexlab{c}}.

\bibitem[Wu et~al.(2023)Wu, Wang, Ge, Lu, Zhou, Shan, and Luo]{wu2023pi}
Chengyue Wu, Teng Wang, Yixiao Ge, Zeyu Lu, Ruisong Zhou, Ying Shan, and Ping
  Luo.
\newblock $pi$-tuning: Transferring multimodal foundation models with optimal
  multi-task interpolation.
\newblock In \emph{International Conference on Machine Learning}, pp.\
  37713--37727. PMLR, 2023.

\bibitem[Xin et~al.(2020)Xin, Tang, Lee, Yu, and Lin]{xin2020deebert}
Ji~Xin, Raphael Tang, Jaejun Lee, Yaoliang Yu, and Jimmy Lin.
\newblock Deebert: Dynamic early exiting for accelerating bert inference.
\newblock \emph{arXiv preprint arXiv:2004.12993}, 2020.

\bibitem[Xue et~al.(2020)Xue, Constant, Roberts, Kale, Al-Rfou, Siddhant,
  Barua, and Raffel]{xue2020mt5}
Linting Xue, Noah Constant, Adam Roberts, Mihir Kale, Rami Al-Rfou, Aditya
  Siddhant, Aditya Barua, and Colin Raffel.
\newblock mt5: A massively multilingual pre-trained text-to-text transformer.
\newblock \emph{arXiv preprint arXiv:2010.11934}, 2020.

\bibitem[Yang et~al.(2019)Yang, Bender, Le, and Ngiam]{yang2019condconv}
Brandon Yang, Gabriel Bender, Quoc~V Le, and Jiquan Ngiam.
\newblock Condconv: Conditionally parameterized convolutions for efficient
  inference.
\newblock \emph{Advances in Neural Information Processing Systems}, 32, 2019.

\bibitem[Ye et~al.(2022)Ye, Zha, and Ren]{ye2022eliciting}
Qinyuan Ye, Juan Zha, and Xiang Ren.
\newblock Eliciting transferability in multi-task learning with task-level
  mixture-of-experts.
\newblock \emph{arXiv preprint arXiv:2205.12701}, 2022.

\bibitem[Yin \& Zhou(2018)Yin and Zhou]{yin2018arm}
Mingzhang Yin and Mingyuan Zhou.
\newblock Arm: Augment-reinforce-merge gradient for stochastic binary networks.
\newblock \emph{arXiv preprint arXiv:1807.11143}, 2018.

\bibitem[Yu et~al.(2022)Yu, Artetxe, Ott, Shleifer, Gong, Stoyanov, and
  Li]{yu2022efficient}
Ping Yu, Mikel Artetxe, Myle Ott, Sam Shleifer, Hongyu Gong, Ves Stoyanov, and
  Xian Li.
\newblock Efficient language modeling with sparse all-mlp.
\newblock \emph{arXiv preprint arXiv:2203.06850}, 2022.

\bibitem[Zadouri et~al.(2023)Zadouri, {\"U}st{\"u}n, Ahmadian, Ermi{\c{s}},
  Locatelli, and Hooker]{zadouri2023pushing}
Ted Zadouri, Ahmet {\"U}st{\"u}n, Arash Ahmadian, Beyza Ermi{\c{s}}, Acyr
  Locatelli, and Sara Hooker.
\newblock Pushing mixture of experts to the limit: Extremely parameter
  efficient moe for instruction tuning.
\newblock \emph{arXiv preprint arXiv:2309.05444}, 2023.

\bibitem[Zamir et~al.(2018)Zamir, Sax, Shen, Guibas, Malik, and
  Savarese]{zamir2018taskonomy}
Amir~R Zamir, Alexander Sax, William Shen, Leonidas~J Guibas, Jitendra Malik,
  and Silvio Savarese.
\newblock Taskonomy: Disentangling task transfer learning.
\newblock In \emph{Proceedings of the IEEE conference on computer vision and
  pattern recognition}, pp.\  3712--3722, 2018.

\bibitem[Zeiler \& Fergus(2014)Zeiler and Fergus]{zeiler2014visualizing}
Matthew~D. Zeiler and Rob Fergus.
\newblock Visualizing and understanding convolutional networks.
\newblock In \emph{13th European Conference on Computer Vision}, 2014.

\bibitem[Zhang et~al.(2021)Zhang, Chu, Zhmoginov, Howard, Jou, Zhu, Zhang, Hwa,
  and Kovashka]{zhang2021basisnet}
Mingda Zhang, Chun-Te Chu, Andrey Zhmoginov, Andrew Howard, Brendan Jou, Yukun
  Zhu, Li~Zhang, Rebecca Hwa, and Adriana Kovashka.
\newblock Basisnet: Two-stage model synthesis for efficient inference.
\newblock In \emph{Proceedings of the IEEE/CVF Conference on Computer Vision
  and Pattern Recognition}, pp.\  3081--3090, 2021.

\bibitem[Zhang et~al.(2022)Zhang, Lin, Liu, Li, Sun, and
  Zhou]{zhang2022moefication}
Zhengyan Zhang, Yankai Lin, Zhiyuan Liu, Peng Li, Maosong Sun, and Jie Zhou.
\newblock Moefication: Transformer feed-forward layers are mixtures of experts.
\newblock In \emph{Findings of the Association for Computational Linguistics:
  ACL 2022}, pp.\  877--890, 2022.

\bibitem[Zhou et~al.(2022)Zhou, Lei, Liu, Du, Huang, Zhao, Dai, Chen, Le, and
  Laudon]{zhou2022mixture}
Yanqi Zhou, Tao Lei, Hanxiao Liu, Nan Du, Yanping Huang, Vincent Zhao, Andrew
  Dai, Zhifeng Chen, Quoc Le, and James Laudon.
\newblock Mixture-of-experts with expert choice routing.
\newblock \emph{arXiv preprint arXiv:2202.09368}, 2022.

\bibitem[Zoph et~al.(2022)Zoph, Bello, Kumar, Du, Huang, Dean, Shazeer, and
  Fedus]{zoph2022designing}
Barret Zoph, Irwan Bello, Sameer Kumar, Nan Du, Yanping Huang, Jeff Dean, Noam
  Shazeer, and William Fedus.
\newblock Designing effective sparse expert models.
\newblock \emph{arXiv preprint arXiv:2202.08906}, 2022.

\bibitem[Zuo et~al.(2021)Zuo, Liu, Jiao, Kim, Hassan, Zhang, Zhao, and
  Gao]{zuo2021taming}
Simiao Zuo, Xiaodong Liu, Jian Jiao, Young~Jin Kim, Hany Hassan, Ruofei Zhang,
  Tuo Zhao, and Jianfeng Gao.
\newblock Taming sparsely activated transformer with stochastic experts.
\newblock \emph{arXiv preprint arXiv:2110.04260}, 2021.

\end{thebibliography}
\clearpage
\appendix
\section{Compute resources used}
\label{sec:compute_resources}
Here we provide details on the compute resources used in our experiments. All models were trained on 48GB A6000s, except for the Ensemble method, which was trained on 80GB A100s. 
The training time for each T5-GLUE experiment was approximately 108 hours while each ResNet-DomainNet experiment required approximately 11 hours of training.  

\section{Experiment Details}
\label{sec:experiment_details}
In this section, we provide details on the experimental setup and hyperparameter choices for the T5-GLUE and ResNet-DomainNet experiments described in the main text. 
We implemented Adamix \citep{wang2022adamix} and ran with the hyperparameters listed in the below subsections.
In Latent Skills \citep{ponti2022combining}, we use Adapters consistent with all other experiments and chose learning rate ratio of 10 for skill matrix, which we found to be best after sweeping for $\{1,10,100\}$ in both the settings.
For exact implementation details of above methods, we refer the reader to each of the respective works. 

It is also important to mention that SMEAR's memory footprint is comparable to that of other methods. 
Since each expert's size is moderate, we compute expert outputs by preparing an expert for each example and applying them to the examples in parallel utilizing \texttt{torch.bmm} operation. 
Thus, none of the methods need an extra weight-sized tensor. 
Training on T5-GLUE requires around 30GB of memory for SMEAR and other methods, with no significant differences observed in ResNet-DomainNet.

\subsection{T5-GLUE}
\label{sec:t5_glue_hyps}
GLUE consists of nine datasets (SST-2 \citep{socher2013recursive}, CoLA \citep{warstadt2019neural}), MNLI \citep{williams2017broad}, RTE \citep{bentivogli2009fifth}, QQP \citep{shankar2017first}, MRPC \citep{dolan2005automatically}, STS-B \citep{cer2017semeval}, QNLI \citep{rajpurkar2016squad}, and WNLI \citep{levesque2012winograd}) that cover a wide range of natural language processing tasks.
Following convention, we exclude WNLI and use the remaining eight datasets. 
We use the prompted form of these datasets available in PromptSource \citep{bach2022promptsource}, which maps each example into a natural language request-and-response form.
During training, we randomly select a prompt templates for each example; during evaluation, we evaluate each example using all of its dataset's templates.
In the T5-GLUE experiments in this paper, we concatenated all 8 datasets of GLUE and perform multitask training.
T5 models were trained for $600k$ steps using a learning rate of $3e^{-4}$, with $2k$ warmup steps, and batch size of $128$. 
The AdamW optimizer was used with its default settings. 
We ran the ST-Gumbel estimator with a $\tau$ value of $10$ and an anneal rate of $1e^{-6}$ by sweeping $\tau$ in the range of $\{1,10\}$ and the anneal rate in the range of $\{1e^{-4},1e^{-5},1e^{-6}\}$.
For the REINFORCE estimator, we used the same values as in \cite{clark2022unified}, $\alpha$ = $1e^{-2}$, $\beta$ = $5e^{-4}$, and $\gamma$ = $1e^{-2}$.
The adapters here use $\text{swish}$ non-linearity in between.  

\subsection{ResNet-DomainNet}
\label{sec:res_domainnet_hyps}
In the ResNet-DomainNet experiments, all the domains from DomainNet were concatenated to perform multitask training similar to T5-GLUE.
ResNet models were trained for $100k$ steps with batch size of $128$ and a learning rate of $1e^{-3}$, with no warm up, using Adam optimizer.
We used $\tau$ value of $10$ and anneal rate of $1e^{-4}$ for the ST-Gumbel estimator by sweeping $\tau$ in the range of $\{1,10\}$ and the anneal rate in the range of $\{1e^{-4},1e^{-5},1e^{-6}\}$.
The values of $\alpha$, $\beta$, and $\gamma$ for the REINFORCE estimators are same as in T5-GLUE experiments.
The hyperparameter that weighs entropy regularization in Dselect-$k$ is chosen as $0.1$ which worked best among $\{0.01, 0.1, 1\}$. 
The adapters also used a $\text{swish}$ non-linearity in between.  

\section{Expert dropout}
\label{sec:expert_dropout}
Table \ref{tab:dropout_ablation} illustrates the impact of expert dropout on methods that do not make use of metadata but learn adaptive routing, namely methods that are trained through gradient estimation and SMEAR.  
We conduct this ablation on a single seed to limit the amount of computation. 
The results show that SMEAR benefits from an improvement of 2.9\% on T5-GLUE, and Top-$k$ achieves a 1.2\% and 0.2\% improvement on T5-GLUE and ResNet-DomainNet respectively. 
As a result, we include expert dropout for these two methods when discussed in the main text.
However, expert dropout has a negative impact on the performance of ST-Gumbel and REINFORCE methods, and thus, we exclude it for these two methods in the main text.

\begin{table*}[h]
\centering
\begin{tabular}{l c c c}
  \toprule
  Routing & T5-GLUE & ResNet-DomainNet\\
  \midrule
  Top-$k$ & 77.2 & 59.8 \\
  \; w/ Expert dropout 0.1 & 78.4 (+1.2) & 60.0 (+0.2)\\
  ST-Gumbel & 78.3 & 58.3 \\
  \; w/ Expert dropout 0.1 & 77.2 (-1.1)& 57.9 (-0.4)\\
  REINFORCE & 79.8 & 59.8 \\  
  \; w/ Expert dropout 0.1 & 78.2 (-1.6)& 59.8 (+0.0)\\
  SMEAR & 80.2 & 62.0 \\
  \; w/ Expert dropout 0.1 & 83.1 (+2.9)& 62.0 (+0.0)\\
  \bottomrule
\end{tabular}
\captionof{table}{Performance comparision of different adaptive routing methods w and w/o dropout on a single seed. The results indicate that SMEAR and Top-$k$ method benefit from the expert dropout, while ST-Gumbel and REINFORCE are negatively affected.}
\label{tab:dropout_ablation}
\end{table*}

\section{License}
\label{sec:license}
T5 is licensed under Apache 2.0. 
The ResNet model we used is licensed under BSD 3-Clause License. 
QNLI uses CC BY-SA 4.0 license. 
MultiNLI uses data sources of multiple different licenses \citep{williams2017broad}. 
CoLA, SST-2, RTE, MRPC, STS-B, QQP, and DomainNet allow non-commercial research use cases. 
Our code for T5-GLUE is based on Hyperformer\citep{mahabadi2021parameter}, which is shared under Apache 2.0.
The code for ResNet-DomainNet is developed by us.

\section{Ethics Statement}
\label{sec:societal_impacts}
We are not aware of any negative ethical implications of our work.
Our work does not involve human subjects and is primarily focused on diagnosing issues with an efficient class of neural networks.
While conditional computation has been used to design extremely large neural networks \citep{shazeer2017outrageously,fedus2021switch,du2022glam} that have high computational costs (and, correspondingly, energy usage), our work primarily focuses on smaller-scale models.

\section{Full results on T5-GLUE and ResNet-DomainNet}
\label{sec:full_results}
We show the full results of T5-GLUE in \cref{tab:glue_results} and ResNet-DomainNet in \cref{tab:domainnet_results}.
\begin{table}
    \centering
\begin{adjustbox}{angle=90}
    \begin{tabular}{l c c c c c c c c c c c c}
      \toprule
      Routing & RTE  & SST-2 & MRPC  & MRPC  & STS-B  & STS-B  & QQP  & QQP  & MNLI  & QNLI  & CoLA & Average \\
       & acc & acc & f1 & acc & pearson &  spearman & f1 & acc & acc & acc & mcc & \\
      \midrule
      SMEAR & $69.9_{2.6}$ & $90.9_{0.8}$ & $90.5_{1.5}$ & $86.9_{2.2}$ & $87.0_{0.7}$ & $86.6_{0.8}$ & $86.9_{0.3}$ & $90.1_{0.2}$ & $84.9_{0.5}$ & $90.2_{0.6}$ & $33.8_{6.4}$ & $81.6_{1.0}$ \\
      $1\times$ parameters  & $72.3_{2.1}$ & $92.1_{0.5}$ & $89.9_{0.5}$ & $86.0_{0.8}$ & $85.5_{0.8}$ & $85.3_{0.9}$ & $87.0_{0.3}$ & $90.2_{0.2}$ & $84.1_{0.5}$ & $89.9_{0.7}$ & $20.1_{8.1}$ & $80.2_{0.8}$ \\
      $1\times$ compute & $67.3_{3.3}$ & $91.9_{0.2}$ & $89.2_{2.7}$ & $85.5_{3.4}$ & $87.4_{0.8}$ & $87.3_{0.7}$ & $85.6_{0.3}$ & $89.2_{0.2}$ & $84.5_{0.7}$ & $89.9_{0.5}$ & $5.1_{3.7}$ & $78.4_{1.1}$ \\
      Adamix & $70.2_{3.2}$ & $92.4_{0.6}$ & $87.4_{1.4}$ & $83.3_{1.2}$ & $86.7_{0.6}$ & $86.6_{0.7}$ & $85.7_{0.2}$ & $89.4_{0.1}$ & $85.4_{0.3}$ & $90.5_{0.4}$ & $5.1_{1.1}$ & $78.4_{0.4}$ \\
      Hash & $58.8_{2.7}$ & $85.6_{1.3}$ & $77.7_{2.6}$ & $68.5_{3.2}$ & $65.4_{2.4}$ & $65.2_{1.8}$ & $76.8_{0.2}$ & $82.8_{0.3}$ & $72.0_{0.9}$ & $80.0_{0.5}$ & $2.9_{2.5}$ & $66.9_{0.9}$ \\
      Tag & $71.7_{2.9}$ & $90.3_{0.5}$ & $85.4_{0.7}$ & $79.5_{0.8}$ & $82.2_{1.1}$ & $81.5_{1.2}$ & $86.2_{0.4}$ & $89.5_{0.3}$ & $84.4_{0.8}$ & $87.9_{0.9}$ & $25.1_{8.3}$ & $78.5_{1.2}$ \\
      Latent Skills & $70.4_{4.6}$ & $90.8_{0.8}$ & $90.0_{1.0}$ & $85.8_{2.1}$ & $86.6_{1.5}$ & $86.3_{1.4}$ & $86.4_{0.4}$ & $89.8_{0.3}$ & $84.9_{0.9}$ & $89.3_{1.4}$ & $30.8_{5.5}$ & $81.0_{1.6}$ \\
      Top-$k$ & $68.2_{2.3}$ & $92.5_{0.4}$ & $88.6_{0.7}$ & $84.7_{1.7}$ & $87.7_{1.6}$ & $87.4_{1.7}$ & $85.3_{0.4}$ & $89.0_{0.5}$ & $84.9_{0.9}$ & $90.1_{0.9}$ & $2.0_{2.0}$ & $78.2_{0.9}$ \\
      ST-Gumbel & $67.6_{2.3}$ & $92.1_{0.7}$ & $88.8_{1.0}$ & $84.8_{1.7}$ & $86.9_{1.0}$ & $86.8_{0.8}$ & $85.7_{0.1}$ & $89.2_{0.2}$ & $84.5_{0.3}$ & $89.1_{0.5}$ & $1.3_{1.7}$ & $77.9_{0.4}$ \\
      REINFORCE & $70.9_{3.3}$ & $92.6_{0.5}$ & $89.8_{1.6}$ & $86.0_{2.1}$ & $87.4_{0.6}$ & $87.2_{0.5}$ & $86.1_{0.3}$ & $89.5_{0.2}$ & $85.8_{0.4}$ & $90.8_{0.7}$ & $14.1_{6.9}$ & $80.0_{0.8}$ \\
      Ensemble & $72.9_{1.6}$ & $91.5_{0.4}$ & $90.9_{1.4}$ & $87.7_{1.4}$ & $85.7_{1.5}$ & $85.1_{1.6}$ & $86.8_{0.3}$ & $90.1_{0.2}$ & $84.7_{0.5}$ & $89.8_{0.6}$ & $33.7_{6.1}$ & $81.7_{1.0}$ \\
      SMEAR $2 \times$ & $70.9_{3.1}$ & $90.9_{0.7}$ & $89.5_{1.1}$ & $85.8_{1.0}$ & $86.9_{0.9}$ & $86.5_{1.0}$ & $86.8_{0.4}$ & $90.1_{0.3}$ & $84.4_{0.5}$ & $89.6_{0.8}$ & $33.1_{6.5}$ & $81.3_{1.1}$ \\ 
      \bottomrule
    \end{tabular}
\end{adjustbox}
    \caption{Full T5-GLUE results.}
    \label{tab:glue_results}
\end{table}

\begin{table}
    \centering
\begin{adjustbox}{angle=90}
\begin{tabular}{l c c c c c c c}
  \toprule
  Routing & Clipart & Infograph & Painting & Quickdraw & Real & Sketch & Final Accuracy \\
  \midrule
  SMEAR & $64.2_{0.1}$ & $31.2_{0.3}$ & $57.8_{0.3}$ & $62.3_{0.1}$ & $74.3_{0.1}$ & $56.0_{0.2}$ & $62.0_{0.1}$ \\
  $1\times$ parameters & $63.3_{0.3}$ & $29.8_{0.3}$ & $56.4_{0.3}$ & $61.5_{0.1}$ & $72.9_{0.1}$ & $54.9_{0.4}$ & $60.8_{0.1}$ \\
  $1\times$ compute & $60.2_{0.3}$ & $27.9_{0.3}$ & $54.8_{0.1}$ & $59.0_{0.2}$ & $72.3_{0.1}$ & $52.6_{0.2}$ & $59.0_{0.1}$ \\
  Adamix & $58.9_{0.2}$ & $27.0_{0.2}$ & $54.1_{0.2}$ & $57.2_{0.3}$ & $72.1_{0.1}$ & $51.2_{0.2}$ & $58.0_{0.2}$ \\
  Hash & $53.5_{0.3}$ & $23.4_{0.3}$ & $49.8_{0.4}$ & $48.6_{0.3}$ & $68.5_{0.1}$ & $45.7_{0.2}$ & $52.4_{0.1}$ \\
  Tag & $62.8_{0.4}$ & $30.2_{0.3}$ & $58.0_{0.2}$ & $61.7_{0.2}$ & $74.1_{0.1}$ & $55.1_{0.3}$ & $61.4_{0.1}$ \\
  Latent Skills & $64.5_{0.4}$ & $31.2_{0.4}$ & $58.9_{0.1}$ & $61.6_{0.3}$ & $74.2_{0.1}$ & $56.3_{0.2}$ & $61.9_{0.2}$ \\
  Top-$k$ & $61.6_{0.2}$ & $29.6_{0.2}$ & $55.8_{0.4}$ & $60.2_{0.3}$ & $73.0_{0.2}$ & $53.5_{0.1}$ & $60.0_{0.1}$ \\
  ST-Gumbel & $59.9_{0.3}$ & $27.6_{0.4}$ & $54.5_{0.3}$ & $58.1_{0.5}$ & $72.1_{0.2}$ & $51.9_{0.4}$ & $58.5_{0.2}$ \\
  REINFORCE & $61.3_{0.3}$ & $29.1_{0.2}$ & $55.9_{0.2}$ & $60.4_{0.3}$ & $72.8_{0.1}$ & $53.6_{0.2}$ & $60.0_{0.1}$\\
  DSelect-$k$ & $60.6_{0.2}$ & $28.4_{0.3}$ & $55.1_{0.3}$ & $59.5_{0.3}$ & $72.5_{0.2}$ & $52.5_{0.4}$ & $59.3_{0.2}$ \\
  Soft MoE & $62.7_{0.2}$ & $29.3_{0.3}$ & $56.5_{0.2}$ & $60.3_{0.2}$ & $73.3_{0.1}$ & $54.9_{0.1}$ & $60.5_{0.1}$ \\
  Ensemble & $65.7_{0.1}$ & $32.3_{0.0}$ & $58.5_{0.3}$ & $63.7_{0.2}$ & $74.6_{0.1}$ & $57.6_{0.2}$ & $62.9_{0.1}$ \\
  SMEAR $2 \times$ & $65.3_{0.1}$ & $31.8_{0.1}$ & $58.4_{0.5}$ & $63.3_{0.3}$ & $74.9_{0.2}$ & $57.3_{0.1}$ & $62.8_{0.1}$ \\
  \bottomrule
\end{tabular}
\end{adjustbox}
    \caption{Full ResNet-DomainNet results.}
    \label{tab:domainnet_results}
\end{table}

\section{Routing distribution in all routing blocks}
\label{sec:router_dist_all_layers}
\begin{figure*}[h]
 
    \centering
    \includegraphics[width=0.9\textwidth]{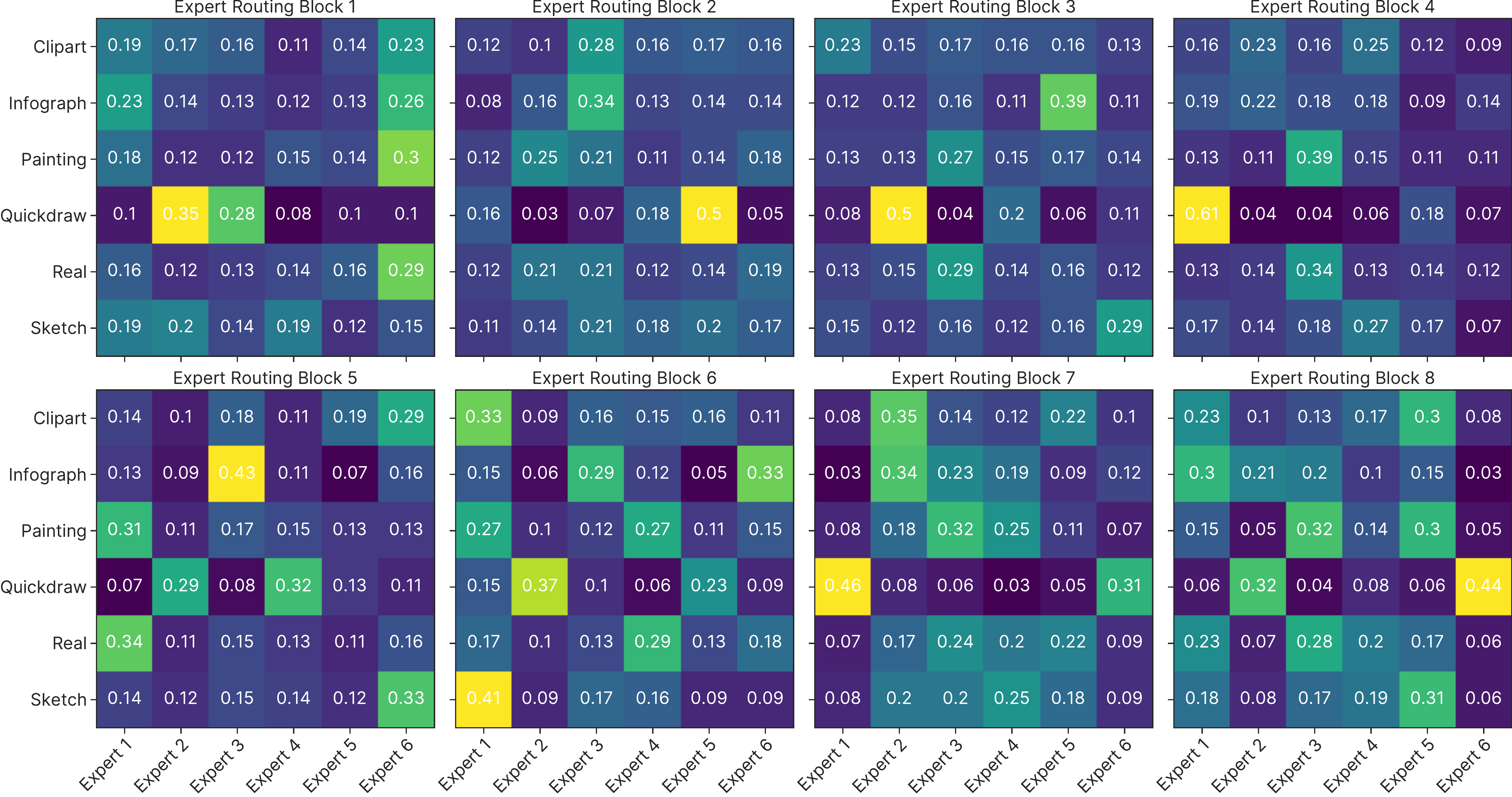}
    \caption{Routing distribution learnt by SMEAR in the routing blocks of ResNet-DomainNet}
    \label{fig:domainnet_routing}    

\end{figure*}

\begin{figure*}[htp]
 
    \centering
    \includegraphics[width=0.9\textwidth]{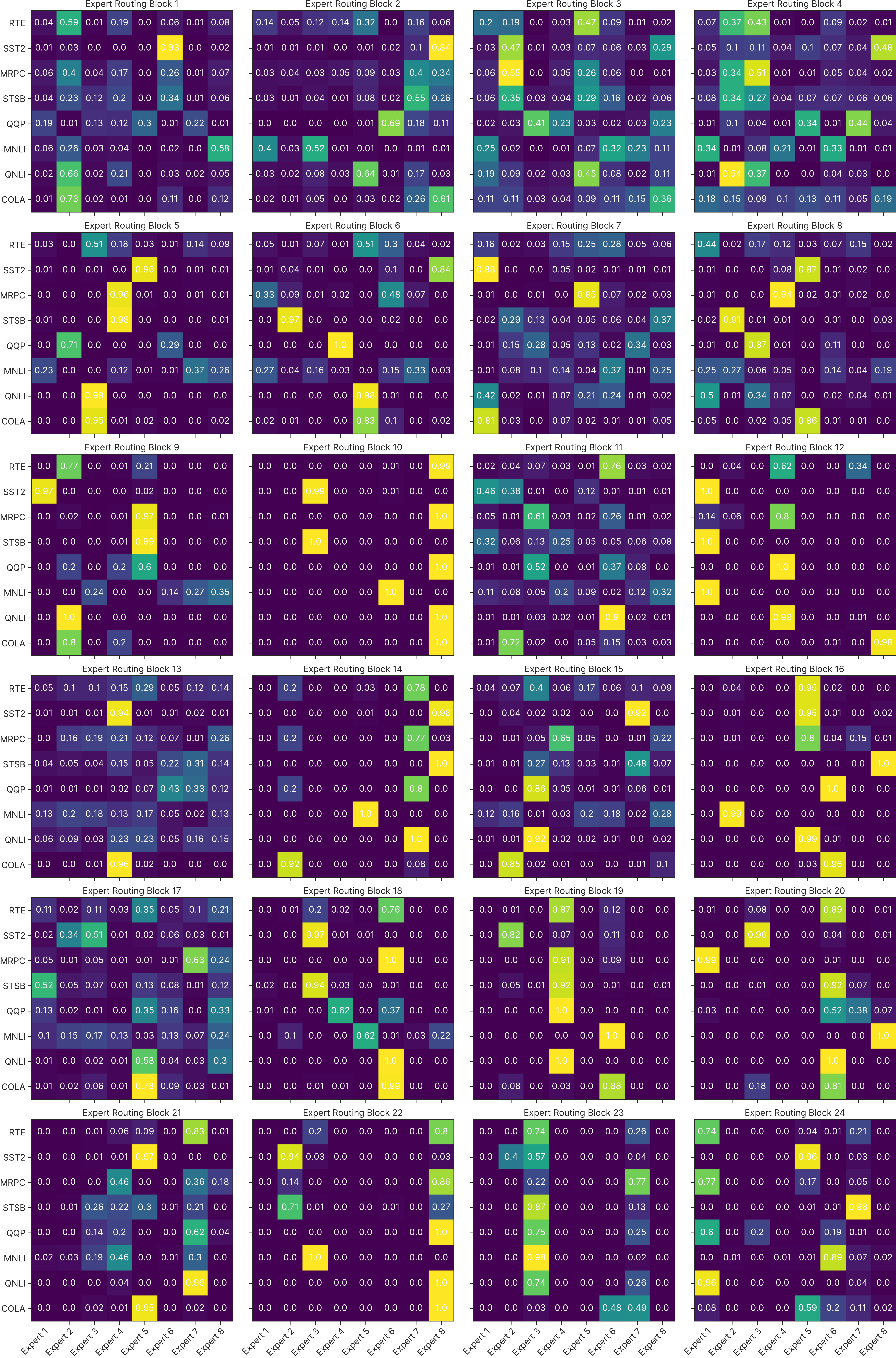}
    \caption{Routing distribution learnt by SMEAR in the encoder routing blocks (1-24) of T5-GLUE}
    \label{fig:glue_routing_layer0to23}    

\end{figure*}

\begin{figure*}[htp]
 
    \centering
    \includegraphics[width=0.9\textwidth]{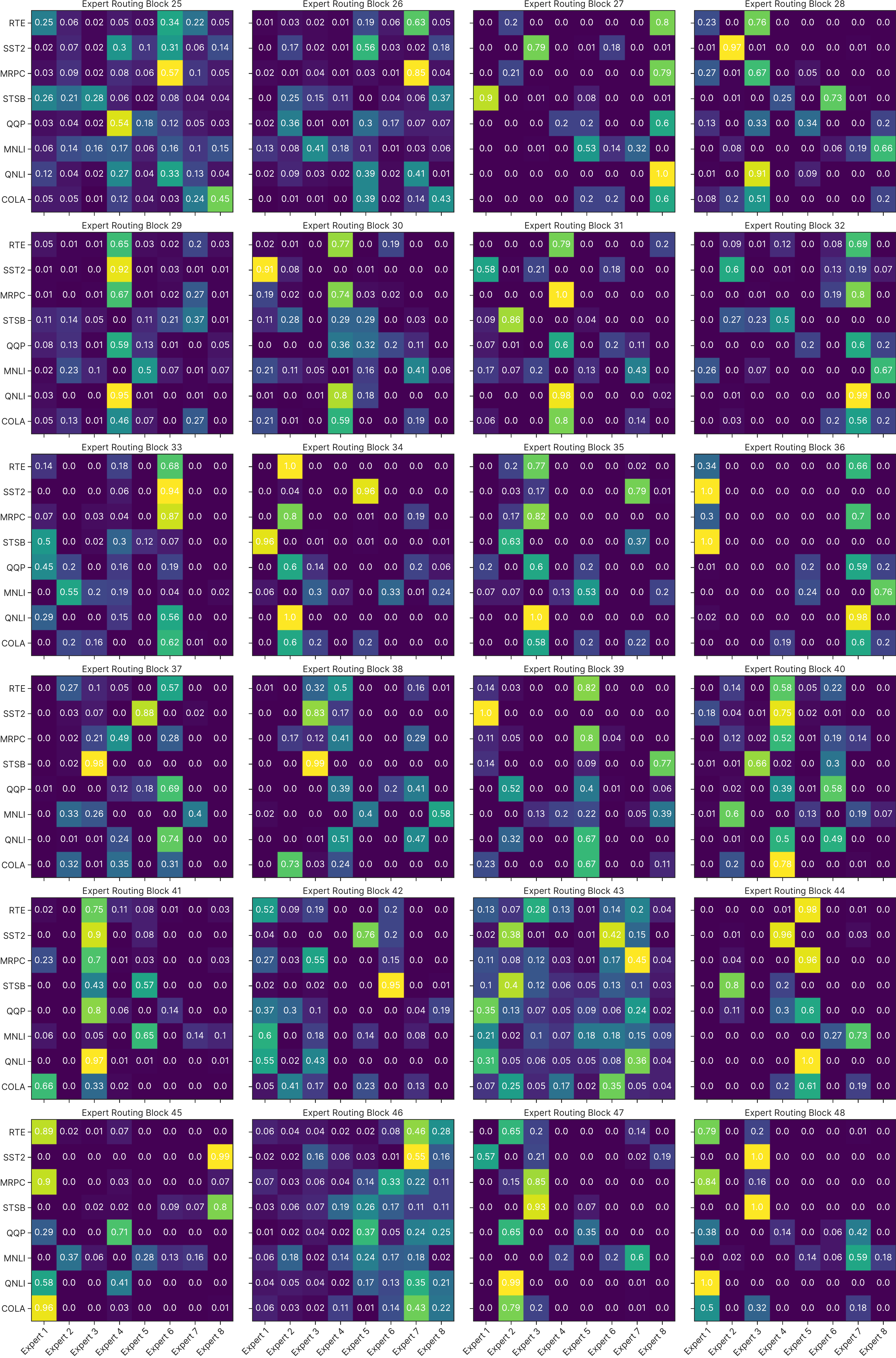}
    \caption{Routing distribution learnt by SMEAR in the decoder routing blocks (25-48) of T5-GLUE}
    \label{fig:glue_routing_layer24to48}    

\end{figure*}

\begin{figure*}[h]
 
    \centering
    \includegraphics[width=0.9\textwidth]{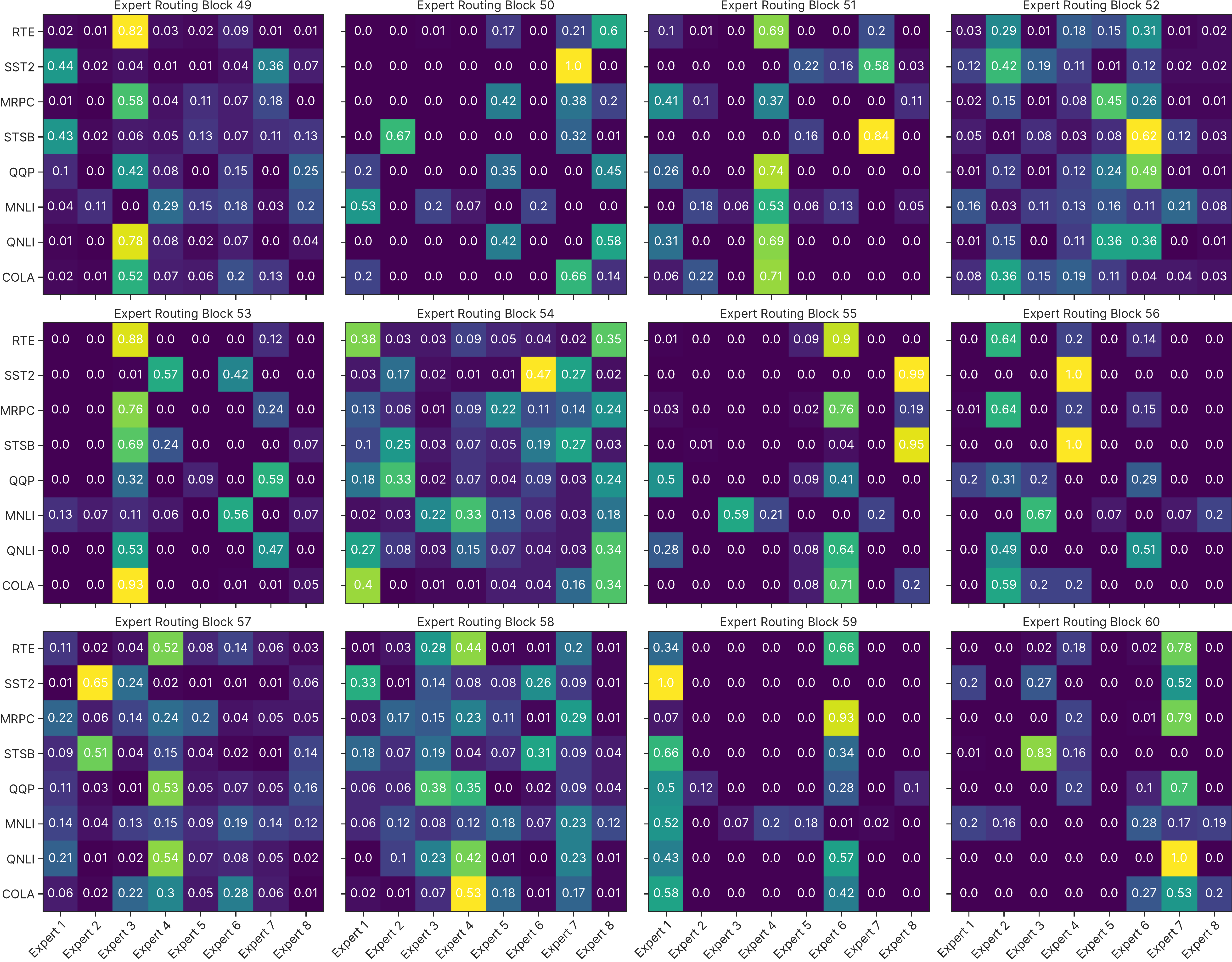}
    \caption{Routing distribution learnt by SMEAR in the decoder routing blocks (49-60) of T5-GLUE}
    \label{fig:glue_routing_layer48to60}    

\end{figure*}

Here, we present the routing distribution across all routing blocks in both T5-GLUE and ResNet-DomainNet as learned by SMEAR. This is depicted in \cref{fig:domainnet_routing}, \cref{fig:glue_routing_layer0to23}, \cref{fig:glue_routing_layer24to48}, and \cref{fig:glue_routing_layer48to60}.

\begin{figure*}[h]
 
    \centering
    \includegraphics[width=0.9\textwidth]{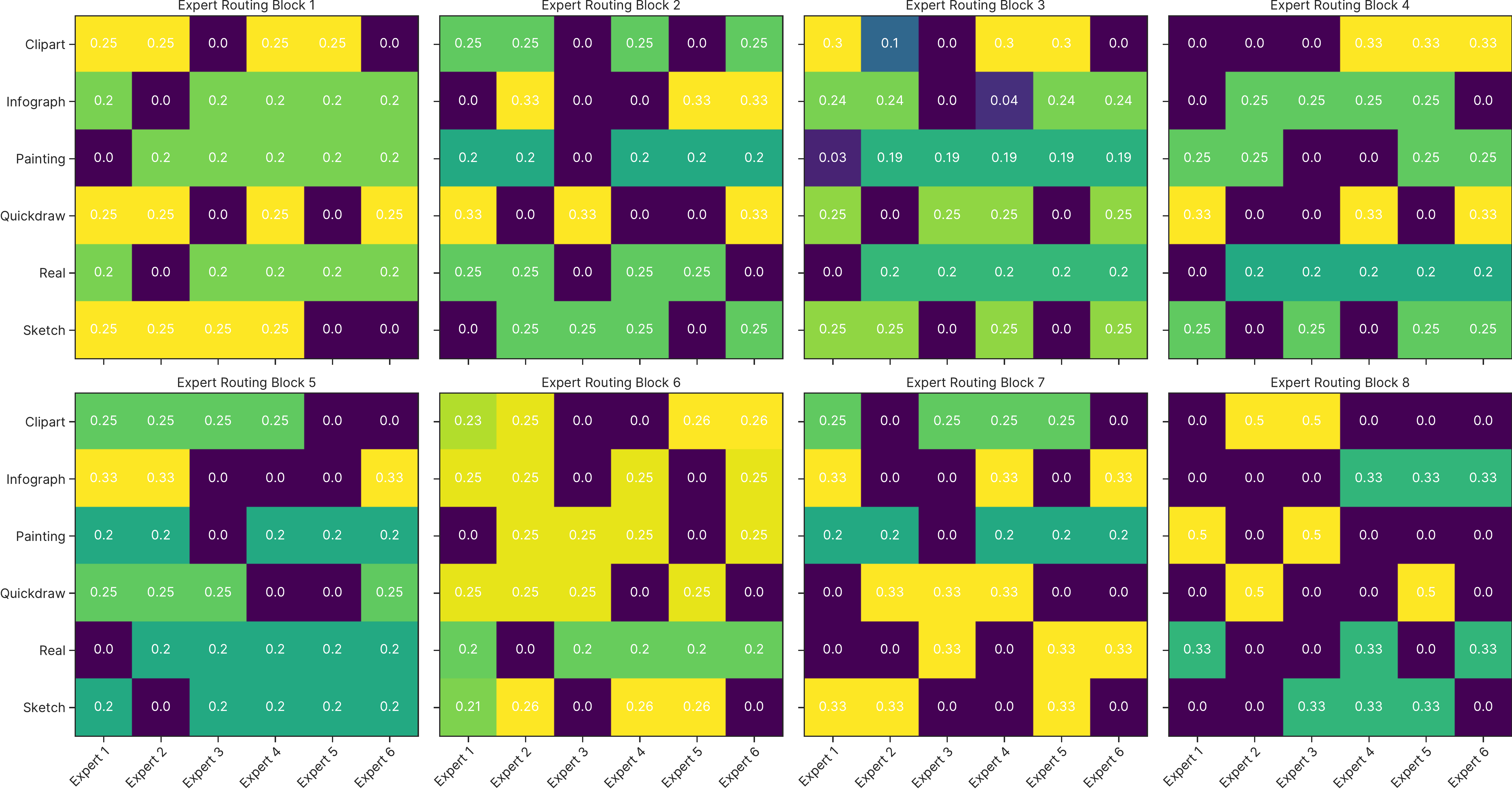}
    \caption{Routing distribution learnt by Latent Skills routing in the routing blocks of ResNet-DomainNet}
    \label{fig:Latentskills_domainnet_routing}    

\end{figure*}

\begin{figure*}[htp]
 
    \centering
    \includegraphics[width=0.9\textwidth]{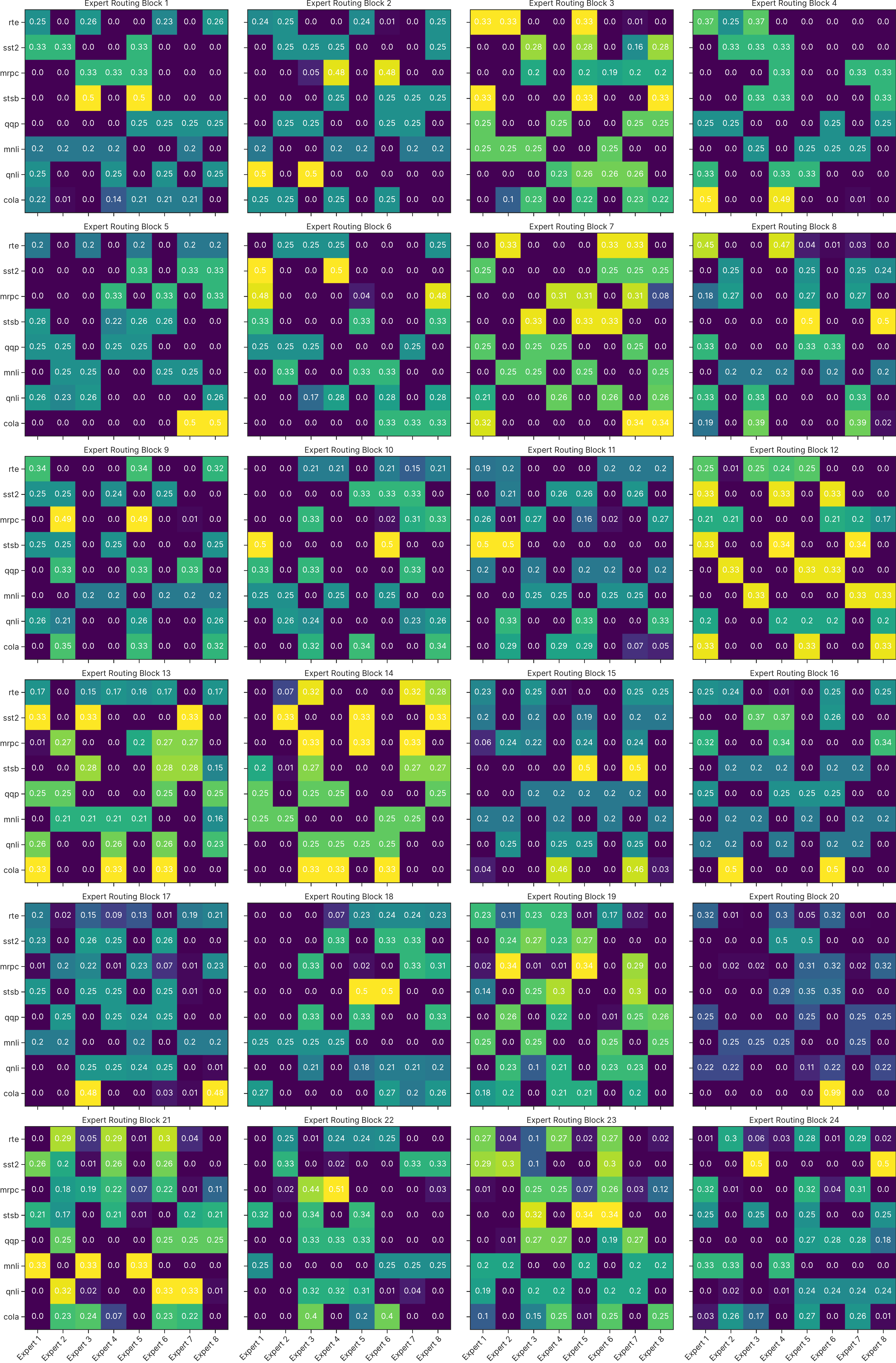}
    \caption{Routing distribution learnt by Latent Skills in the encoder routing blocks (1-24) of T5-GLUE}
    \label{fig:Latentskills_glue_routing_layer0to23}    

\end{figure*}

\begin{figure*}[htp]
 
    \centering
    \includegraphics[width=0.9\textwidth]{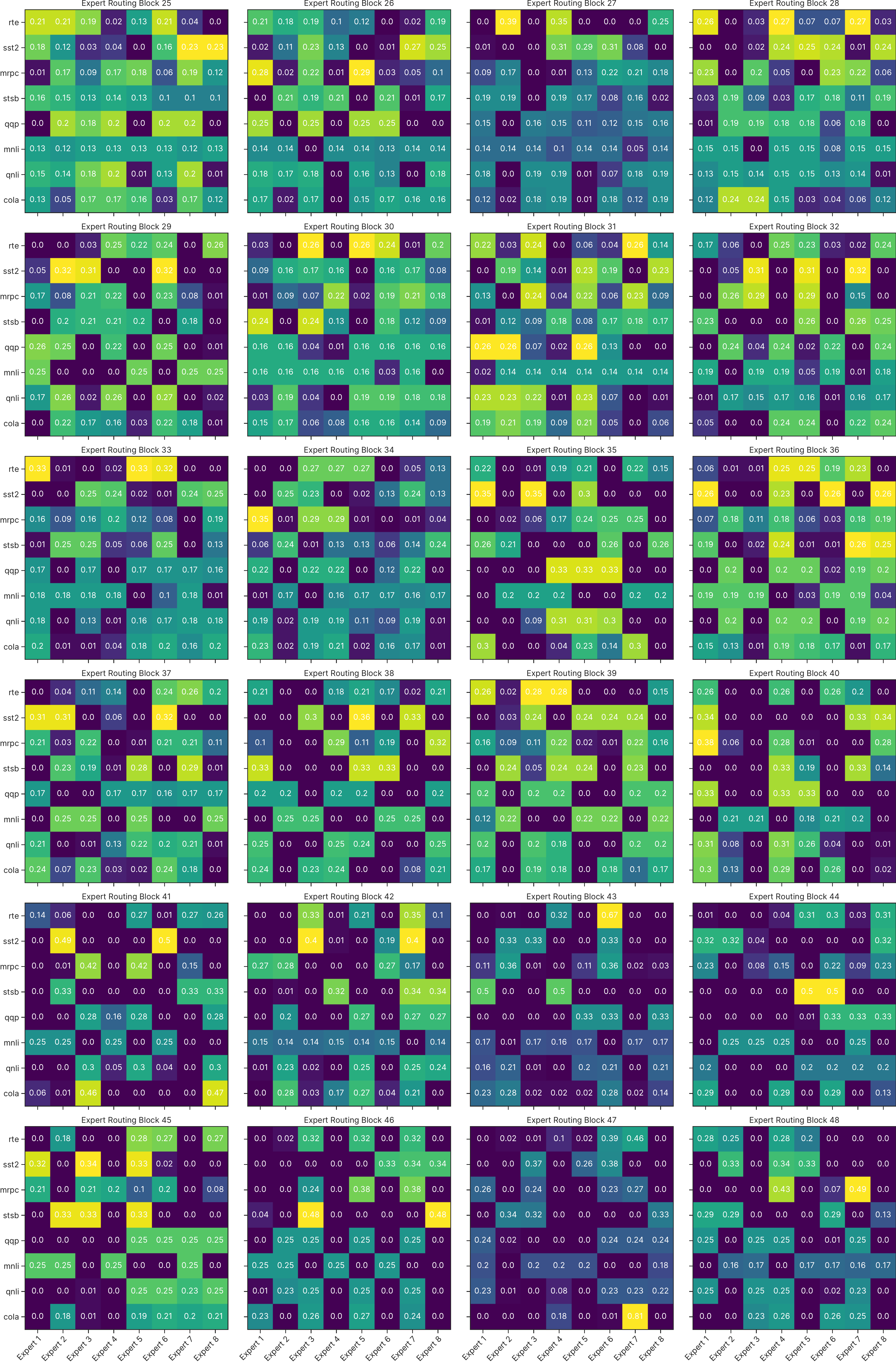}
    \caption{Routing distribution learnt by Latent Skills in the decoder routing blocks (25-48) of T5-GLUE}
    \label{fig:Latentskills_glue_routing_layer24to48}    

\end{figure*}

\begin{figure*}[h]
 
    \centering
    \includegraphics[width=0.9\textwidth]{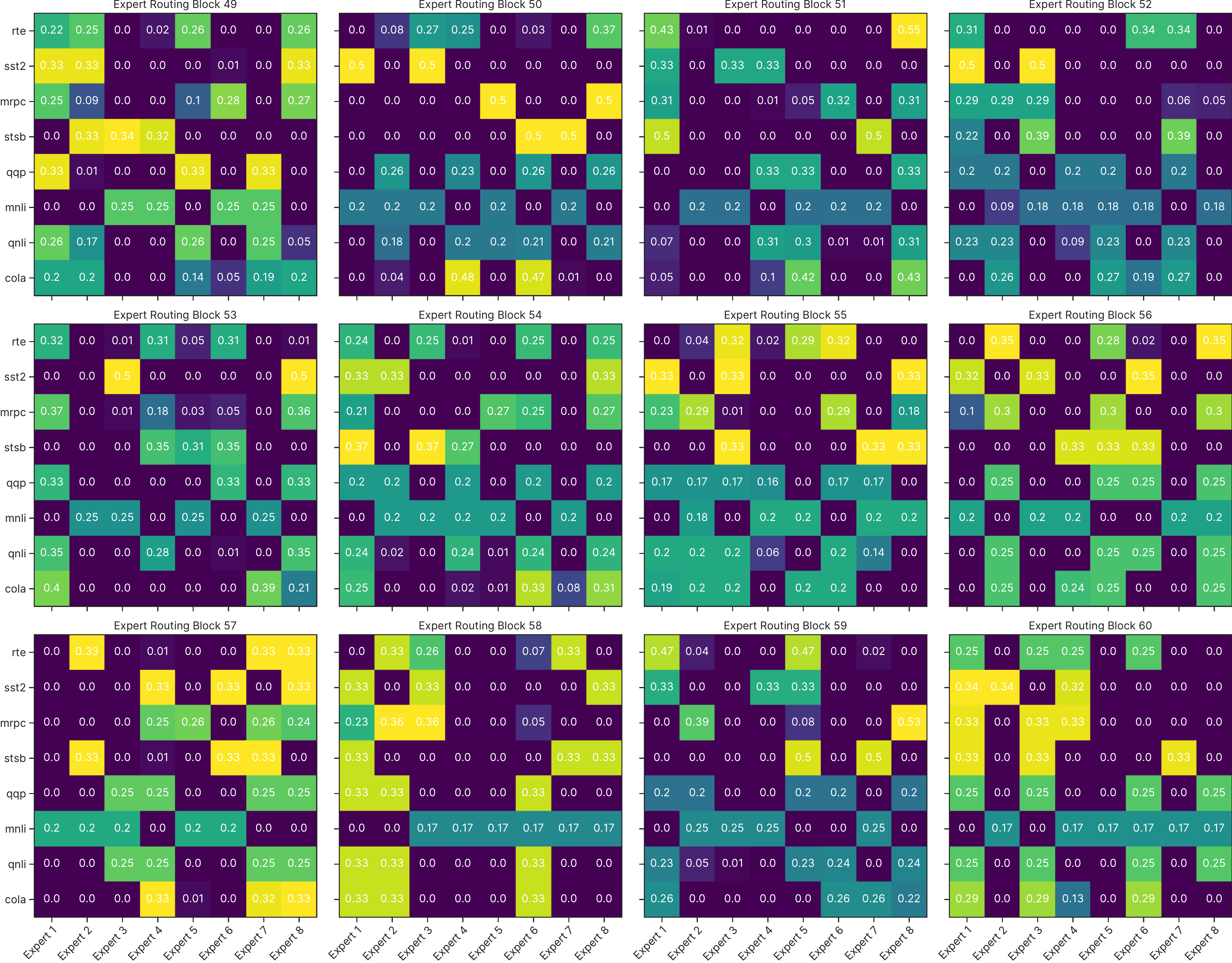}
    \caption{Routing distribution learnt by Latent Skills in the decoder routing blocks (49-60) of T5-GLUE}
    \label{fig:Latentskills_glue_routing_layer48to60}    

\end{figure*}

\begin{figure*}[h]
 
    \centering
    \includegraphics[width=0.9\textwidth]{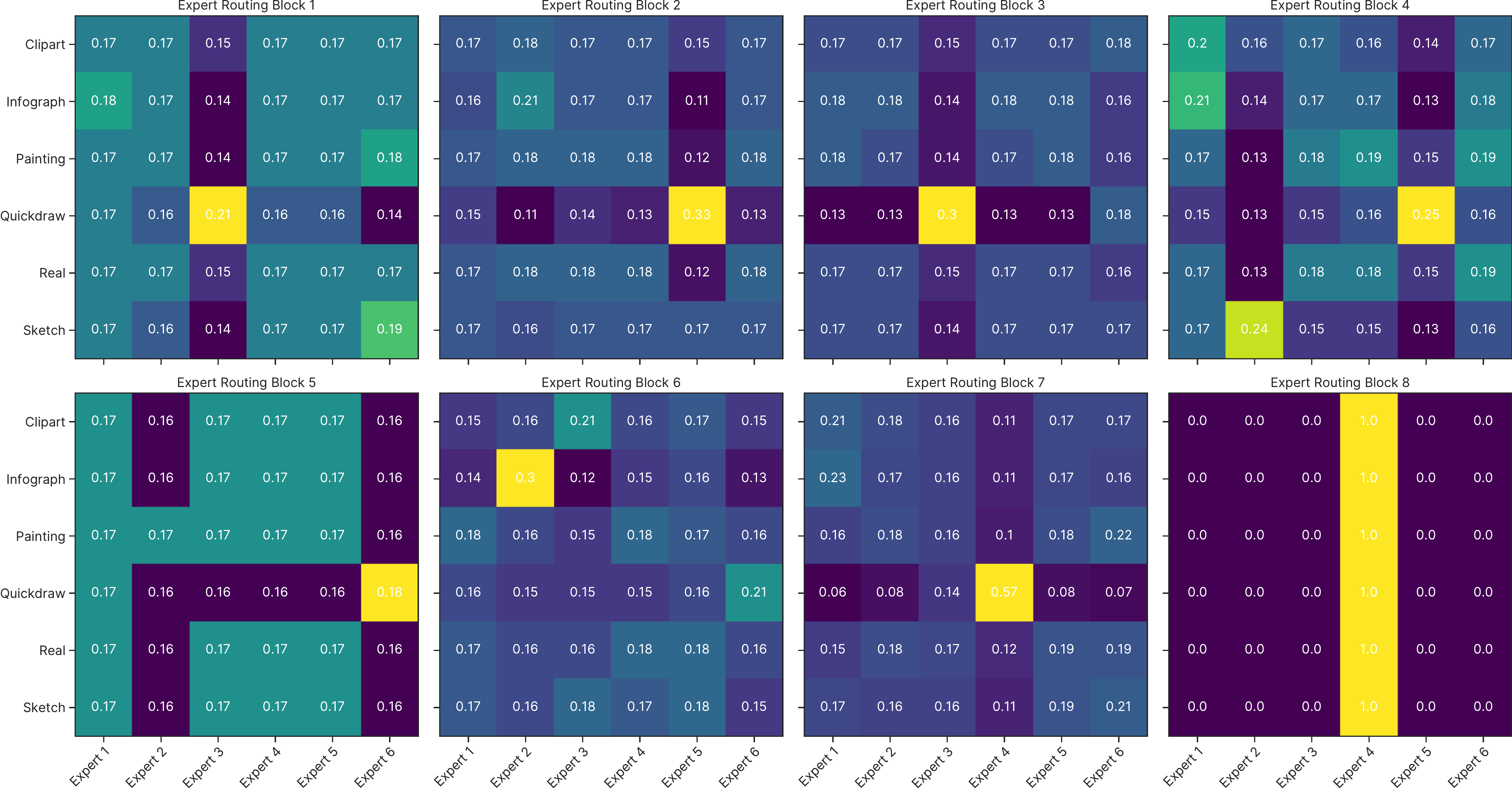}
    \caption{Routing distribution learnt by Top-$k$ routing in the routing blocks of ResNet-DomainNet}
    \label{fig:Topk_domainnet_routing}    

\end{figure*}

\begin{figure*}[htp]
 
    \centering
    \includegraphics[width=0.9\textwidth]{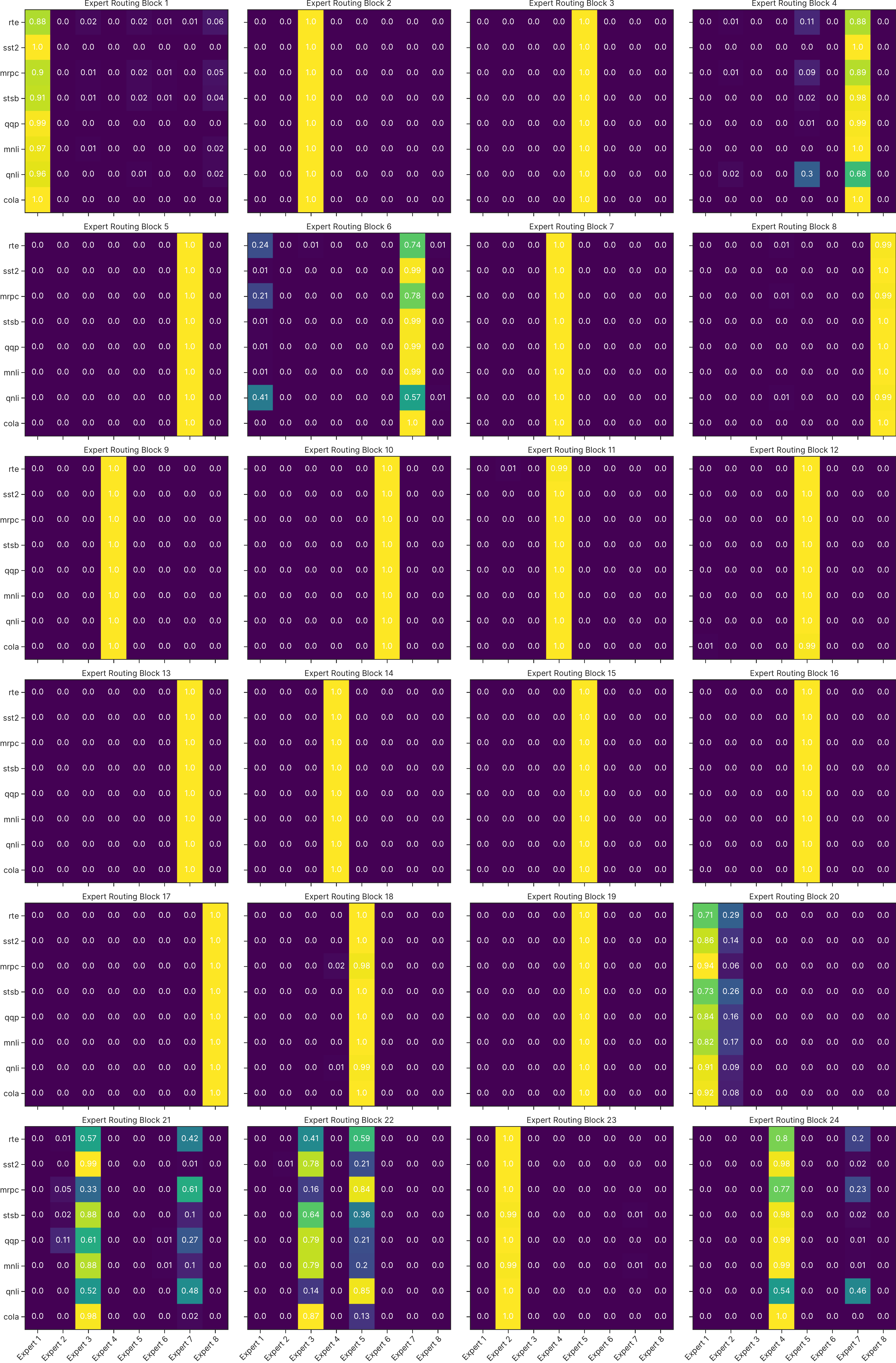}
    \caption{Routing distribution learnt by Top-$k$ in the encoder routing blocks (1-24) of T5-GLUE}
    \label{fig:Topk_glue_routing_layer0to23}    

\end{figure*}

\begin{figure*}[htp]
 
    \centering
    \includegraphics[width=0.9\textwidth]{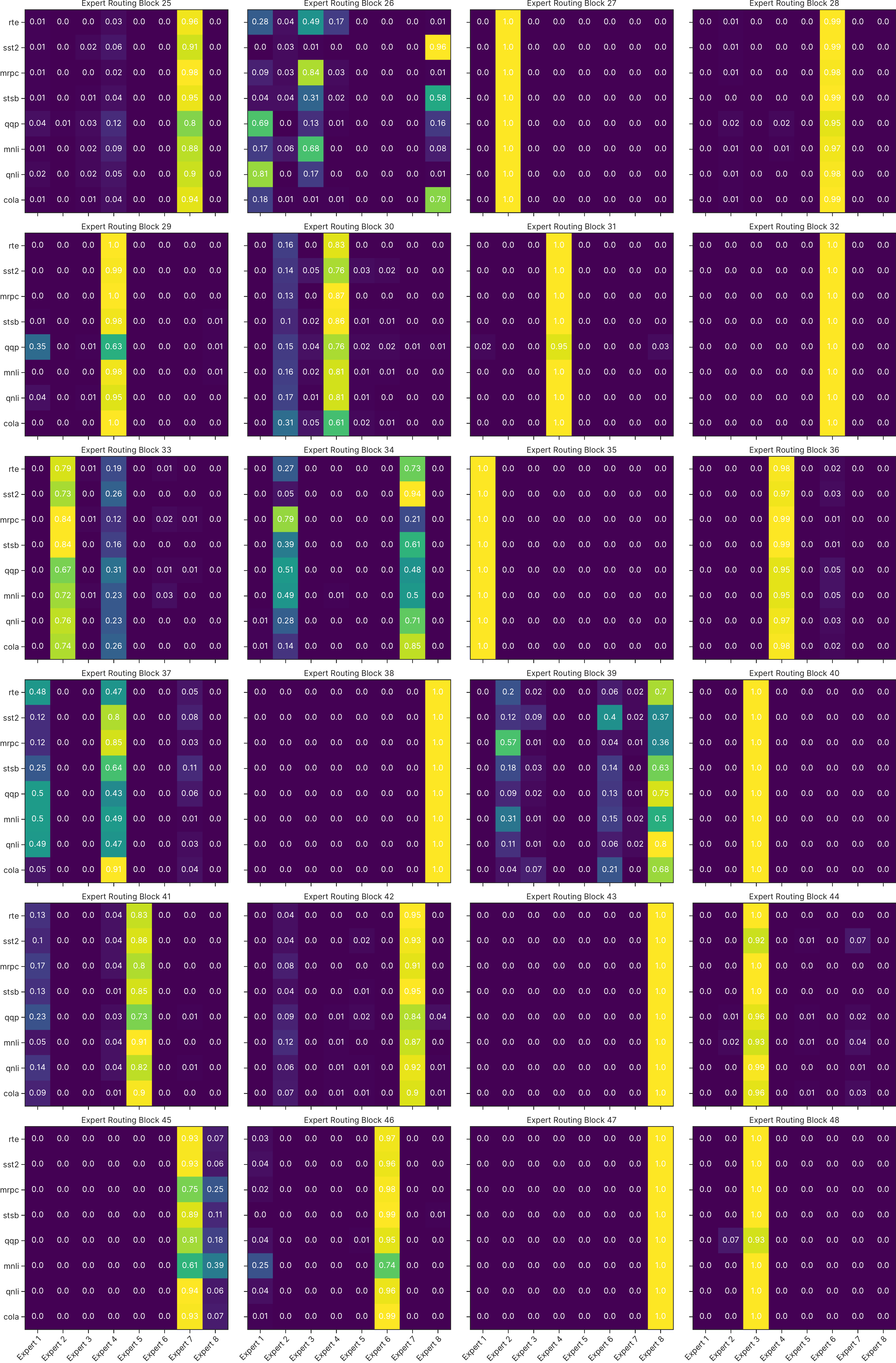}
    \caption{Routing distribution learnt by Top-$k$ in the decoder routing blocks (25-48) of T5-GLUE}
    \label{fig:Topk_glue_routing_layer24to48}    

\end{figure*}

\begin{figure*}[h]
 
    \centering
    \includegraphics[width=0.9\textwidth]{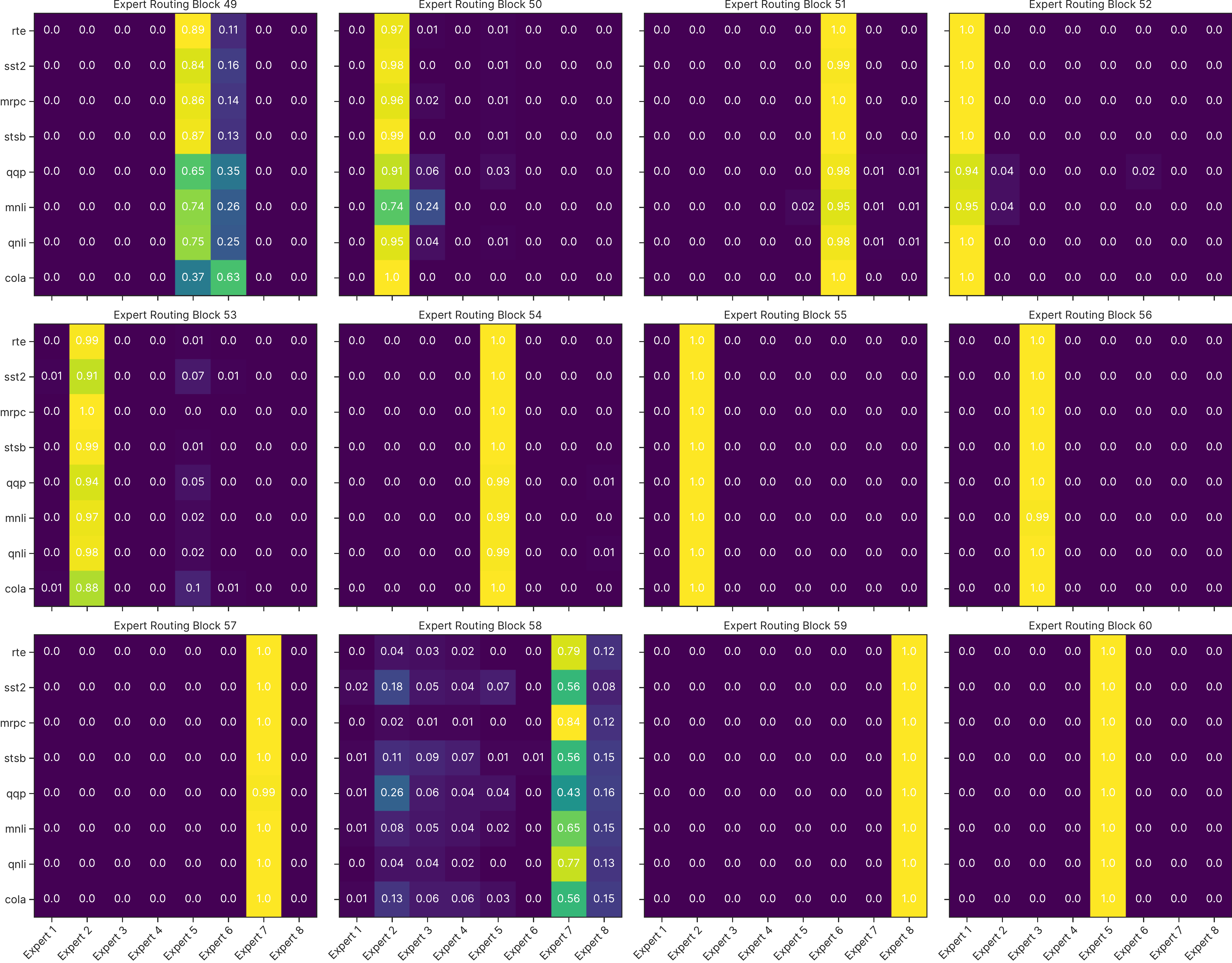}
    \caption{Routing distribution learnt by Top-$k$ in the decoder routing blocks (49-60) of T5-GLUE}
    \label{fig:Topk_glue_routing_layer48to60}    

\end{figure*}

\begin{figure*}[h]
 
    \centering
    \includegraphics[width=0.9\textwidth]{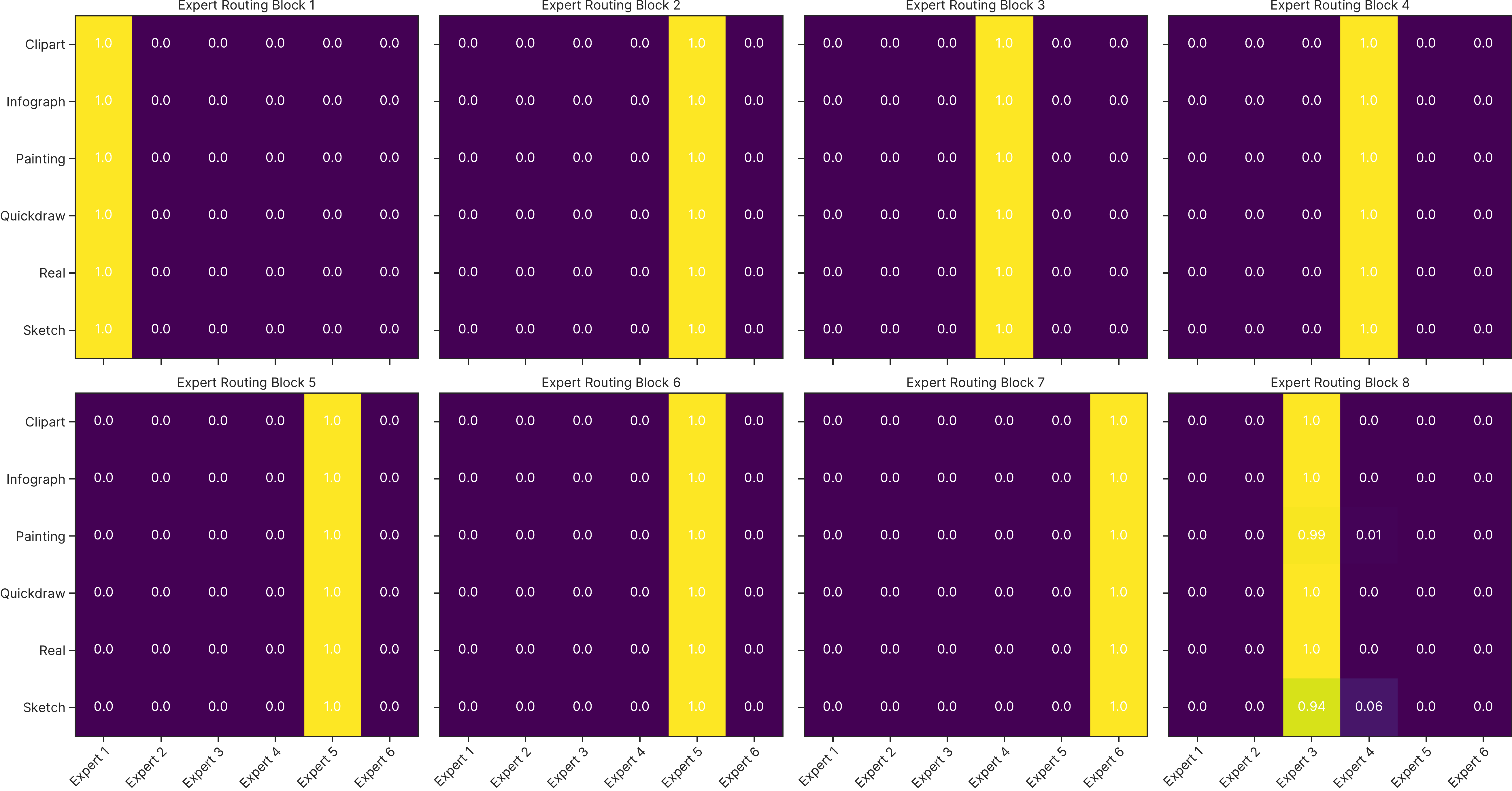}
    \caption{Routing distribution learnt by ST-Gumbel routing in the routing blocks of ResNet-DomainNet}
    \label{fig:STGumbel_domainnet_routing}    

\end{figure*}

\begin{figure*}[htp]
 
    \centering
    \includegraphics[width=0.9\textwidth]{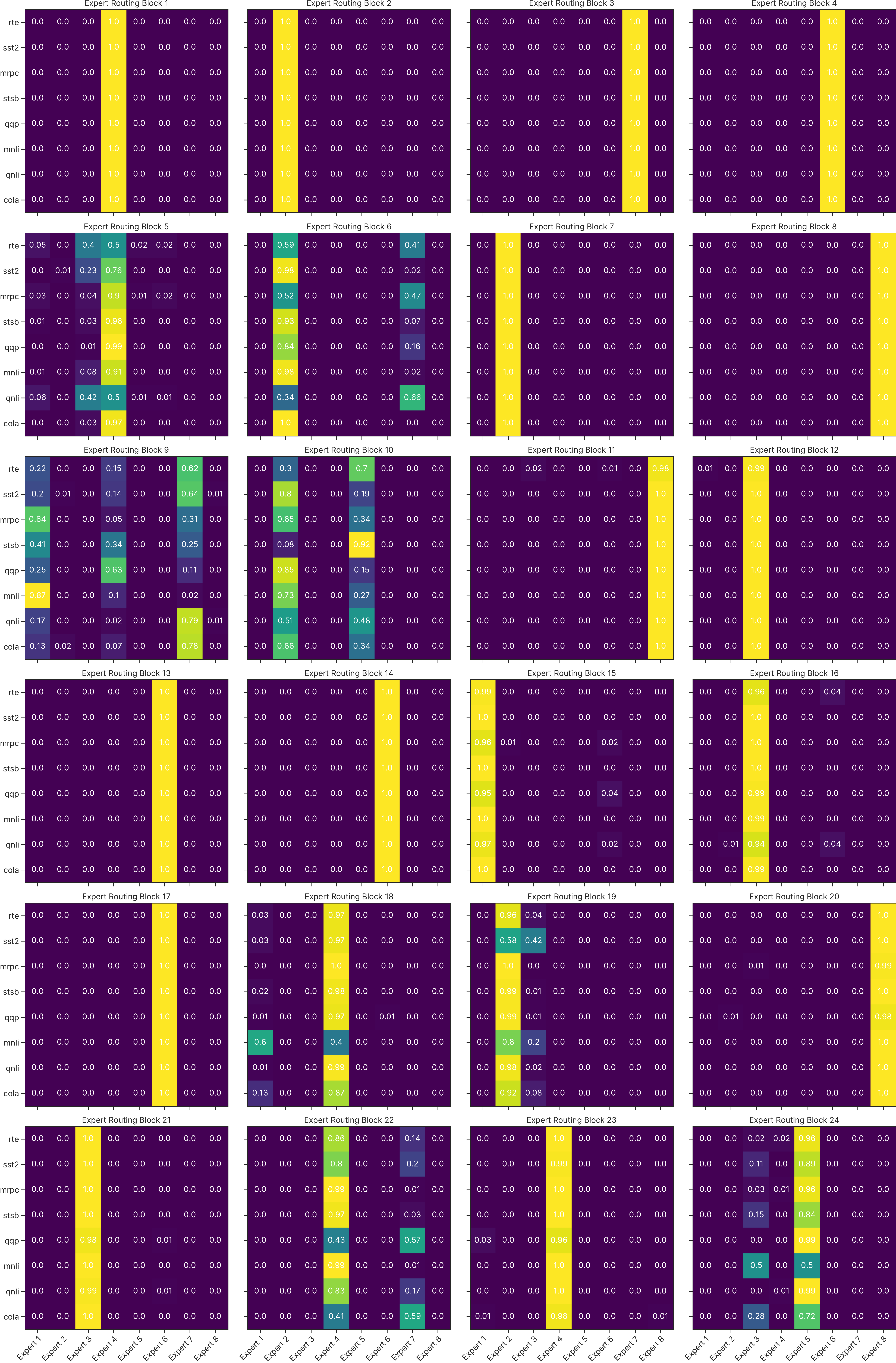}
    \caption{Routing distribution learnt by ST-Gumbel in the encoder routing blocks (1-24) of T5-GLUE}
    \label{fig:STGumbel_glue_routing_layer0to23}    

\end{figure*}

\begin{figure*}[htp]
 
    \centering
    \includegraphics[width=0.9\textwidth]{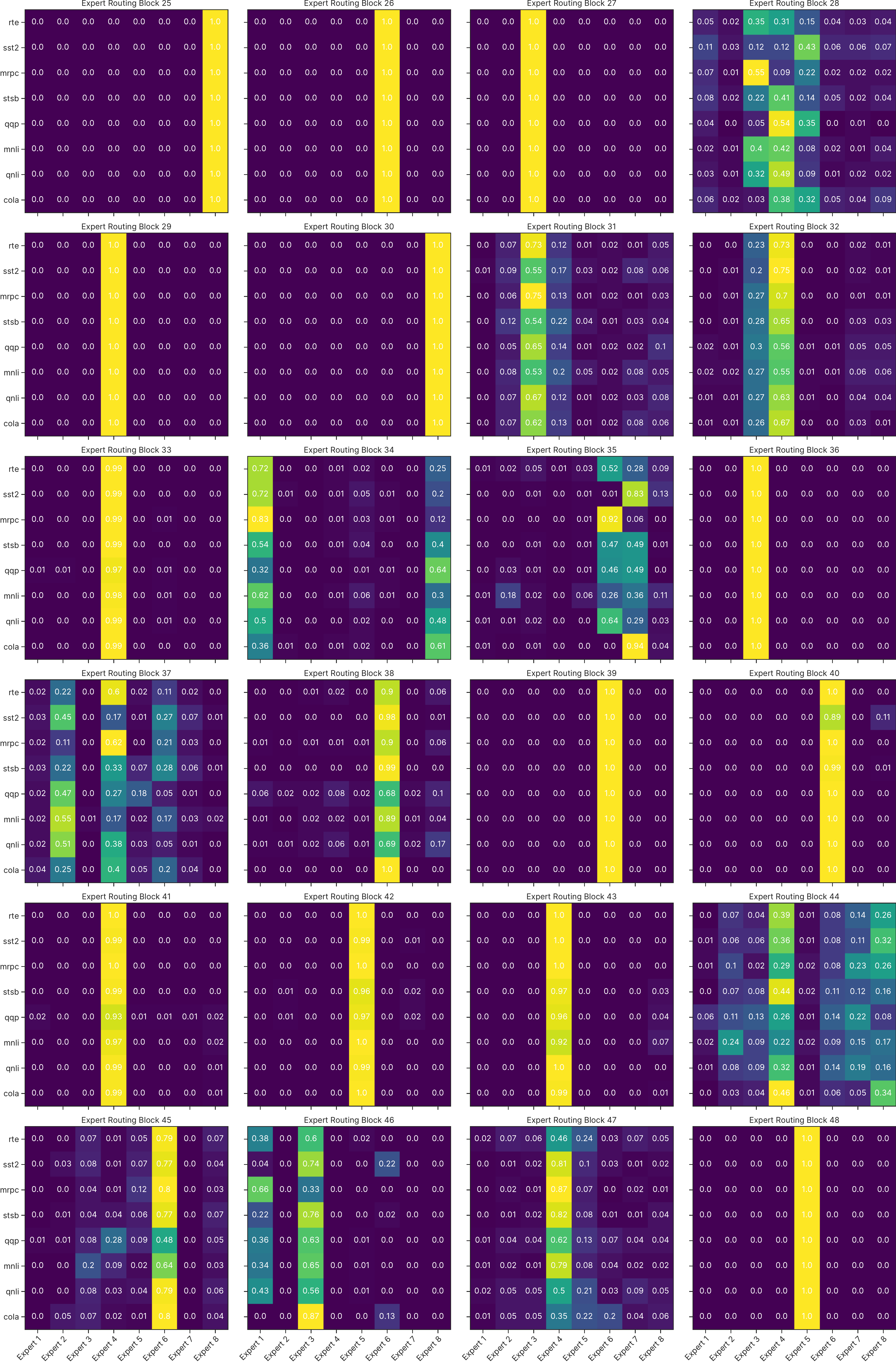}
    \caption{Routing distribution learnt by ST-Gumbel in the decoder routing blocks (25-48) of T5-GLUE}
    \label{fig:STGumbel_glue_routing_layer24to48}    

\end{figure*}

\begin{figure*}[h]
 
    \centering
    \includegraphics[width=0.9\textwidth]{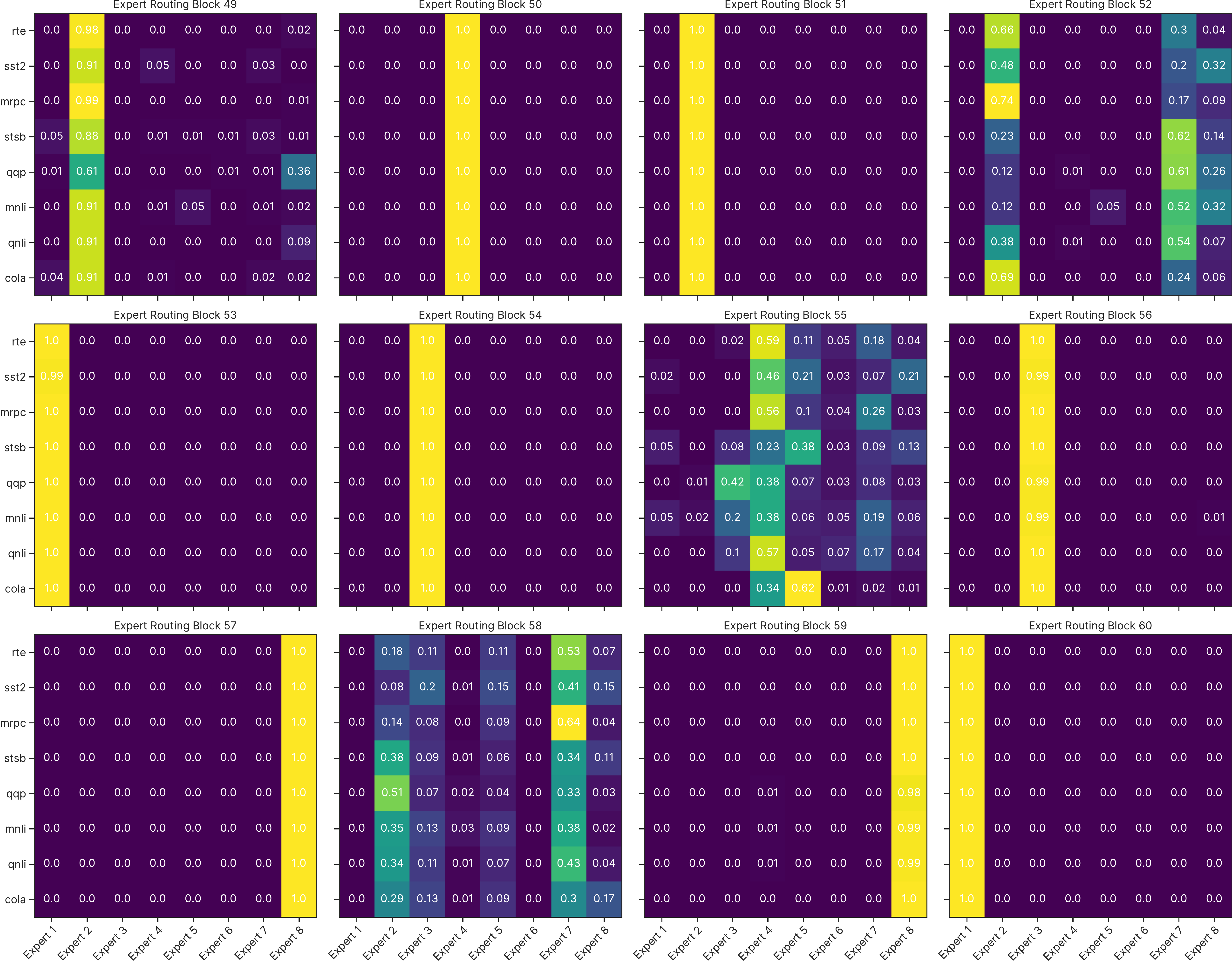}
    \caption{Routing distribution learnt by ST-Gumbel in the decoder routing blocks (49-60) of T5-GLUE}
    \label{fig:STGumbel_glue_routing_layer48to60}    

\end{figure*}

\begin{figure*}[h]
 
    \centering
    \includegraphics[width=0.9\textwidth]{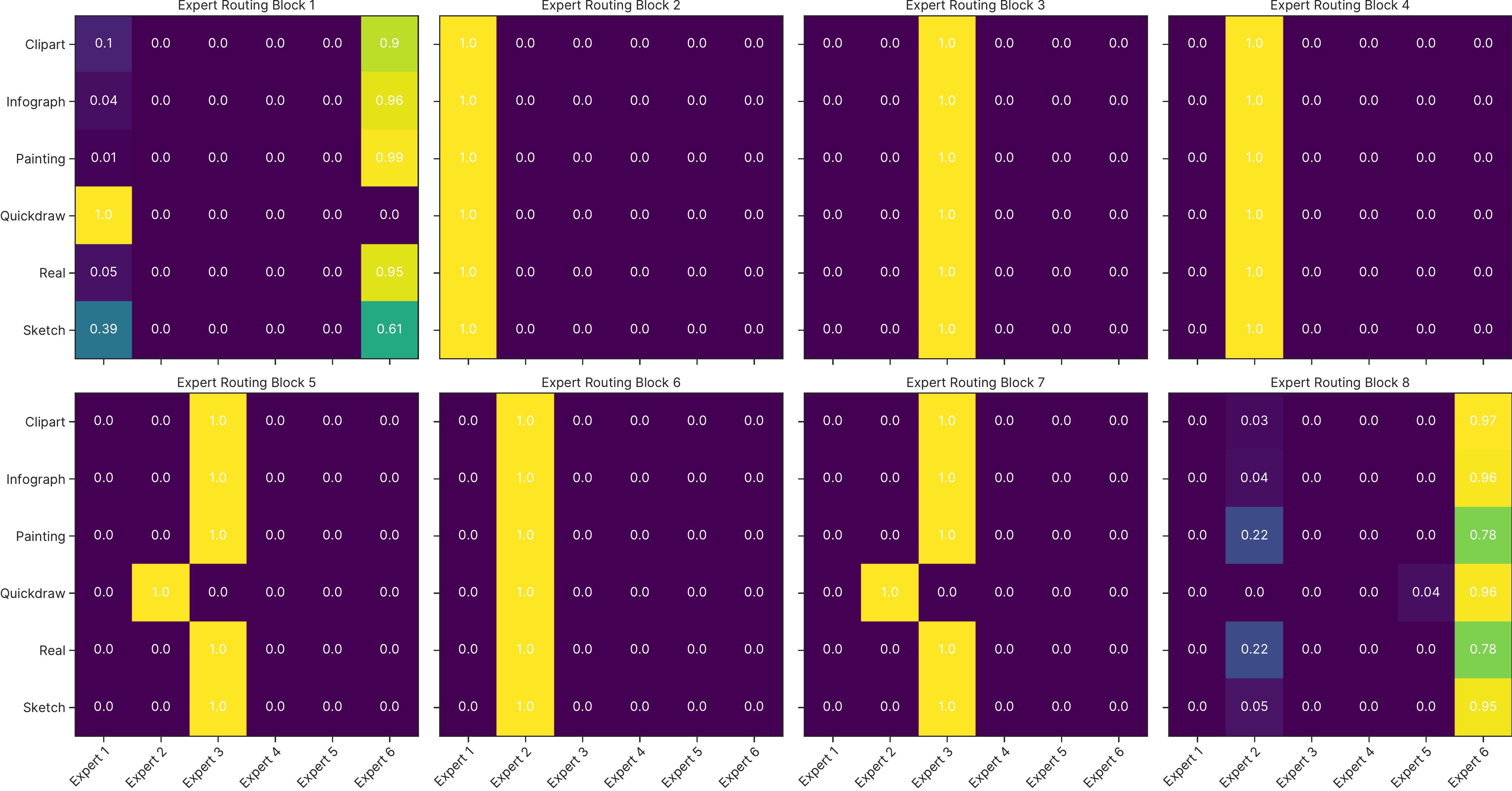}
    \caption{Routing distribution learnt by REINFORCE routing in the routing blocks of ResNet-DomainNet}
    \label{fig:Reinforce_domainnet_routing}    

\end{figure*}

\begin{figure*}[htp]
 
    \centering
    \includegraphics[width=0.9\textwidth]{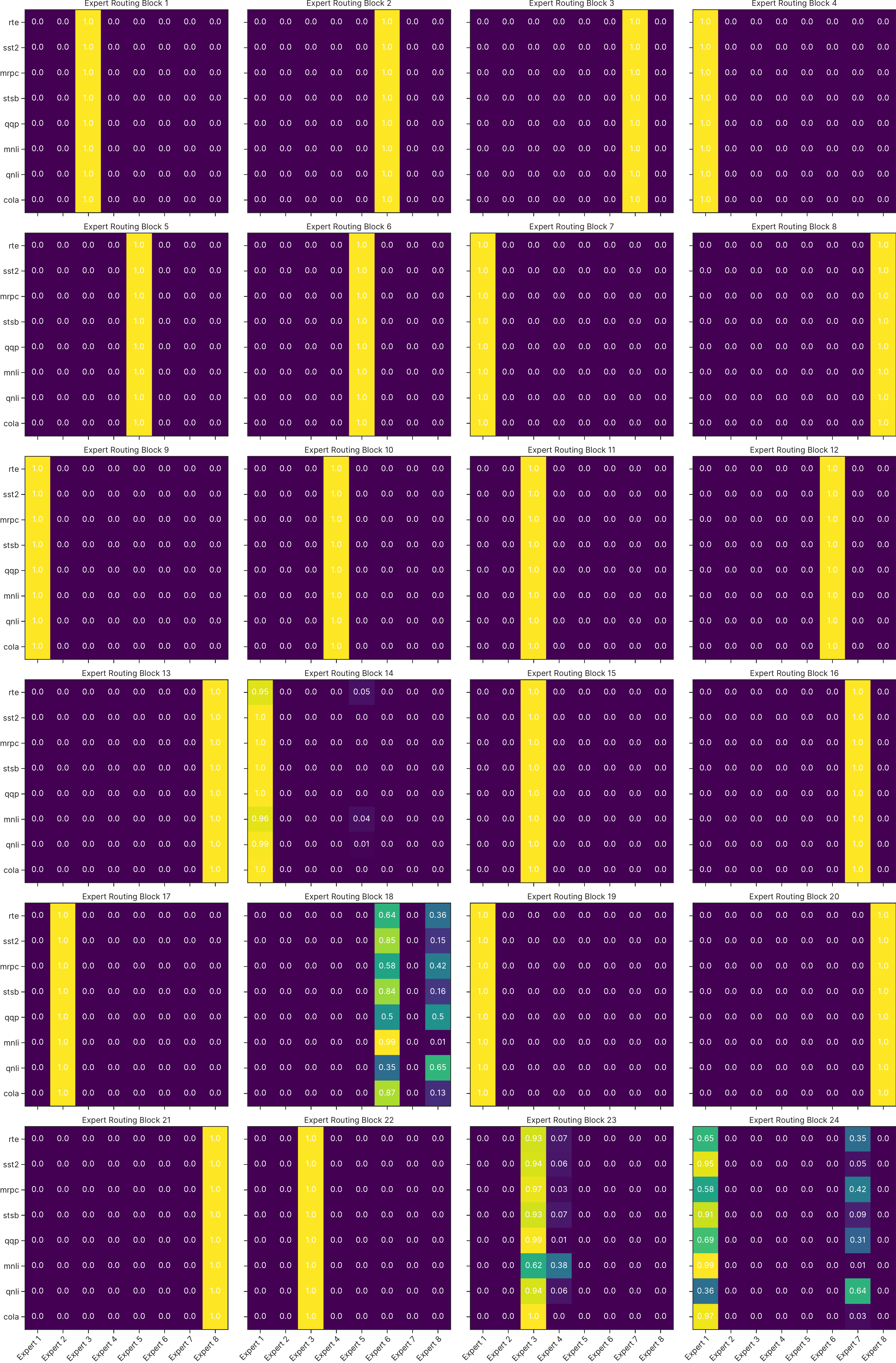}
    \caption{Routing distribution learnt by REINFORCE in the encoder routing blocks (1-24) of T5-GLUE}
    \label{fig:Reinforce_glue_routing_layer0to23}    

\end{figure*}

\begin{figure*}[htp]
 
    \centering
    \includegraphics[width=0.9\textwidth]{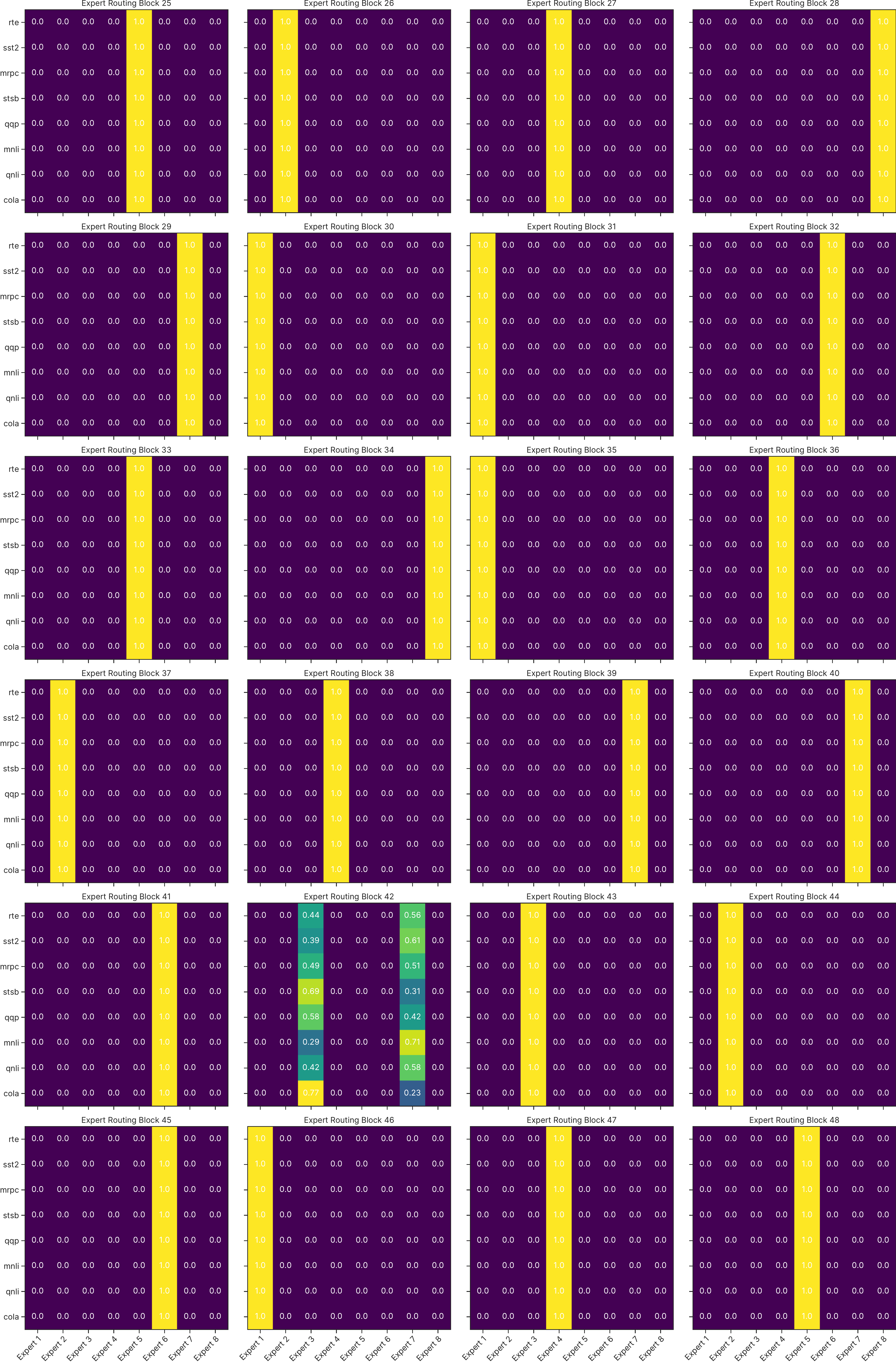}
    \caption{Routing distribution learnt by REINFORCE in the decoder routing blocks (25-48) of T5-GLUE}
    \label{fig:Reinforce_glue_routing_layer24to48}    

\end{figure*}

\begin{figure*}[h]
 
    \centering
    \includegraphics[width=0.9\textwidth]{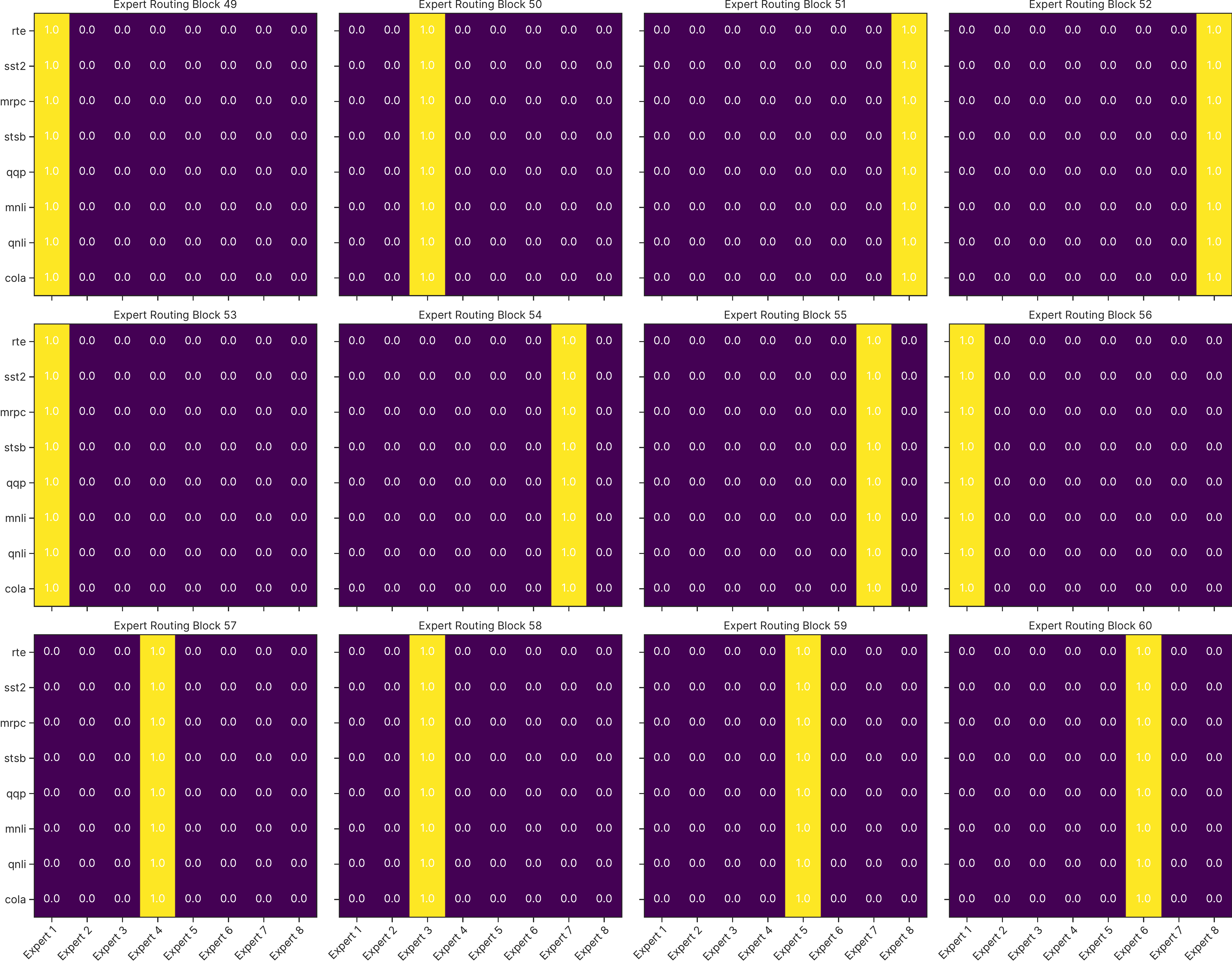}
    \caption{Routing distribution learnt by REINFORCE in the decoder routing blocks (49-60) of T5-GLUE}
    \label{fig:Reinforce_glue_routing_layer48to60}    

\end{figure*}

\begin{figure*}[h]
 
    \centering
    \includegraphics[width=0.9\textwidth]{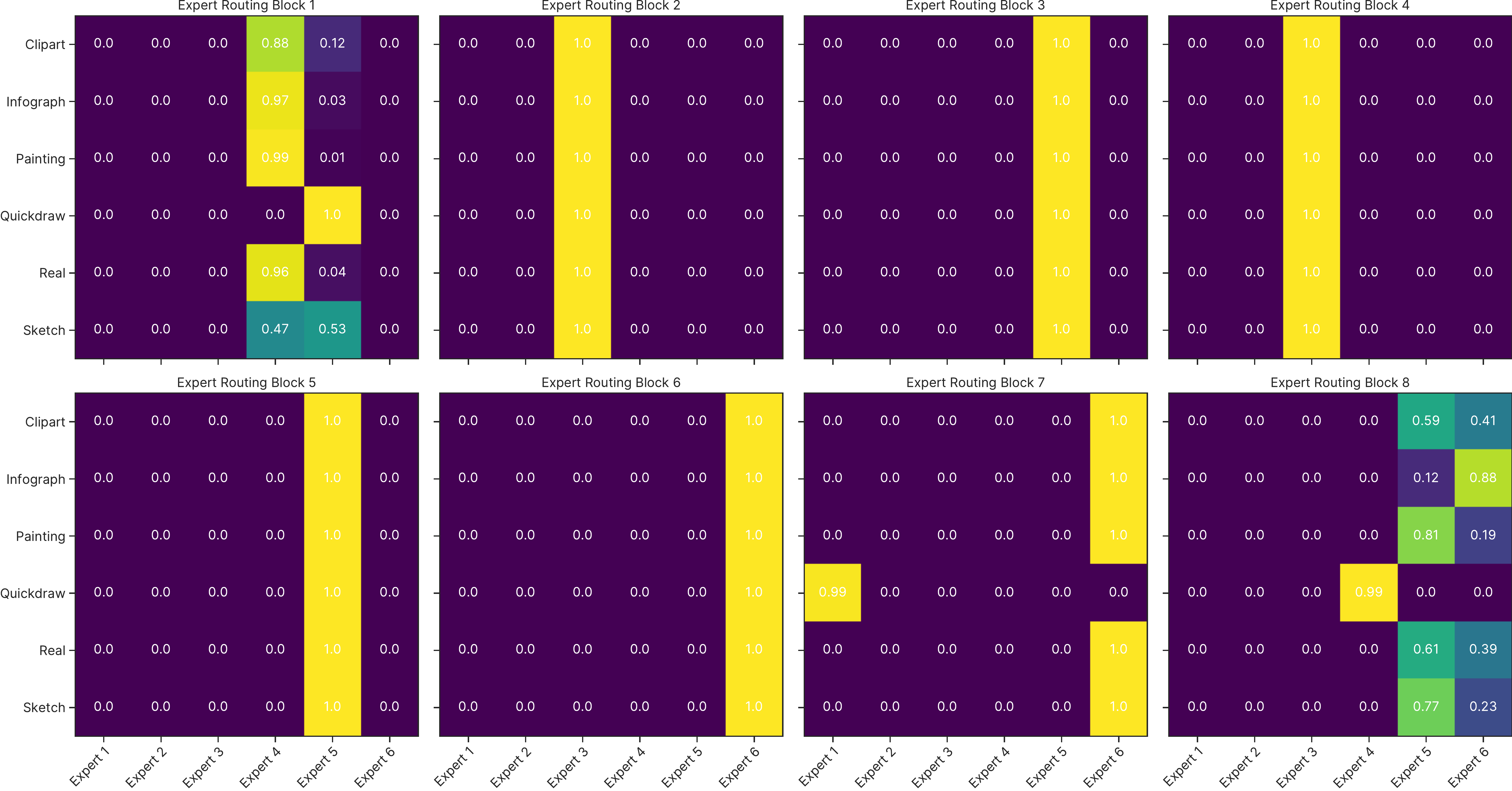}
    \caption{Routing distribution learnt by Dselect-$k$ routing in the routing blocks of ResNet-DomainNet}
    \label{fig:Dselectk_domainnet_routing}    

\end{figure*}

\begin{figure*}[h]
 
    \centering
    \includegraphics[width=0.9\textwidth]{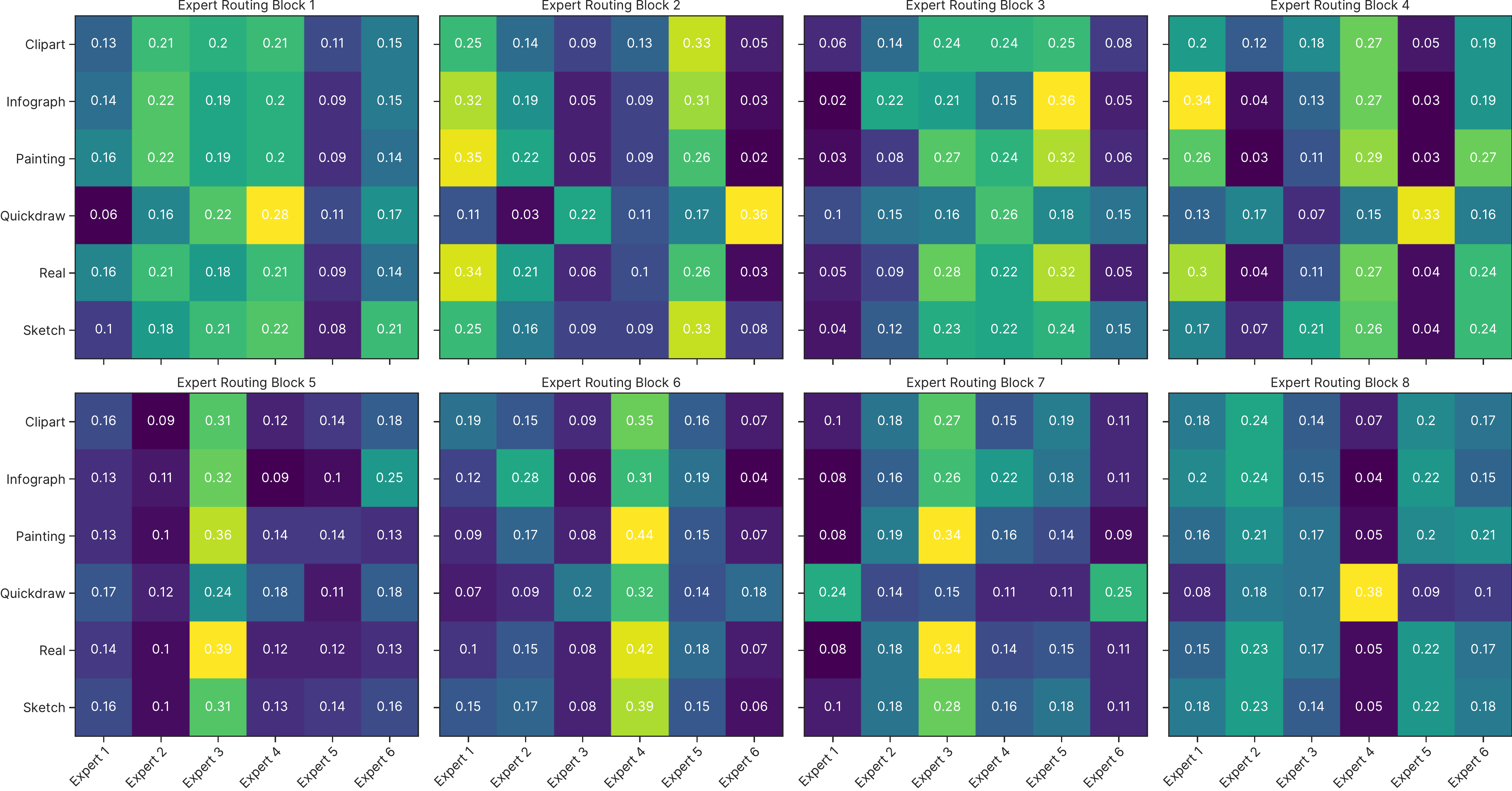}
    \caption{Routing distribution learnt by Ensemble routing in the routing blocks of ResNet-DomainNet}
    \label{fig:Ensemble_domainnet_routing}    

\end{figure*}

\begin{figure*}[htp]
 
    \centering
    \includegraphics[width=0.9\textwidth]{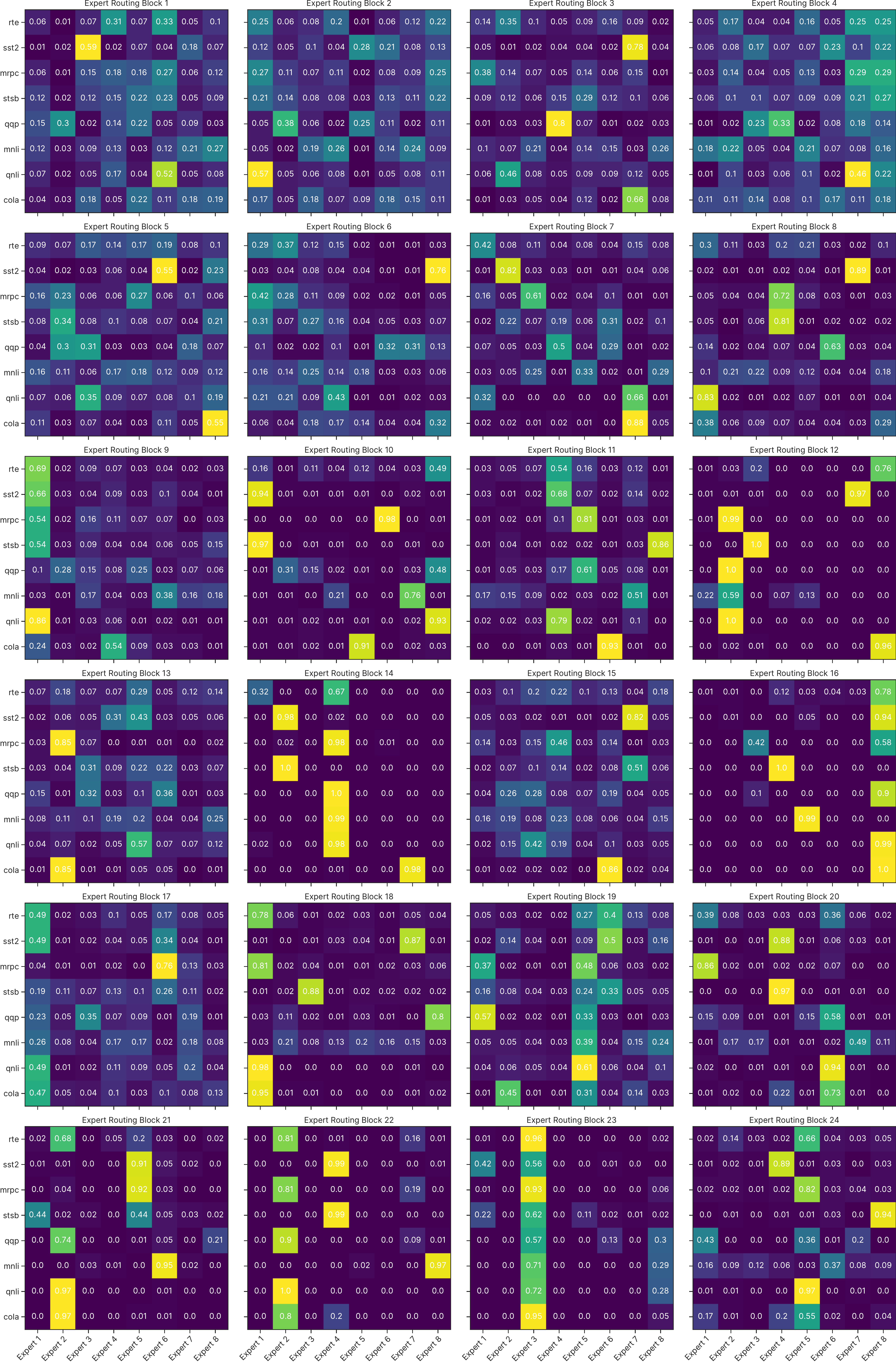}
    \caption{Routing distribution learnt by Ensemble in the encoder routing blocks (1-24) of T5-GLUE}
    \label{fig:Ensemble_glue_routing_layer0to23}    

\end{figure*}

\begin{figure*}[htp]
 
    \centering
    \includegraphics[width=0.9\textwidth]{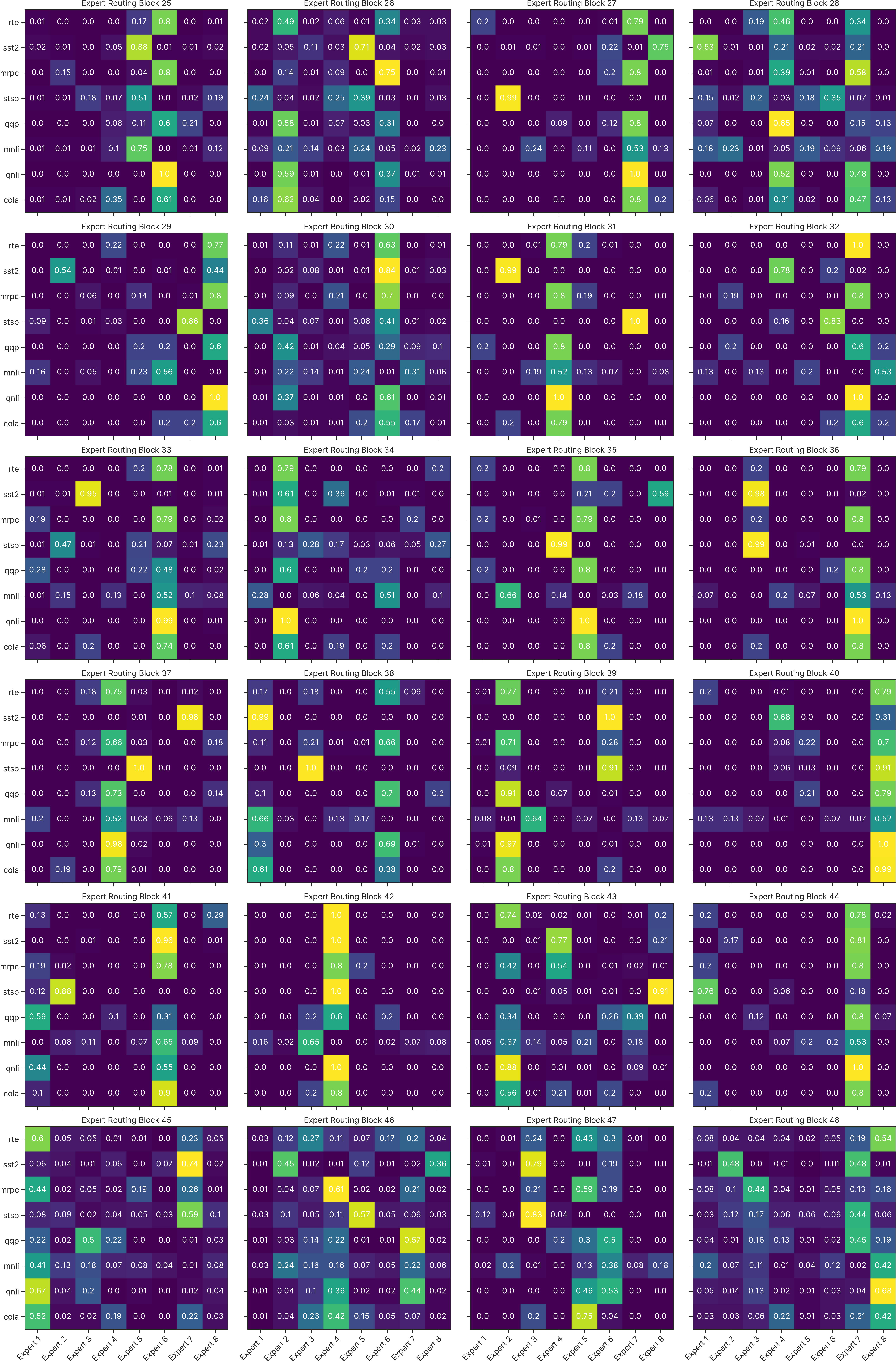}
    \caption{Routing distribution learnt by Ensemble in the decoder routing blocks (25-48) of T5-GLUE}
    \label{fig:Ensemble_glue_routing_layer24to48}    

\end{figure*}

\begin{figure*}[h]
 
    \centering
    \includegraphics[width=0.9\textwidth]{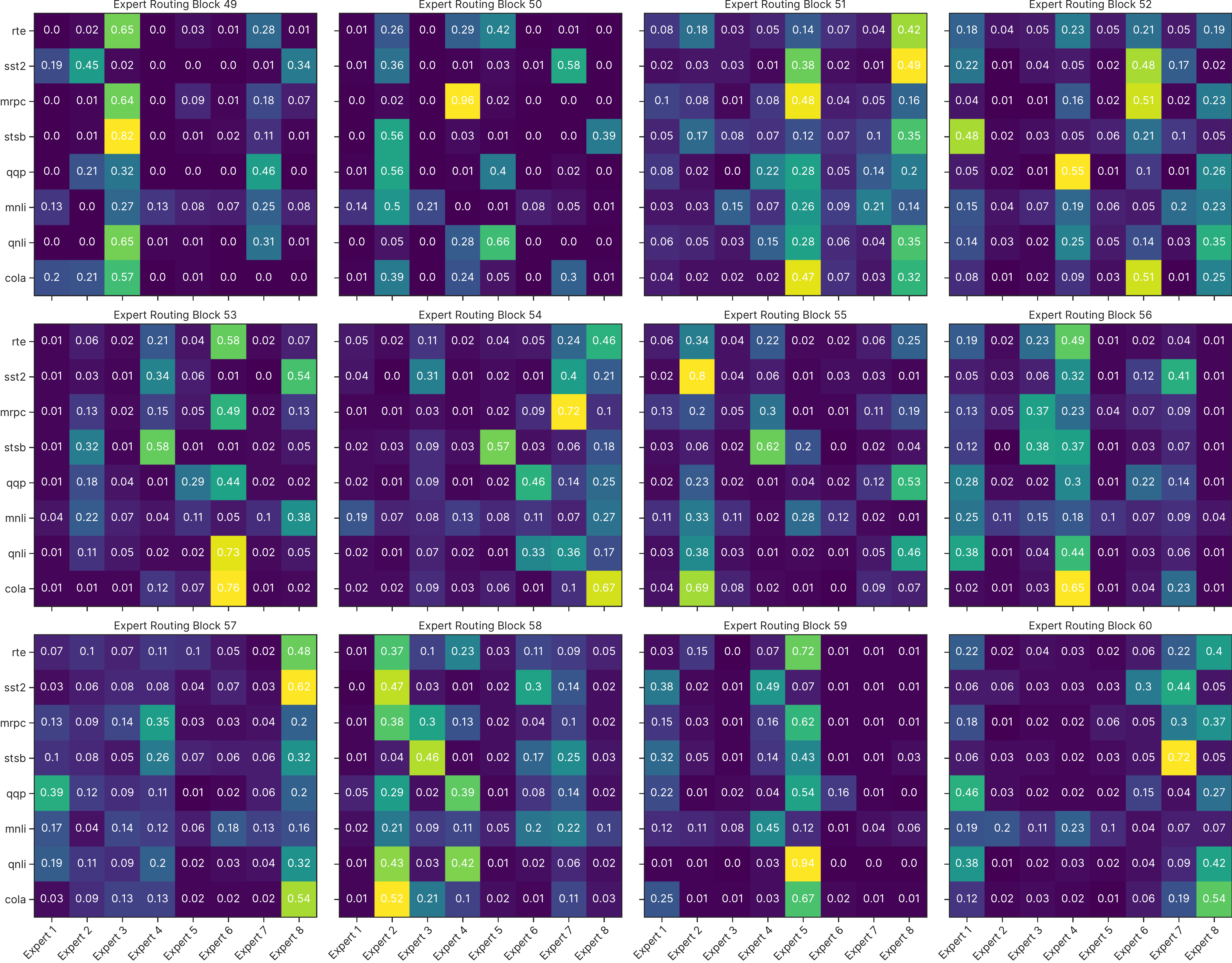}
    \caption{Routing distribution learnt by Ensemble in the decoder routing blocks (49-60) of T5-GLUE}
    \label{fig:Ensemble_glue_routing_layer48to60}    

\end{figure*}

For other learned routing methods, refer to the corresponding figures for detailed routing distributions:
\begin{itemize}
    \item \textbf{Latent Skills}: \cref{fig:Latentskills_domainnet_routing}, \cref{fig:Latentskills_glue_routing_layer0to23}, \cref{fig:Latentskills_glue_routing_layer24to48}, and \cref{fig:Latentskills_glue_routing_layer48to60}.
    \item \textbf{Top-$k$}: \cref{fig:Topk_domainnet_routing}, \cref{fig:Topk_glue_routing_layer0to23}, \cref{fig:Topk_glue_routing_layer24to48}, and \cref{fig:Topk_glue_routing_layer48to60}.
    \item \textbf{ST-Gumbel}: \cref{fig:STGumbel_domainnet_routing}, \cref{fig:STGumbel_glue_routing_layer0to23}, \cref{fig:STGumbel_glue_routing_layer24to48}, and \cref{fig:STGumbel_glue_routing_layer48to60}.
    \item \textbf{REINFORCE}: \cref{fig:Reinforce_domainnet_routing}, \cref{fig:Reinforce_glue_routing_layer0to23}, \cref{fig:Reinforce_glue_routing_layer24to48}, and \cref{fig:Reinforce_glue_routing_layer48to60}.
    \item \textbf{Dselect-$k$}: \cref{fig:Dselectk_domainnet_routing}
    \item \textbf{Ensemble}: \cref{fig:Ensemble_domainnet_routing}, \cref{fig:Ensemble_glue_routing_layer0to23}, \cref{fig:Ensemble_glue_routing_layer24to48}, and \cref{fig:Ensemble_glue_routing_layer48to60}.
\end{itemize}

\end{document}